# Sensitive Information Detection: Recursive Neural Networks for Encoding Context

Jan Neerbek

PhD Dissertation

Department of Computer Science
Aarhus University
Denmark

# Sensitive Information Detection: Recursive Neural Networks for Encoding Context

A Dissertation
Presented to the Faculty of Science and Technology
of Aarhus University
in Partial Fulfillment of the Requirements
for the PhD Degree

by
Jan Neerbek
June 30, 2020

# Abstract

The amount of data for processing and categorization grows at an ever increasing rate. At the same time the demand for collaboration and transparency in organizations, government and businesses, drives the release of data from internal repositories to the public or 3rd party domain. This in turn increase the potential of sharing sensitive information. The leak of sensitive information can potentially be very costly, both financially for organizations, but also for individuals. In this work we address the important problem of sensitive information detection. Specially we focus on detection in unstructured text documents.

We show that simplistic, brittle rule sets for detecting sensitive information only find a small fraction of the actual sensitive information. Furthermore we show that previous state-of-the-art approaches have been implicitly tailored to such simplistic scenarios and thus fail to detect actual sensitive content.

We develop a novel family of sensitive information detection approaches which only assumes access to labeled examples, rather than unrealistic assumptions such as access to a set of generating rules or descriptive topical seed words. Our approaches are inspired by the current state-of-the-art for paraphrase detection and we adapt deep learning approaches over recursive neural networks to the problem of sensitive information detection. We show that our *context-based* approaches significantly outperforms the family of previous state-of-the-art approaches for sensitive information detection, so-called *keyword-based* approaches, on real-world data and with human labeled examples of sensitive and non-sensitive documents.

A key challenge in the field of sensitive information detection is the lack of publicly available real-world datasets on which to train and/or benchmark on. This is due to the inherent sensitive nature of the data in question. We address this issue in this work by releasing publicly labeled examples of sensitive and non-sensitive content. We release a total of 8 different types of sensitive information over 2 distinct sets of documents. We utilize efforts by human domain experts in labeling both datasets for 4 complex types of informational content for each set of documents. This release totals $750,000$ labeled sentences with their parse trees for the research community to make use of.

# Resumé

Mængden af information tilgængeligt som skal kunne automatisk håndteres og bearbejdes vokser eksplosivt. Dette sker samtidigt med øget fokus på deling af data og krav om transparens. Dette øger risikoen for deling af potentielt følsomme oplysninger som ikke skulle have været delt. Sådanne fejlagtige delinger og afsløringer af følsomme oplysninger er forbundet med høje omkostninger. I denne afhandling adresseres det voksende og komplekse problemområde omkring at finde følsomme informationer ved hjælp af datalogiske algoritmer. Specifikt fokuseres på at finde følsomme oplysninger i ustrukturerede tekst dokumenter.

Vi påviser at simple regelsæt kun finder en relativ lille del af de faktiske følsomme oplysninger i tekst dokumenter. Vi påviser også at tidligere udgivende algoritmer til at finde følsom information har båret en indbygget svaghed således at disse algoritmer kun kan finde simpel følsom information.

Vi udvikler og beskriver en hel ny familie af algoritmer til at finde følsom informationer. Vi antager adgang til eksempler af følsomme og ikke-følsomme dokumenter. Dette er i kontrast til tidligere algoritmer med potentielt urealistiske antagelser, så som at al følsom information kan indfanges af små sæt af emneord. Vores nye algoritmer er inspireret af algoritmer indenfor "paraphrase detection". Vi tilpasser og forandre disse algoritmer baseret på dybe rekursive kunstige neurale netværk. Vi efterviser eksperimentelt at vores algoritmer er signifikant bedre til at finde følsom information end de tidligere udviklede og anvendte, såkaldte emneord-baserede algoritmer.

En stor udfordring for den forsatte forskning, udvikling og kvalitetssikring indenfor det samfundsmæssige vigtige felt; at finde og beskytte følsomme informationer, er manglen på adgang til relevant data. Da følsomme informationer, i sagens natur, er følsomme, er der en absolut mangel på datasæt som kan bruges til at forske og udvikle nye tilgange og algoritmer. Dette fundamentale problem adresseres i denne afhandling. Vi frigiver 8 forskellige typer af følsomme oplysninger over 2 meget forskellige datasæt af dokumenter. Vi anvender angivelser af følsomhed foretaget af domæne eksperter. For hvert datasæt frigiver vi 4 dokument-niveau angivelser af forskellige slags følsomme informationer. Sammenlagt frigiver vi mere end 750.000 sætninger med angivelse af følsomhed og med semantiske træer over sætningerne som forskere og udviklere kan gøre brug af.

# Acknowledgments

This dissertation would not have been possible without the support and help from many people around me – I am immensely grateful to all of you!

A special thanks goes to my advisor, Ira Assent, for many fruitful discussions, invaluable feedback and idea bouncing, and also for believing in the project even when the going was tough. Also a special thanks to Peter Dolog from Aalborg University who was always ready for another visit and who provided directions, in-sights and goals for all half-baked ideas I came up with. I have learned a lot from preparing manuscripts together with both of you!

A great thanks to my employment place, Alexandra Institute, which provided funding for the during of my PhD project. A kind thanks to Mirko Presser who made this journey possible, Peter Andersen for taking the chance and Anders Kofod-Petersen for wanting to get this done the right way.

A shout-out to my fellow group members at the Data-Intensive Systems Group, both past and present for facilitating a creative and friendly researching environment. A thanks to visiting Professor/PostDoc, Leon Derczynski, for suggesting recursive neural networks to me and friendly talks and valuable feedback through the years.

A special shout-out to my fellow PhD student, Manuel Rafael Ciosici, for interesting talks, crazy ideas, gym, way too many burgers and an occasional hug.

A thank you to Susanne Bødker and Aslan Askarov for being part of my PhD support group and for discussions on studying, academia and life in general.

Lastly I want to thank my family for bearing with me during the PhD project and a very special thanks goes to my wife, Mette, who is always there for me. Thank you!

*Jan Neerbek,*
*Aarhus, June 30, 2020.*

# Contents

# Part I

# Overview

# Chapter 1

# Introduction

Collaboration requires sharing of information. This is particularly true for large organizations such as companies or public institutions. Such organizations face the need to handle and share large sets of documents without sharing or leaking any potentially sensitive information. The work in this dissertation is about the important problem of detecting such sensitive information in text documents.

Text documents which are to be shared, whether only with some 3rd parties or released publicly, may contain information that is *sensitive* in nature. There are many types of information which may be sensitive. For companies examples of such sensitive information are intellectual property, or progress on new products. For public institutions or other organizations this could be information on ongoing dealings, information on strategy or similar. This information can be personal information about a specific individual. For instance information such as name, address, gender, specific sicknesses, etc. This kind of sensitive information is referred to as *private* information.

Any documents which contain sensitive information need to be protected, i.e., the sensitive information must not be publicized. A key challenge in this setting is to distinguish the sensitive information from the non-sensitive information. This is known as *sensitive information detection*, i.e. identifying sensitive information in text documents [10, 16, 38, 42, 84].

Sensitive information detection is of great importance in a number of applications, where unintended leak of sensitive information may incur sever negative consequences for businesses or authorities. The need is also driven by a growing trend of transparency in both governmental institutions and in private companies. For instance, the recent years has seen a number of government laws and initiatives to promote the exchange and openness of information, e.g. the Freedom of Information Act in USA[1] and EU[2] and more recently the GDPR protection regulation[3].

[1] `https://www.foia.gov/`

[2] `http://www.europarl.europa.eu/RegData/PDF/r1049_en.pdf`

Due to this growth in data intended for release, the risk of accidentally leaking sensitive information has increased tremendously. The consequences of unintended leakage of sensitive information to the public (or to the competition) may be very damaging for public authorities or businesses. To put this into perspective, the fine for not processing sensitive data according to regulation can be as large as 4% of yearly turnover for companies according to the GDPR regulation[4]. However the cost in terms of lost business data and customer trust may be even bigger: in 2017 the Poneman Institute conducted an independent study, sponsored by IBM. The study calculates cost over the complete life cycle of a data breach, i.e., detection, escalation, notification (and mitigation), and after-the-fact (ex-post) response [81, 82]. In the study it was found that the cost to an organization, on average, for a data breach of sensitive information is $3.6 million.

The detection of sensitive information is important to organizations, both businesses and public institutions. However, sensitive information detection is also important on the individual personal level. There can be both social and financial costs in connection with leak of individual private information.

As 80% of all information exchanged is unstructured [28] this also means that a large fraction of leaked sensitive information comes from unstructured, text documents. The Freedom of Information initiative requires millions of text documents need to be screened yearly [84]. Thus, the number of documents becomes so large that manual, human inspection becomes very costly or even intractable.

In this dissertation we investigate sensitive information detection in text documents using ML algorithms. A Machine Learning (ML) algorithm is an algorithm which learns from data. An often used definition is "A computer program is said to learn from experience E with respect to some class of tasks T and performance measure P, if its performance at tasks in T, as measured by P, improves with experience E" [72].

In this work we refer to an ML algorithm as a "model". We have some data from the task T and we update the model to make the performance (for example, accuracy of prediction) better when evaluated on the task.

For certain information types, sensitivity may be centered around certain words. For instance, mentions of AIDS in connection to an individual and whether an individual is sick, are considered sensitive. However other types of sensitive information may be more *complex* (see Chapter 2 for details) in the sense that they cannot be easily identified via the presence of certain words alone. For example certain financial transactions may be considered sensitive for a company but other financial transaction are perhaps not sensitive. The two types of financial transactions may be described, using similar words, as in the example below. Thus, for a complex sensitive information type the *context* in which a word appears may influence whether we deem the word to be sensitive. In this work context denotes the sentence or document in which the word appear, which may influence whether a text is considered sensitive. Sensitivity in relation to context turns out to be a critical property of complex sensitive information and indeed is a main theme in this dissertation.

[3] `https://www.eugdpr.org`

[4] `http://europa.eu/rapid/press-release_MEMO-18-387_en.htm`

Next we give an example of complex sensitive information from our released, labeled dataset [75] which is derived from the public available Enron dataset. In one of the sensitive information tasks from our released datasets we have labels for the sensitive information type covering the financial concept of "prepay transactions". An example of a possible sensitive leak is a short email containing the following main text:

```
Attached are the credit worksheets for Royal Bank of
Canada and Bow River Trust.  As we discussed we have
proposed letters of credit for the approved form of
collateral pending further discussion with Treasury
regarding funding impact.  We may have to move to cash
margining if necessary.
```

This text does not contain the word *prepay* but still this short email is about a prepay transaction. A co-occurrence word model assign probability to each word that this word is indicative of the sensitive topic. As we show in our experiments keyword-based models assign low probability to all words in the example above and thus mis-classifies the text. However using models that consider a full sentence at a time we develop approaches which detect semantic sensitive information as in the above example.

To address the particular needs when detecting such *complex* sensitive information types we create a new family of approaches which take context into account and are thus suited for complex sensitive information detection. We compare empirically with previous approaches and finds that our family of approaches find previously undetected sensitive information.

A key challenge when working with sensitive information detection is that actual labeled sensitive information is not commonly available. Due to the nature of the type of data there is a natural barrier in obtaining sensitive data from organizations as they do not wish their sensitive information to be made public [5].

The lack of public datasets with labeled sensitive information also makes it difficult to compare or benchmark different approaches against each other. Previous works report accuracies which are typically based on different datasets and different definitions of sensitive information and are thus difficult to compare and evaluate.

Furthermore as we discuss in Chapter 4.2 for realistic benchmark purposes datasets for benchmarkings must have the critical property that the sensitive and non-sensitive documents are similar in structure and thus harder to distinguish than sensitive datasets where artificial sensitive documents have been inserted into an existing open dataset and where the sensitive documents are completely different in form than the non-sensitive documents.

To address these shortcomings we create 8 benchmark tasks for sensitive information detection based on 2 corpora of documents. In total we release publicly over $750,000$ new, labeled, processed sentences for sensitive detection training and benchmarking. Each of our definition of sensitive information types is given as a textual definition and also through a training set. To ensure that our sensitive information definitions are not overly simplistic we based our labels on labels provided by human annotators.

Our first 4 benchmark tasks are based on the Enron dataset which consists of internal and external communication from a number of Enron employees released as part of an investigation of the practices in the now defunct company Enron around 2002 [58]. The Enron dataset has been released nearly completely unaltered and as such is one of the few real-world examples of a dataset containing several different kinds of sensitive information. Furthermore while the Enron dataset has been utilized by other authors [16, 42, 84] as example of a real-world dataset, we recognize that to find non-simplistic sensitive information we need true labels of sensitive information and to obtain this we need human annotators. Here we propose to use labels generated in the Legal track of Text Retrieval Conference (TREC) around 2010, we argue that some of the different topics labeled by human annotators can be viewed as types of sensitive information and as such together with Enron text documents an unique large scale real-world dataset with several types of sensitive information labeled by human annotators.

In [76], also included in Chapter 10 and in detail discussed in Section 5.2, we propose our second set of benchmark tasks. Similar to the labeled Enron dataset our second dataset is a real-world dataset with communication and documents from a large corporation, Monsanto. These documents was released to the public in 2017 as part of the ongoing trial of Monsanto (see Section 5.2). The our processing, labeling and final publication of re-worked Monsanto dataset is motivated by the need to have more than one dataset with real-world sensitive information for evaluation and benchmarking of various sensitive information detection approaches. Furthermore we also need to validate that biases and dataset specific structures in the Enron dataset is not captured by sensitive information detection approaches, e.g. that our approaches generalize to other types of sensitive information. We also note that the Enron dataset is from around 2002 and thus more than 15 years old and the need for an updated dataset with documents representative of communication today is needed.

The Monsanto documents released in the trial are unlabeled. Therefore we propose to use labels generated by lawyers at Baum, Hedlund, Aristei & Goldman[7] as labels for different types of sensitive information. There are 4 different types of sensitive information labeled. For each type we both study lawyer document labels as sentence label in a *silver* dataset with weaker (more noisy) labels and also a *golden* dataset with true sentence level labels from a group of annotators. For each sensitive information type we release both the silver and golden datasets, in total 8 preprocessed datasets with constituency parse trees for each labeled sentence.

Previous work in detection of sensitive information in textual data rely on Bayesian statistics based on co-occurrence counting of words in documents from various domains [16] [38, 84, 91] similarly to the classical and seminal techniques for classification by topic [14] or sentiment analysis [77]. We refer to method based on co-occurrence counting as *keyword-based* approaches because they count occurrences of *keywords*. We formalize this for the sensitive detection domain in Chapter 2 and discuss the keyword-based approaches in Chapter 3. In general keyword-based models for text classification tend to be brittle and very sensitive to small changes in training data [94].

In particular, intuitively if we have a sentence with sensitive information, e.g. a sensitive sentence, and the sentence is rewritten, i.e. *paraphrased*, such that the sentence still carries the same information, then, clearly, the sentence is still sensitive. This intuition prompts us to look into current best approaches for paraphrase detection and create new approaches which combine the ability to detect sensitive information with the ability to detect paraphrased information. This is the topic of Chapter 4.

So-called Recursive Neural Networks (RecNN) are among the current best approaches for paraphrase detection [88]. Such models can learn from text and structure at the same time and this appears to be beneficial for problems where natural language structure is important such as for instance sentiment analysis [90] or paraphrase detection [88].

We create RecNN models for sensitive information detection which take text and structure representations and detect sensitive information. We empirically demonstrate that such RecNN approaches consistently outperform keyword based approaches.

Neural network models and recursive neural network models have high costs in terms of training times. This is in contrast with keyword based approaches which often can be implemented in one pass counting over the data. To ensure that we can train full RecNN models on a smaller computational budget we create a new efficient training strategy for neural networks and empirically demonstrate that this strategy reduces training times drastically without sacrificing performance of the model. This is discussed further in Chapter 9.

Detection of sensitive information constitutes a part of so-called *Data Leak Prevention (DLP)*. In this work we present new approaches to sensitive information detection for unstructured documents, which is also sometimes referred to as *Data Leak Detection (DLD)*. The approaches presented in this work are inspired by current deep learning NLP methods which automatic learn semantic and syntactic features through recursive neural networks from the dataset.

## 1.1 Research Questions

In this dissertation we address the important problem of complex sensitive information detection using machine learning for automatically detection. The work done here and in the publications which form part of this dissertation address the following research questions

**RQ1** How can we create state-of-the-art ML models, such as recursive neural networks, for complex sensitive information detection?

**RQ2** How can we efficiently train our models?

**RQ3** How do we benchmark different sensitive information detection approaches?

These questions form the boundaries for our investigations, developments and experiments in the following chapters. In Section 6.2 we relate our contributions to the research questions posed above.

## 1.2 Thesis Structure

Part I is outlined as follows; Above we introduced and motivated the problem domain. In Section 1.1 we outline the research question we aim to answer with this work. In Chapter 2 we discuss the drawback of simplistic sensitive information detection which do not properly address the context of potential leaks. We define the critical property of sensitivity in *context* which turns out to guide our work on advanced methods for detecting sensitive information. In Chapter 3 we discuss previous work on sensitive information detection, we characterize approaches which do not consider context as *keyword-based* approaches. We group previous work into 3 main groups of approaches; $n$-gram based (Section 3.1), Inference rule based (Section 3.2) and Pointwise mutual information based (Section 3.3). We are able to show that all 3 groups are in fact keyword-based. Furthermore we show this implies that previous state-of-the-art are blind to certain kinds of sensitive information. In Chapter 4 we discuss similarities between paraphrase detection and sensitive information detection. We make slight modifications to the problem formulation of paraphrase detection such that we transform the paraphrase problem into a new variant. We show how we can adapt approaches for paraphrase detection into approaches for the new paraphrase detection variant, and, more importantly for our purposes, how to adapt the approaches to sensitive information detection. This we do in Section 4.1 where we introduce recursive neural networks for sensitive information detection.

In Chapter 4.2 we address the need for public available benchmark dataset and evaluations of sensitive information detection approaches. We present two new datasets with human annotated documents and part of our contribution is the release of more than $750,000$ labeled sentences and their constituency parse trees as easy available training data for recursive models and as evaluation of sensitive information detection approaches. In Section 5.1 we present a dataset based of the Enron dataset and in Section 5.2 we present our second dataset which is based on recently released documents from the Monsanto trial.

We end Part I with an outlook in Chapter 6 where we discuss possible future research directions in Section 6.1 and finally conclude with a discussion of our contributions in Section 6.2.
An overview of notation used in this work can be found in Table 1.1.
In Part II we include published work, Chapter 7 - Chapter 10:

**Chapter 7** Jan Neerbek, Ira Assent and Peter Dolog. Detecting Complex Sensitive Information via Phrase Structure in Recursive Neural Networks. *Advances in Knowledge Discovery and Data Mining*, PAKDD '18. [74].

**Chapter 8** Jan Neerbek, Ira Assent and Peter Dolog. TABOO: Detecting unstructured sensitive information using recursive neural networks. In *Proceedings of the 33rd International Conference on Data Engineering*, ICDE '17. [73].

**Chapter 9** Jan Neerbek, Peter Dolog and Ira Assent. Selective Training: A Strategy for Fast Backpropagation on Sentence Embeddings. *Advances in Knowledge Discovery and Data Mining*, PAKDD '19. [75].

**Chapter 10** Jan Neerbek and Morten Eskildsen and Peter Dolog and Ira Assent. A real-world data resource of complex sensitive sentences based on documents from the Monsanto trial. *Proceedings of The 12th Language Resources and Evaluation Conference*, LREC '20 [76].

In accordance with GSST regulations, parts of this thesis was also used in the progress report for the qualifying examination. Furthermore I hereby assert that I, the author, independently and in my own words have formulated Part I of this dissertation. In all included material from previously published work, Part II, (Chapter 7 - Chapter 10), I have done the main work both in terms of development of ideas, research and experiments and also in terms of actual writing of the articles. This is also certified by my co-authors in co-author statements which have been obtained and which were submitted together with this dissertation as required.

| **Notation** | **Meaning** |
|---|---|
| $D$ | A distribution or finite dataset $D$ of documents |
| $V_D$ | The vocabulary of words encountered in $D$ |
| $\mathbf{v}$ | A vector with components $\mathbf{v}[i]$ |
| $\mathbf{M}$ | A matrix with components $\mathbf{M}[i,j]$ |
| $w$ | A word, such that $w \in V_D$ |
| $(w_1, w_2, \ldots, w_n)$ | A sequence (vector) of words |
| $d$ | A document, i.e. $d = (w_1, w_2, \ldots, w_{n_d})$, where $n_d$ is the length of the document |
| $\{w_1, w_2, \ldots, w_n\}$ | A set of words, e.g. $i \in [1; n] : w_i \in V_D$ |
| $\mathbb{1}[.]$ | The indicator function which is 1 if the argument is true and 0 otherwise |
| $\mathbb{R}$ | The real numbers |
| $\mathcal{L}$ | A labeling function $\mathcal{L} : D \rightarrow \{0, 1\}$. See also Definition 1 |
| $\vert.\vert$ | Size of argument. I.e. length of sequence or number of elements in set |
| $\Vert\mathbf{v}\Vert$ | Euclidean norm of vector $\mathbf{v}$, i.e. $\Vert\mathbf{v}\Vert = \sqrt{\mathbf{v}^T\mathbf{v}}$ |
| $sen(w)$ | A score for how sensitive word (or words) $w$ is (are) |
| $tp, fp, tn, fn$ | True Positive, False Positive, True Negative, False Negative. For classification into *positive* (here: sensitive) or *negative* (here: non-sensitive). True if classification is correct, False otherwise |
| $Acc$ | Accuracy, $Acc = \frac{tp+tn}{tp+fp+tn+fn}$ |
| recall | recall $= \frac{tp}{tp+fn}$ |
| precision | precision $= \frac{tp}{tp+fp}$ |
| $F1$ | $F1 = 2\frac{\text{precision}\cdot\text{recall}}{\text{precision}+\text{recall}} = \frac{2tp}{2tp+fn+fp}$ [26] |

Table 1.1: Notation used in Part I. Note in each of the previous published articles, Chapter 7 - Chapter 9, notation is slightly adapted to a particular publishing venue and thus not completely captured in this table.

# Chapter 2

# Sensitive Information

The definition of *sensitive* is highly domain specific. In healthcare and government institutions there is a lot of focus on no leaking of Personally identifiable information (PII) [60] whereas in organizations, both private and public, there is (also) focus on not leaking informational content such as documents containing information on, for instance, inventions or strategic plans. Whereas in text classification classifying might happen according to one-word class definitions, e.g. a *Sports* class [14], in this work we focus on sensitive information which often is a very complex construct and not easily captured by a series of keywords. E.g. where the definition of what constitutes sensitivity is known implicitly to domain experts, but a formal definition is not given or may even be impossible to obtain. Thus we use a data-driven approach where we focus on sensitive content identified by human annotators, making use of state-of-the-art ML research. Given a dataset, data and actual labels define what is sensitive, and we refer to it as the *sensitive information type* or as the type of sensitive information.

When we consider sensitive information we always do this in context of some *domain*. Examples of domains includes a domain of a single company, or a domain of a governmental organization. A domain can also be a certain type of information such as personal information, where an example is information about a person's health and sicknesses.

We focus on sensitive content in *textual units*. Following [16], we consider a textual unit to be the relevant unit of context. Depending on the task and domain this could be a document, paragraph, sentence, phrases or words. In this work we consider sensitive content in the scope of *sentences*. Working on the sentence level has value for potential redacting purposes as it allows us to remove just the sensitive sentences and leave large parts of an otherwise non-sensitive document untouched. This allows users to browse non-sensitive parts of a document released based on, say, the Freedom of Information act discussed in the previous section.

## 2.1 Definitions

We assume explicit access to a *distribution* of sentences $S$. E.g. for a domain we have a set of sentences, where each sentence belongs to some distribution.

$$s \in S$$

and consists of a sequence of words

$$s = (w_1, w_2, \dots, w_{n_s}), \quad \forall i : w_i \in V_S$$

where $n_s$ is the number of words in document $s$ and $V_S$ is the vocabulary for the distribution $S$.

The sentences are generated by some process in the domain which is generally not known to us. The size of the distribution $S$ can be theoretically unbounded if we can obtain new samples from the domain process and in principle have access to as many samples as we wish, or the distribution can in fact be a dataset of sentences of some finite size. The latter case is typical when a released set of sentences with or without labels are available. Thus we slightly abuse notation and let $S$ denote both an infinite sized probabilistic distribution of sentences and also a finite sized dataset of sentences.

Two examples of such finite sized datasets are the Enron and Monsanto sets. with a definition of sensitive information given as a representative sample of sentences exhibiting some sensitive content. For example, a domain could be *Monsanto* and $S$ is the released dataset as discussed in Section 5.2.

Given a domain and a distribution of sentences $S$, we further need a definition of what constitutes sensitive information. A number of different definitions have been proposed, e.g. see Chapter 4.2 for concrete domain definitions based on human annotators and also Chapter 3 for previous work.

For now we abstract from any particular sensitive information definition. In general we assume access to a labeling function $\mathcal{L}$, i.e. for a given dataset or distribution $S$ our labeling function is a function from sentences in $S$ to the binary alphabet $\{0, 1\}$

$$\mathcal{L} : S \to \{0, 1\}. \tag{2.1}$$

This labeling function defines, for a particular domain, which sentences are sensitive.

**Definition 1** ***Sensitive Definition*** *Given a dataset or distribution of sentences $S$ and a labeling function $\mathcal{L}$. We say the labeling function* provides *a definition of sensitive content over the sentences $S$ such that*

$\mathcal{L}(s) = 1$ *implies* ***some*** *parts of the sentence are sensitive*
$\mathcal{L}(s) = 0$ *implies* ***none*** *parts of the sentence are sensitive*

For the sensitive information detection problem, one goal for approximating, machine learning models becomes having the model predict as well as possible the labeling function as given by Definition 1.

Whereas other sensitive information detection approaches (see Chapter 3) assume that the sensitive information is defined by a simple, known, generative function, e.g. seed words, the labeling function defined above is not necessarily generative. That is, we do not assume a known generative model for sensitive information. Instead we rely on labeled samples of documents and proceed to models for finding sensitive information based solely on this labeling function over a given domain. This is a crucial property compared to approaches which assume seed-words or another known generative process for labels, see also Chapter 3 and the discussion of the assumptions made in such approaches.

Also note from Definition 1 that the problem is asymmetric in how the labels are awarded. If $\mathcal{L}(s) = 0$ then the sentence $s$ is known to be completely safe, with no sensitive information. The situation for a sensitive sentence $\mathcal{L}(s) = 1$ is different. In a sensitive sentence *some* sensitive information is present. That is, large portions of the sentence might be, and in practice often are, non-sensitive.

Next we define sensitivity as a function of the *context*. The context is the textual unit in which a word appears. In this work we consider the (sensitivity) context of a word to be the sentence in which it appears, but our approach can be generalized to considering the document in which the word appears to be context.

In the following we use the sentence as context. We introduce two core notions: context-less and context-based sensitive information. Context-less sensitive information types are then characterized by being *invariant* to the context (here: sentence) in which they appear. In a context-less approach, a word can be assigned a fixed sensitivity score independently of the sentence.

Please note that depending on the model, other information, such as groups of neighboring words may be part of the sensitivity score definition. However, if the sensitivity score definition is the same for multiple contexts then we say that the sensitive score definition (and the sensitive information type) is context-less. Otherwise it is context-based.

As we demonstrate in this work for complex sensitive information considering only context-less sensitivity is not always sufficient. Some sensitive information is undetected by such approaches. With this intuition we define context-based sensitive information as *conditional sensitivity* of words given a context; Given a sentence $s$ we denote the conditional sensitivity of a particular word (given $s$) as

$$sen(w|s)$$

I.e. the sensitivity score of $w$ in context of $s$. If another sentence $s'$ consists of the same sequence of words, i.e. $s' = s$ then we require $sen(w|s) = sen(w|s')$.

We capture this property in the following definition.

We define *context-less sensitive information* as

**Definition 2** ***context-less*** *Given a set of conditional sensitivities*

$$\Delta = \{sen(w_0, w_1, \ldots, w_n|s) \mid \forall i : w_i \in V_S \textit{ and } s \in S\}$$

*then, if for all n-grams* $w_0, w_1, \ldots, w_n$ *and pair of sentences* $s, s'$ *we have*

$$\forall (w_0, w_1, \ldots, w_n), s, s' \qquad sen(w_0, w_1, \ldots, w_n|s) = sen(w_0, w_1, \ldots, w_n|s')$$

*we say that the sensitivity defined by* $\Delta$ *defines* context-less sensitive information.

For *context-based sensitive information* we similarly define

**Definition 3** ***context-based*** *Given a set of conditional sensitivities*

$$\Delta = \{sen(w_0, w_1, \ldots, w_n|s) \mid \forall i : w_i \in V_S \textit{ and } s \in S\}$$

*then if there exists a n-gram* $w_0, w_1, \ldots, w_n$ *and a pair of sentences* $s, s'$ *such that*

$$\exists (w_0, w_1, \ldots, w_n), s, s' \qquad sen(w_0, w_1, \ldots, w_n|s) \neq sen(w_0, w_1, \ldots, w_n|s')$$

*we say that the sensitivity defined by* $\Delta$ *defines* context-based sensitive information.

In [74] (Chapter 7) we previously published a similar definition. I.e. Definition 4 in Section 7.2.

The definitions, Definition 2 and Definition 3 again reflect an asymmetry similar to the asymmetry of what constitutes a sensitive sentence.

Intuitively the two definitions reflect the difference between having a fixed context of consideration ("context-less") and having a fluid context of consideration ("context-based"). The widely used $n$-grams (see Section 3.1) often provide strong results for natural language problems. In a sense $n$-gram based approaches are in-between context-less and context-based. E.g. on one hand, given a fixed $n$ we may learn an $n$-gram which can be both sensitive and non-sensitive, and would require a larger context to correctly detect sensitive content. On the other hand, most sensitive content can be captured by an $n$-gram approach if just $n$ is set high enough.

In our published work, Part II, we contrast previous work, including $n$-grams, with more complex methods. To be aligned with our published work, $n$-grams are considered context-less approaches in the following.

## 2.2 Examples of sensitive content

To make the types of sensitive sentences which we encounter in our datasets more concrete we here add a few illustrative examples of sensitive vs non-sensitive sentences.

Example from the *GHOST* type of sensitive information from the Monsanto dataset, see Chapter 4.2. In the ghostwriting information type, Monsanto had people write articles and columns and discussions of other people listed as authors.

```
But I suspect that is wishful thinking Are you interested
in writing a column on this topic?
```

We note that while the human annotators have marked this sentence as being sensitive (with regards to *GHOST*) the sentence itself does not discuss ghost-writing or who gets credit so the sensitivity is not attached to any one word but rather to the intent of the communication.

Two more examples, both from the Monsanto *REGUL* dataset are the following:

```
Regulatory Affairs has shared these recent publications
with IARC and is continuing to share directly with
panelists IARC has a history of questionable and
politically charged rulings on the carcinogenic
properties of products such as cell phones, coffee and
caffeine.
```

and, (also from *REGUL* dataset)

```
rats Study 6:  MONSANTO Ill GLOBAL REGULATORY Monsanto
Company Confidential AND GOVERNMENT AFFAIRS BMD Modeling
of Pancreatic Islet Tumor Data in Male Rats Approach
```

The first *REGUL* example is actually sensitive according to the sensitive information type (Monsanto's dealing with regulatory bodies) whereas the second example, even though it contains the word "REGULATORY" is labeled non-sensitive. From the data we see that the second example is actual experiments reporting internally in Monsanto whereas the first example is discussion on reporting to regulatory bodies (IARC) from internally in Monsanto that both make use of related words, but have different labels with respect to the sensitive information type.

# Chapter 3

# Keyword-based Approaches

A common approach to detecting sensitive information in texts is based on word counting, i.e. using statistics on words for prediction. In this chapter we look at the 3 common families of approaches; $n$-gram, inference rules and pointwise mutual information based approaches. We focus here on their usage in sensitive information detection literature. We show how they are based on word counting and we show these types of approaches all share some common properties. Specifically they all have a fixed context size and in sensitive information detection the three approaches are used with a sensitive information definition which is by nature fixed for the corpus on which the approaches are applied on. Given the operator supplied fixed context size and the static definition of sensitive we propose that $n$-gram, inference rules and pointwise mutual information based approaches to finding sensitive information all only can find context-less sensitive content, as defined in Definition 2. We show that all three families of methods essentially consider the same conditional probabilities to this end. This motivates finding new approaches which does consider context when assigning sensitivity scores, which we do in Chapter 4. In our published works we contrast these 3 approaches with more complex approaches and to be aligned with publications in Part II we consider these $n$-gram, inference rules and pointwise mutual information approaches to be context-less in this chapter.

A defining common characteristic of these approaches, as we show in this chapter, is that they use words as the necessary statistic for assigning sensitivity score to texts. Approaches which are completely without context only consider a word at a time. This implies that sensitivity scores are assigned to individual words

$$sen(w)$$

I.e. context-less as defined in Definition 2.

For approaches which use a fixed sized context, this can be extended to the case where we are working with sequences of words, $n$-grams

$$sen(w_1, w_2, \ldots, w_n) \tag{3.1}$$

which is the sensitivity of the sequence (ordered set) of words $w_1$, $w_2$, up to $w_n$. We refer to approaches which only considers statistics on the form Equation (3.1) as being *keywords-based* because of the fact that these approaches only considers the probability of (key-)words. By definition keywords-based approaches are context-less in the sense of Definition 2.

Assume we are given a set of documents $D$, with vocabulary $V_D$ and a labeling function $\mathcal{L}(d)$ over all documents $d \in D$. We can obtain a large set of describing probabilities by counting over words, e.g. we denote the count of the sequence ($n$-gram) $w_1, w_2, \ldots, w_n$ occurring in $D$ as

$$C(w_1 \wedge w_2 \wedge \ldots \wedge w_n) = \sum_{d \in D} \mathbb{1}[(w_1 \wedge w_2 \wedge \ldots \wedge w_n) \in d]$$

where $\mathbb{1}[\ldots]$ is the indicator function which has value 1 if the parameter is true and value 0 otherwise. The use of the logical *and* ($\wedge$) operator represents the fact that the words occur together.

Similarly we can count the number of occurrences of $w_1, w_2, \ldots, w_n$ in documents that contain sensitive content, i.e. where $\mathcal{L}(d) = 1$

$$C(w_1 \wedge w_2 \wedge \ldots \wedge w_n, \mathcal{L}(d) = 1) = \sum_{d \in D} \mathbb{1}[(w_1 \wedge w_2 \wedge \ldots \wedge w_n) \in d \text{ and } \mathcal{L}(d) = 1].$$

Given these counts we form estimators for the probabilities

$$\Pr(w_1 \wedge w_2 \wedge \ldots \wedge w_n) = \frac{C(w_1 \wedge w_2 \wedge \ldots \wedge w_n)}{|D|}$$
$$\Pr(w_1 \wedge w_2 \wedge \ldots \wedge w_n, \mathcal{L}(d) = 1) = \frac{C(w_1 \wedge w_2 \wedge \ldots \wedge w_n, \mathcal{L}(d) = 1)}{|D|}.$$

We can apply Bayes Law and obtain an estimate of the conditional probability of a document $d$ containing sensitive content given the observation that $w_1, w_2, \ldots, w_n$ occurred in $d$

$$\Pr(\mathcal{L}(d) = 1 | w_1 \wedge w_2 \wedge \ldots \wedge w_n) = \frac{C(w_1 \wedge w_2 \wedge \ldots \wedge w_n, \mathcal{L} = 1)}{C(w_1 \wedge w_2 \wedge \ldots \wedge w_n)}. \quad (3.2)$$

Thus if we only consider one $n$-gram at a time in order to make a prediction of whether the text under consideration is sensitive or not, then we become context-less beyond our chosen size $n$ context. We show in this chapter that the current approaches in sensitive information detection described here all consider sensitivity one $n$-gram at a time and thus are context-less in the sense of Definition 2. Note, however that considering one $n$-gram at a time is a choice. Another choice could be to look at the distribution of $n$-grams and their sensitivity scores in a text. Doing this could would make the new approach being less context-less, e.g. it would potentially be possible to detect sensitivity scores over longer sequences of words than the original $n$-gram. In previous work in sensitive information detection no such extension has been investigated and as such the approaches considered here only consider statistics with fixed size context similar to Equation (3.2).

Such probabilities are assigned regardless of the document and the context in which the word sequence occurs. Thus the probability can be viewed as a context-less sensitivity score for the word sequence and it follows that any approach which estimates sensitivity using estimators on this form is *keyword-based* and only captures *context-less* sensitive information in the sense of Definition 2.

## 3.1 $n$-gram based

In the $n$-gram model we are concerned with $n$-grams that co-occur together, e.g. $n$ words that co-occur together in document. In some sense $n$-grams are the prototypical example of a keyword-based approach. Often a $n$ value of 2, e.g. a bi-gram model, yields surprisingly strong results, e.g. [38]. Whereas higher $n$ tends to perform less well as our models do not generalize as well and become brittle with respect to small changes in data [94].

In the $n$-gram language model we assume we can assign a probability $\Pr(.)$ to word sequences, i.e., $\Pr(w_1, w_2, \ldots, w_n)$, where $w_i \in V_D$. The motivation for looking at the $n$-gram probabilities for sensitive information detection is expectation that the distribution of word sequences in sensitive documents is different from that in non-sensitive documents. I.e. an application of the Distributional Hypothesis by Harris [41], which can also be informally stated as

*words (meaning) are similar if they appear in similar contexts* [34].

Thus we consider the document as a series of probabilistic events (i.e. words). The probability of document $d$ occurring is then the joint probability of the document

$$\Pr(d) = \Pr(w_1, w_2, \ldots, w_{n_d}).$$

By applying the chain rule of probability [49] we can write this in the following form

$$\Pr(w_1, w_2, \ldots, w_{n_d}) = \prod_{i=1}^{n_d} \Pr(w_i | w_1 \ldots w_{i-1}).$$

In the $n$-gram model we approximate the terms in the above expression by making the assumption that every word only depends on the previous $n-1$ words. This is referred to as the Markov $n$-gram independence assumption [78]

$$\Pr(w_1, w_2, \ldots, w_{n_d}) = \prod_{i=1}^{n_d} \Pr(w_i | w_{i-n+1} \ldots w_{i-1}) \qquad \text{(Markov assumption)}.$$

The estimation of the Markovian probabilities is usually done by counting number of occurrences of $(w_1, w_2, \ldots, w_n)$ in documents over the dataset and then normalizing (i.e., dividing) by the size of the dataset, as we saw in the introduction to this chapter. Consider a bi-gram example, we estimate the probability of a sequence $(w_i, w_j)$ by counting the number of occurrences of the sequence and dividing by the number of occurrences of any (in this example, bi-gram) sequence beginning with $w_i$. Thus we obtain (first for the bi-gram example)

$$\Pr(w_j|w_i) = \frac{C(w_i \wedge w_j)}{\sum_{w_k} C(w_i \wedge w_k)}.$$

Generally we obtain for word $w_j$ and any length $n > 0$ word sequence $w_1, w_2, \ldots, w_n$

$$\Pr(w_j|w_1, w_2, \ldots, w_n) = \frac{C(w_1 \wedge w_2 \wedge \ldots \wedge w_n \wedge w_j)}{\sum_{w_k} C(w_1 \wedge w_2 \wedge \ldots \wedge w_n \wedge w_k)}. \tag{3.3}$$

Approaches based on $n$-grams with lower order of $n$, has been previously applied for sensitive information detection. Hart et al. [42] uses a traditional bag-of-words classification approach, i.e., 2-grams (word frequencies exceeding a threshold frequency) to detect sensitive information. The evaluation finds that SVMs outperform Naïve Bayesian and Rocchio score on pairwise document set evaluation. In [47] private information is detected in Twitter feeds using broad topical (e.g. Latent Dirichlet Allocation) and $n$-gram features.

Approaches for text sanitization where we not only detect sensitive content but also apply strategies other than removing to retain *utility* of the resulting text also can be based on $n$-grams such as the uni-gram approach $t$-pat [48], but these approaches are similarly sensitive to more context. Indeed the authors in [5] show that doubling the context of $t$-pat from uni-gram to bi-grams result in better detection and sanitation, however they do not investigate for higher values of $n$.

Using $n$-gram approaches assign a fixed context size to every probability considered. For the approaches described above these probabilities are considered independently of each other. As argued in the introduction to this chapter we conclude that such simple use of $n$-gram approaches do not capture the contextual sensitive information and can ultimately only assign a fixed, context-less sensitivity value to each $n$-gram and thus more complex sensitive information is not addressed.

## 3.2 Inference rule based

Inference rule models for sensitive information detection are inspired by association rule mining where the goal is to find frequently occurring itemsets [2]. In the sensitive information detection setting we consider frequently co-occurring words as elements in frequently itemsets. I.e. similar to $n$-grams we, in inference rule based sensitive information detection, base our statistics on co-occurrence counts. A inference rule may take the form

$$A \rightarrow B$$

which is to be interpreted as *the observation of condition* $A$ *(left-hand-side) implies the consequence* $B$ *(right-hand-side).* In association rule mining $A$ is a frequent itemset and $B$ is also a frequent itemset. For our purposes $A$ is some expression over words occurring in our data and $B \in \{0, 1\}$, i.e. $B$ is a prediction of either "sensitive" ($B = 1$) or "non-sensitive" ($B = 0$) . For each rule we associate a probability known as the *confidence* of the rule. First we define the structure of the condition $A$ and then we define the estimated confidence of the rules using $A$ as a condition.

In general the condition can be a function of the words occurring. From [16] we extract 3 different ways to combine observed word occurrences

$$w_1 \wedge w_2 \tag{3.4}$$

is true for document $d$ if sequence $(w_1, w_2)$ occurs in $d$. Similar a logic formula

$$w_1 \vee w_2 \tag{3.5}$$

is true for a document $d$ if any of $(w_1)$ *or* $(w_2)$ occur in $d$. Finally we define the logic formula

$$\neg w_1 \tag{3.6}$$

as true if the word $(w_1)$ does *not* occur in the document in $d$.

Using the given three basic constructive logic formulas we can generate more complex Boolean formulas. I.e. by substituting the words in the formulas with previous derived formulas we may obtain complex formulas dependent on several words; $A = f(w_1, w_2, \ldots, w_k)$. Each formula $A$ has an associated weight, referred to as the *support*

$$\text{Supp}(A) = \frac{C(A)}{|D|} \tag{3.7}$$

I.e. $\text{Supp}(A)$ is the fraction of the number of occurrences of $A$ in D. Furthermore each rule $A \rightarrow B$ (where $A$ now is the condition) has an associated *confidence*,

$$\text{Conf}(A \rightarrow B) = \frac{\text{Supp}(A \wedge B)}{\text{Supp}(A)} = \frac{C(A \wedge B)}{C(A)} \tag{3.8}$$

Intuitively confidence tells us given $A$ how certain are we of $B$, whereas support tell us how often $A$ occurs in the dataset, e.g. how important is the *set* of rules that have $A$ as a condition. In association rule mining we are typically interested in rules, $A \rightarrow B$ where the condition $A$ occurs often (e.g. has high support) and where the rule has high confidence (e.g. if we encounter $A$ then there is high probability for also encountering $B$). However as argued in [16] in sensitive information detection we are interested also in rules with where the condition has low support, because we wish to detect all leaks, not just the most frequent ones.

However note that we are not interested in rules with low *confidence* since low confidence implies a weak connection to the sensitive information.

The earliest work in the domain, [16], defines sensitive information as words frequently occurring together with sensitive seed words. Such sensitive information is learned using word-to-word inference rules. In the first part of the empirical evaluation, a particular topic ("HIV" in this case) is assumed to be sensitive, and Wikipedia documents are screened. In the second part, the topic is set to the "University of Wharton" on the Enron dataset. These datasets are thus well suited to study the identification of clear-cut sensitive topics as a classification task where co-occurrence of words with seed words can be reasonably assumed. In our work, however, we target complex sensitive topics that are not necessarily captured by co-occurrence with seed words.

Grechanik et al. [38] extends the inference rule approach to sensitive information in software artifacts, such as comments or function and variable names in the software source code. The goal is to avoid actual past incidents where software code was inadvertently shared without removing comments about (opinions of) business partners or other business secrets. Such leaks can obviously be damaging to company business. In their work, word-to-word inference rules are applied to keyword based sensitive topics that are to be redacted from source code files. For evaluation, a set of keywords are chosen by the authors as being sensitive. As before, this means that complex sensitive topics are not captured. The sensitive information extracted is redacted using synonyms obtained from publicly available word ontologies. The performance of the approach is measured in terms of increasing entropy (based on word count ratios) when comparing the original data and the redacted versions.

Cumby and Ghani [23, 24] uses inference control to detect and sanitize sensitive information. They develop the notion of *$k$-confusability* where they consider an adversarial classifier $H$ which generates a total ordering of possible sensitive classes that a document can be in and the document is $k$-confusable if it's rank in the ordering is at least $k$. The authors uses uni-gram as features and thus assign context-less sensitivity scores to all uni-grams. Several experiments are conducted to remove sensitive information under the assumption of context-less sensitive information and while keeping various measures of utility high.

Geng et al. [33] detect quasi-identifiers (entities) (QIE) and sensitive entities (SE). Where examples of QIE might include a persons age, address or name and examples SE might include private information such as diseases, sexual orientations and so on. The authors apply inference rules to discover the sets of SE. The entities, QIE, are obtained as Named Entities from the text and probabilities are obtained though counting of search results with the QIE subsets. The extraction of both SE and QIE sets are based on an assumption of context-less sensitivity scores on uni-grams. High accuracy experimental results are obtained for PII and PHI from user interaction on a online health discussion forum.

While it is in principle possible to use any possible logic formulas, in the literature only single word to sensitive prediction inference rules has been reported [16, 23, 24, 33, 38]. Such approaches limit the context to a fixed size and therefore such approaches does not as such classify a document as being sensitive, but rather generates a set of context-less sensitivity scored words or $n$-grams.

## 3.3 Pointwise mutual information

Pointwise mutual information (PMI) is a statistical measure which can be used for inferring relations between events. We follow [65]and define PMI between two words $w_1, w_2$ as

$$PMI(w_1 \wedge w_2) = \log \frac{\Pr(w_1 \wedge w_2)}{\Pr(w_1)\Pr(w_1)}$$

where log is base-2 log. This definition can be extended to two word sequences ($n$-grams) by replacing $w_1$ and $w_2$ with the $n$-grams in question. In [84] the authors introduce pointwise mutual information as a measure for when words are to be considered sensitive. The authors consider PMI with respect to the sensitive documents, i.e.

$$PMI(w_1 \wedge w_2 \wedge \ldots \wedge w_n, \mathcal{L} = 1) = \log \frac{\Pr(w_1 \wedge w_2 \wedge \ldots \wedge w_n, \mathcal{L} = 1)}{\Pr(w_1 \wedge w_2 \wedge \ldots \wedge w_n)\Pr(\mathcal{L} = 1)}$$

where $(w_1 \wedge w_2 \wedge \ldots \wedge w_n)$ is a sequence of words which we wish to determine if is sensitive or not. Note that the authors in [84] work with seed sets of sensitive words and thus their count of sensitive occurrences really counts documents with sensitive words. I.e., the event $\mathcal{L} = 1$ is the same as the event that we see particular seed words.

Using Bayes' theorem we may rewrite $PMI$ as (See [65], Section 5.4)

$$PMI(w_1 \wedge w_2 \wedge \ldots \wedge w_n, \mathcal{L} = 1) = \log \frac{\Pr(\mathcal{L} = 1 | w_1 \wedge w_2 \wedge \ldots \wedge w_n)}{\Pr(\mathcal{L} = 1)} \quad (3.9)$$

Note that since $\Pr(\mathcal{L} = 1)$ is constant for a particular dataset $D$ and labeling function $\mathcal{L}$ and because the log function is monotonically increasing, we find that the ordering given by $PMI$ score is the same as the ordering given by $n$-grams or inference rules for same $(w_1, w_2, \ldots, w_n)$ as given by Equation (3.2).

When working with $PMI$ we wish to set a threshold for $PMI$ which provides us we good predictors of potential sensitive content [84]. Using this threshold we mark all single words $w$ with high $PMI(w, \mathcal{L} = 1)$ to be sensitive. One particular interesting measure to be utilized as threshold is the *Information Content* $IC(\mathcal{L} = 1)$.

$$IC(\mathcal{L} = 1) = -\log \Pr(\mathcal{L} = 1) = -\log \frac{C(\mathcal{L} = 1)}{|D|}$$

I.e. the probability $\Pr(\mathcal{L} = 1)$ is estimated as the fraction of sensitive documents over the size of the dataset. Thus for any word(s) $w$ if we have

$$PMI(w, \mathcal{L} = 1) = \log \frac{\Pr(\mathcal{L} = 1|w)}{\Pr(\mathcal{L} = 1)} \geq IC(\mathcal{L} = 1)$$

then we consider $w$ to be sensitive also. Furthermore working in domains where the cost of leaks of sensitive information is particular high we can introduce a weighting scheme and thus increase the recall of sensitive information and the possible expense of precision

$$PMI(w, \mathcal{L} = 1) = \log \frac{\Pr(\mathcal{L} = 1|w)}{\Pr(\mathcal{L} = 1)} \geq \frac{1}{\alpha} IC(\mathcal{L} = 1)$$

For some weight $\alpha > 1$. The weight can be adjusted dynamically based on observed performance on a development dataset.

Sánchez and Batet [84] develop and based their approach *C-sanitized* on pointwise mutual information (PMI). This measure is used to define the relationship between sensitive words and other words in the document collection. PMI results in the same ordering as words as in the $n$-gram approach. In [84] the authors also propose using Information Content (IC), as threshold between sensitive and non-sensitive words. Authors introduce both information content $IC$ and the additional parameter $\alpha$ to allow for a stricter definition of sensitivity than using $IC$ alone. In [85] the authors extend the approach to encode the $\alpha$ parameter semantically and obtain a derived approach *(C, g(C))-sanitized*, where $g(C)$ provides a generalization for each sensitive keyword $c$, such that $g(c)$ is a semantic generalization of $c$, e.g. c="AIDS" and g(c)="chronic disorder". Instead of limiting $PMI$ to $IC(c)$ they propose to use $IC(g(c))$ which will detect more sensitive terms related to $c$. The authors argue that for domain users choosing a good generalization $g(c)$ is easier than choosing the $\alpha$ parameter for the optimization of the detection process. As the PMI approaches only consider word counts we conclude that these approaches only detect context-less sensitive information.

## 3.4 Other related work

[18] studies the problem of hiding sensitive information when generating vector representations. For example if a user is using a service on a mobile device

a vector representation of input is sent to the cloud for further processing. The approach studies the trade off between having high performance on the original task vs. low performance for a potential attacker trying to recover the sensitive information from an intermediate vector representation. In their evaluations, LSTMs and feedforward neural networks are used. However, only simple privacy types of sensitive information are considered such as age or gender. Also, the problem is given a state representation from a particular model how to prevent detection of private information. By contrast, we are interested in detecting complex sensitive information with high accuracy.

[97] considers the problem of hiding privacy related information, such as age, gender or race, by *rewriting* the texts. In this setup, a list of words and a privacy attribute are given. The hiding of privacy information is successful if an adversarial classifier has low probability of predicting the attribution. The authors consider several different cost functions to achieve a good performance that balances hiding the privacy attribute in question (making it difficult to predict the attribute from the rewritten text) and being able to recover most of the original text from the rewritten text. The authors consider the ability to recover (most of) the original text as a indicator that their rewritten text has not been too distorted to be useful in downstream tasks. The authors give as an example a job application where we wish to hide the gender of the applicant but where the application should still contain all other relevant information. Performance is evaluated using both automated quantitative metrics and human evaluations.

The approach of using automated language translation to hide privacy attributes is only a limited subset of sensitive information in general. Furthermore, their methodology is focused on hiding/redaction and not detection. It uses language translation approaches and can be seen as a form of paraphrase generation. In this work we focus on detection approaches.

[4] presents a stacked autoencoder approach which can take a context representation vector as additional input. The output is used on three different tasks where similarity of words and sentences are calculated. Note, sensitive here is used in the meaning of the autoencoder being sensitive to context, not as in the sense of sensitive information as discussed elsewhere in our work.

The approach developed in [4] uses an average of a small window of words in the original input as input to their stacked autoencoder. They do not consider recurrent or recursive structures which could be useful for carrying context information. As context input they use an average word embedding over a larger window of input. They then use dimensionality reduction to reduce the average embedding to a dense, low dimensionality context vector. Detection of sensitive or private information is not considered.

[1] introduces models for paraphrase detection in short and noisy texts. A deep component based model is created including a CNN-LSTM component, denoted SentMod, to build a sentence vector representation. Interestingly, the combination of CNN and LSTM can be seen as the CNN providing structure

to the LSTM in a local window ($n$-gram). Where we in our work provide structure in the form of constituency parse trees which are traversed recursively.

This approach performs well on paraphrase detection and is robust to noise in form and spelling of the sentence input. Unfortunately, it is not clear how to extend, in particular, the proposed Pair-wise Features to a sensitive information detection setting where we do not have matching pairs of sensitive and non-sensitive sentences.

[90] introduces a variant of recursive neural networks known as a *Recursive Neural Tensor Network*. It demonstrate stronger performance than AvgVec, which uses the average of input word embeddings as input. These findings motivate the investigation of recursive neural networks which we consider in this work (Chapter 4).

[68] evaluates the use of word embeddings for generating document representations for detecting sensitive information as defined by UK Freedom of Information laws[1]. They achieve strong results on a dataset of 3801 labeled documents. An interesting addition in their approach is the use of several different trained word embeddings as inputs, but where positional information (word order) is lost similar to AvgVec (see above). [68] includes a subset of frequent $n$-grams and POS $n$-grams, which adds a fixed context size. Thus their approach does not consider the full sentence context as in the RecNN based approaches we develop here.

[3] employ bag-of-words representation with TF-IDF weighting per word as paragraph representations. They develop a new dataset from U.S. diplomatic cables released by WikiLeaks. Interestingly, sensitivity annotation is found in the original documents on WikiLeaks. The authors build a dataset with three levels of sensitivity from the documents. The approach generates clusters of similar paragraphs and for each cluster a separate classifier is trained. Experiments are conducted with Naive Bayes and linear SVM classifiers. Their approach does not consider word order or semantic structure like parse trees as in our work. They use clusters to group similar paragraphs together, to help the context-less classifier make better predictions, rather than using clusters of similar sentences to select subsets for training as in our selective training Chapter 9

In his master thesis [39], Varun Gupta considers the problem of paraphrase generation, in particular with the goal of increasing training set size by generating paraphrases of existing training instance. This is a form of data augmentation and can help improve performance of downstream tasks, but does not consider paraphrase detection. In this work, we focus on detecting sensitive information and do not consider data augmentation.

[40] is a shorter article version of his master thesis [39]. Again only paraphrase generation as data augmentation is considered.

[1] http://www.legislation.gov.uk/ukpga/2000/36/contents

Other previous work on sensitive information redaction in the unstructured setting exists. These works focus on the redaction step, e.g. after a list of sensitive words have been found. As such these previous works do not approach the detection problem that we consider in this dissertation, however they still consider only context-less sensitive information as they redact *all* occurrences of a word regardless of content. We brief mention these works in this section.

Chakaravarthy et al. [15] develops the idea of $K$*-safety*, where they require that all subsets of words which might be sensitive occurs at least $K$ times in non-sensitive contexts. This provides a bound on how much an attacker could derive from a redacted document. The authors motivate $K$-safety with the example of removing “AIDS” and other private types of information from patient-doctor communications. Promising experiential results are provided for synthetic generated datasets however the authors assume that each type of sensitive information is given as a keyword and a list of words which may occur together with the keyword in sensitive documents. They do not consider the semantic relationship between words and would assign same sensitivity score to the occurrence of “HIV” or “not HIV” and thus complex types of sensitive information cannot be captured.

Chakaravarthy et al. is inspired from earlier work [86] where the concept of $k$*-anonymity* for authorship is defined. Here the list of sensitive words are provided as quasi-identifiers and this might include $n$-grams or global document statistics such as total number of words without it being further discussed how to obtain such a list.

## 3.5 Our Contributions

We define the *keyword-based* approach. Such keyword-based approaches can only detect *context-less* sensitive information. We conclude that all previous work for sensitive information detection considered here, based on $n$-grams, inference rules and Point-wise Mutual, considers fixed contexts of words and assign for each context (tuple of words) assign a single sensitivity score. These approaches such as they are used within sensitive information detection considers each of these sensitivity scores independently and thus these approaches rely on hard-coded sensitivity scores for context and are consider keyword-based and can only detect context-less information. Our main contribution in this chapter are

- The keyword-based definition over algorithms for sensitive information detection
- $n$-grams approaches are keyword-based
- Inference rules approaches are keyword-based

- PMI approaches are keyword-based

While these contributions are interesting in themselves, they are also important because they lay the groundwork for the next generation of approaches, *context-based* approaches. We discuss our work on such approaches in next chapter.

# Chapter 4

# Beyond Keywords: The Paraphrase Analogy

As discussed in previous chapter, existing methods for sensitive information detection rely on context-less, fixed assignment of sensitivity scores on keywords. Such keyword-based approaches are limited in which types of sensitive information can be detected. In particular keyword-based approaches have two main requirements/challenges:

- It must be possible to assign high, fixed sensitivity scores to all keywords which are deemed sensitive.
- Sensitive sentences must contain words that frequently co-occur with the fixed, sensitive keywords.

As a consequence, more complex sensitive information types which requires context-based sensitivity scores cannot necessarily be captured.

In this chapter we address these two challenges and build an family of approaches for detecting sensitive information. Our approaches do not require keywords and consider context when detecting sensitive information. I.e. our proposed family of detection algorithms do *not* assign context-less sensitivity scores to keywords. We move away from the restrictive assumption that we assume access to descriptive sensitive keywords, instead we consider the more general sensitive information detection problem where all that we have access to are labeled *examples* of sensitive and non-sensitive information. Thus detecting sensitive information becomes the problem of detecting whether a sentence contains the same informational content as some of the sensitive examples, perhaps in a re-written form. We here focus on the ability to detect sensitive meaning rather than just known sensitive keywords. The process of taking one sentence and re-writing this sentence using perhaps different words or a different grammatical structure is known as *paraphrasing* (more on this below). Our work is motivated by this view of detecting (sensitive) meaning as a paraphrase detection problem. For the latter problem, recent machine learning techniques show promising results. Thus we consider an approach, which obtains high accuracy in detecting sentences which are paraphrases of each other, as promising for learning to correctly identify sensitive/non-sensitive sentences. In this chapter we detail how we can adapt selected paraphrase detection approaches to sensitive information detection.

Recall that in paraphrase detection we are given two texts and tasked with determining if these two texts are paraphrases of each other, i.e. if they contain the same informational content [88]. We propose to adapt paraphrase detection approaches to model semantics for sensitive information detection. That is, we observe that while it may be difficult to characterize a particular sensitive information type with a few seed words it is generally much easier to come up with examples of sensitive content. E.g. if we assume access to previous examples of sensitive content then we can detect new instances by looking for *paraphrases* of the known sensitive content.

Inspired by this intuition we develop a novel approach for finding sensitive information [74] (Chapter 7) and show experimentally across a number of datasets that this family of approaches can not only find sensitive content characterize by seed words but also find previously undetected content [74] (Chapter 7), [76] (Chapter 10). See also the discussion in Section 5.3.

In this work we extend current context-less approaches (Chapter 3) for sensitive information detection to include larger context. In this section we argue that certain paraphrase detection approaches are promising candidates for study, in particular those approaches which can be adapted to consider the problem of sensitive vs. non-sensitive context, rather than whether two given sentences are paraphrases of each other. In our result overview Section 5.3 we observe that we can successfully adapt recursive neural networks and LSTM (Long Short-Term Memory) recurrent neural networks (see next section, Section 4.1) for sensitive information detection. We observe further in Section 5.3 that at least for one dataset we see a fraction, of over 20%, of sentences which are only correctly identified by our RecNN (Table 5.7), meaning that at least 20% of the sentences correctly identified by the RecNN approach are not captured by LSTM on that dataset, which indicates as we discuss that RecNN can be a strong approach for detecting sensitive information. Including recursive structure, here in the form of compositionality parse trees, can be important in sensitive information detection Section 4.1.

First we introduce the problem of paraphrase detection and review current state-of-the-art approaches to detecting paraphrasing. In next section, Section 4.1, we introduce neural network approaches inspired from paraphrase detection approaches and show how we can adapt such neural approaches to the problem of detecting sensitive information.

In paraphrase detection we are given two sentences and asked to judge if these two sentences express the same meaning, i.e., if they are *paraphrases* of each other. See also Figure 4.1.

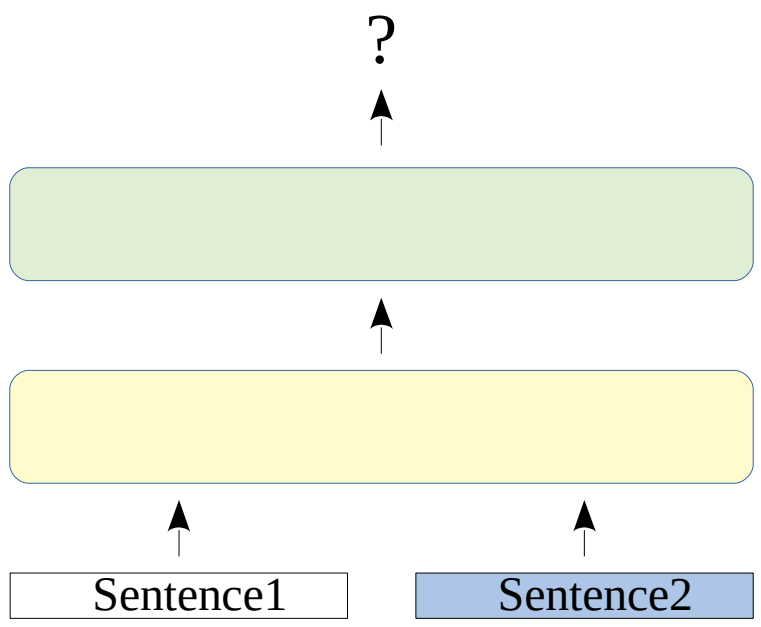

Figure 4.1: Illustration of paraphrase detection. We are given 2 sentences at the bottom. These are processed (bottom of model) and compared (top of model) and a prediction is made (the question mark).

Recall that for sensitive information detection we defined a labeling function, Equation (2.1) (Chapter 2). Similarly in paraphrase detection we may define a labeling function from pairs of sentences to 0 (not paraphrases) or 1 (paraphrases)

$$\mathcal{L}_{\text{paraphrase}} : D \times D \to \{0, 1\}.$$

Adapting paraphrase approaches to the field of sensitive information detection we observe that a sentence which contains sensitive information can express this information in various ways. This means, a paraphrase of a sensitive sentence is still a sensitive sentence. This holds even if we replace some words in the sentence with synonyms, i.e. the sentence would still be both sensitive and a paraphrase. In contrast for keyword-based approaches once the word sensitivity scores are fixed, the approach *cannot* infer if a sentence, which does not contain any sensitive keywords, is in fact sensitive, which means that paraphrase inspired detection approaches should be able to detect sensitive information where keyword-based approaches fail.

Paraphrase detection is a hard problem and thus many of the approaches for this problem rely on semantic understanding of the sentences [88]. An example of two paraphrases is (taken from [88]):

- The judge also refused to postpone the trial date of Sept. 29
- Obus also denied a defense motion to postpone the September trial date

The reader may wish to compare with the sensitive example discussed in Chapter 1. Many of the characteristics of paraphrase sentences also occur in sensitive information detection. For instance in the paraphrase example above we see that several of the words have been replaced by synonyms. For instance the month September is referred to as both “September” and as “Sept” and one sentence refers to the judge by his name “Obus” where the other refers simply to “The judge”. These characteristics we also find in sensitive information. In Chapter 1 we discussed some of the more challenging aspects of sensitive information detection which have a similar structure. E.g. while not containing any sensitive keywords, the sentences are still sensitive because their informational content is a re-write of sensitive content. We observe that current state-of-the-art approaches for sensitive information detection are solely based on counting of co-occurrences and some of the sentences, while being sensitive, do not contain any of the sensitive co-occurrences. We therefore expect paraphrase detection techniques to be beneficial for sensitive information detection.

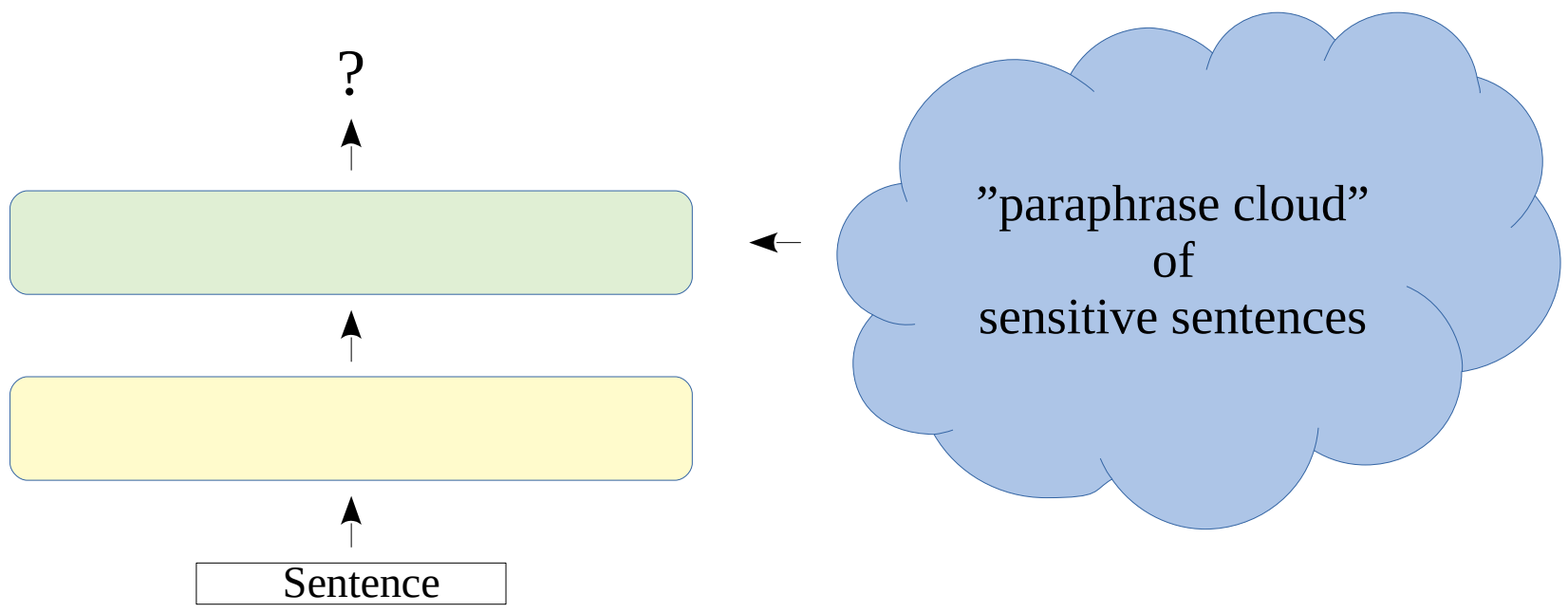

Figure 4.2: Illustration of sensitive information detection. We are given a sentence at the bottom. The sentence is processed (bottom of model) and compared (top of model) with the known sensitive examples. Finally a prediction is made (the question mark). The main difference compared to Figure 4.1 is that we here have to detect sentences which are paraphrases of *any* sentence in the labeled dataset seen at training.

To adapt paraphrase detection approaches to sensitive information detection we conceptually move from a setting as the one found in Figure 4.1 to the one in Figure 4.2. In paraphrase detection we match in pairs, i.e. we only have to determine if a sentence is a paraphrase of the single other sentence given. In Figure 4.1 this is the illustrated as the box with label “Sentence2”. In contrast in Figure 4.2 we must detect sensitive information from paraphrases of a given set of examples. This is referred to as the “paraphrase cloud” in the figure. We train the sensitive information detection approach by optimizing on the model’s ability to predict the labeling function (see also next section, Section 4.1). This training objective will drive the model parameters towards a set of parameters where the model becomes better at predicting the labeling function. By adopting paraphrase detection model architectures we give our models strong preference for generalizing on paraphrases of the training examples. In Figure 4.2 we illustrate this as the “paraphrase cloud”.

The field of paraphrase detection has seen a lot of interest the recent years with new strong models with high performance being published. Here we briefly review previous work, with a particular focus on representation based approaches. Socher et al. [88] applies recursive neural network for paraphrase detection and obtain state-of-the-art results on the Microsoft Research Paraphrase dataset (MSRP)[1]. The authors unfold the node representations into a similarity matrix which is scored using dynamic pooling, e.g. a hierarchical structure of pooling layers. In [13] authors compares different distributional representations with different compositional models and show performance on paraphrase detection. They compare additive, multiplicative and recursive neural network based compositional models. For the recursive neural network based models they use the approach from [88] together with constituency parse trees but argues that they greedily could learn tree structures for the detection. Kiros et al. [53] experiment on data from several domains, among these paraphrase detection. They develop so-called *Skip-Thought Vectors* which are representations over phrases similar to word vectors. A key contribution is the ability to extend representations from known phrases to smaller variations of these phrases (i.e. with some words replaced) by using compositionality of representations. The authors only consider recurrent models for representations and uses logistic regression for prediction but are still able to show promising result on the MSRP dataset.

In [80] the authors investigate the effect of word order information and different compositional models on paraphrase detection. They experiment with additional, multiplicative and a compositional model where adjective-noun phrases are modeled as functional application using matrix multiplication over grammar based parse trees. We note that this functional application can be cast as a linear recursive neural network model. The authors show that models without word order information fail at detecting more complex examples of paraphrases. They also experiment with ability of different compositionality models to distinguish between semantic variations of sentences where we go from a paraphrase to *not* a paraphrase. For instance “A man plays guitar” vs. “A guitar plays a man” or “The man plays no guitar”. They show that the recursive neural network inspired models perform significantly (on average almost an order of magnitude) better for distinguishing paraphrases from these kinds of non-paraphrases.

Compositionality such as represented by parse trees is important for paraphrase detection, as found in Socher et al. [88], For complex sensitive infor-mation detection, compositionality is expected to play a similar role. This is motivated by our observation that particular phrases are indicative of a sensitive information type that goes beyond simple word (co-)occurrences. Compositional context representations in the form of parse trees are thus expected to benefit sensitive information detection. Building complex detection models (such as RecNN) we boost the amount and types of sensitive information which can be detected.

[1] `https://www.microsoft.com/en-us/download/details.aspx?id=52398`

We build adapted the representation based approaches above and use backpropagation to train a recursive neural network particular trained to be good at finding sensitive sentences and which assign similar representations to similar sub-trees in the parse tree. This is discussed in detail in the next section.

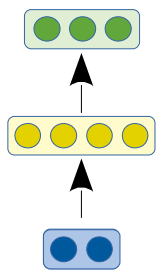

Figure 4.3: Visualization of classic neural network with layers. Input layer has size 2, hidden layer has size 4 and output layer has size 3.

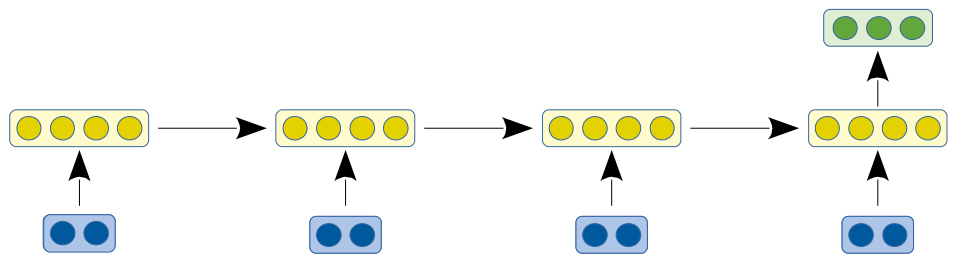

Figure 4.4: Visualization of recurrent neural network (with layers). The main difference to the classical neural network is that the previous hidden state is part of the update to the new hidden state as per Equation (4.3).

## 4.1 Recursive Neural Networks

In this section we lay the technical groundwork for adapting state-of-the-art paraphrase approaches to sensitive information detection. As discussed in previous section these approaches are based on so-called recursive neural networks.

In a regular neural network we transform some input vector $\mathbf{x}$ of fixed length into some output vector $\mathbf{y}$ also of fixed length. Thus for a neural network NN we can denote the operation of the neural network as

$$\mathbf{y} = \mathrm{NN}(\mathbf{x}). \tag{4.1}$$

Let $|\mathbf{x}|$ be the input size and $|\mathbf{y}|$ be the output size of the neural network. We consider *layered* neural networks where the operation in Equation (4.1) can be expressed as a series of operations. We use superscript $\ell$ such that $y^\ell$ denotes the output of layer $\ell$, thus we may write

$$\mathbf{y}^\ell = \mathrm{NN}^\ell(\mathbf{y}^{\ell-1}). \tag{4.2}$$

Let $L$ be the number of layers in the neural network. The input does not count as a layer, but we may still refer to it as the 0th layer, $\mathbf{y}^0 = \mathbf{x}$. We have $\mathbf{y} = \mathrm{NN}(\mathbf{x}) = \mathbf{y}^L = \mathrm{NN}^L(\mathbf{y}^{L-1})$. We follow the literature [34, 37] and refer to all layers $0 < \ell < L$ as *hidden* layers. Layer $L$ is the output layer, neither input nor output layers are considered hidden.

In Figure 4.3 a small example of a layered neural network is shown. The colored horizontal boxes are the vectors given by Equation (4.2). We have the input layer (in blue) with $|\mathbf{y}^0| = |\mathbf{x}| = 2$. We have a hidden layer (in yellow) with size $|\mathbf{y}^1| = 4$ and the output layer (in green) with $|\mathbf{y}| = |\mathbf{y}^2| = 3$. Note the vertical black arrows represent the operation of a particular layer of the neural network, for instance the lower black arrow represent the operation of the 1st neural network layer, $\mathrm{NN}^1$, which transform the input $\mathbf{x}$ to the hidden layer vector $\mathbf{y}^1$ as given by Equation (4.2), $\mathbf{y}^1 = \mathrm{NN}^1(\mathbf{x})$. We refer to $\mathbf{y}^\ell$ as (the value of) layer $\ell$ in the neural network and to $\mathrm{NN}^\ell$ as the operation of $\ell$th layer.

Given this definition of a neural network it is clear that if the functions in all layers are linear in input then the overall function for the entire neural network is also linear. Similarly if the layered functions are polynomial then the complete network is also polynomial.

If each layer consists of a linear combination of the inputs together with a bias term (i.e. an affine transformation) followed by a suitable non-linear function then we can apply the *universal approximation theorem* [25, 45] which states that with this kind of layer transformation and if $L \geq 2$ then our neural network can approximate any suitably discrete or continuous function with bounded input and output to any fixed error $\epsilon > 0$ [34].

Thus it is common to consider neural networks with layers on the following form

$$\mathbf{y}^\ell = \sigma(\mathbf{W}^\ell \mathbf{y}^{\ell-1} + \mathbf{b})$$

where $\sigma(z)$ is the suitable non-linear function, also called the *activation* function. For each layer $\ell$ we denote the length of the input as $N_{\ell-1} = |\mathbf{y}^{\ell-1}|$, and of the output as $N_\ell = |\mathbf{y}^\ell|$. The parameter $\mathbf{W}^\ell$ is a matrix of size $N_\ell \times N_{\ell-1}$, referred to as the weight matrix. The vectors $\mathbf{y}^{\ell-1}, \mathbf{b}$ are a column vectors. By design length of $\mathbf{y}^{\ell-1}$ is $N_{\ell-1}$ and length of both $\mathbf{y}^\ell$ and $\mathbf{b}$ are $N_\ell$.

A wide set of functions can be used as suitable non-linear or activation functions. Common choices include [34, 37]

$$\begin{aligned} \sigma(z) &= \frac{1}{1+e^{-z}} && \text{Sigmoid} \\ \sigma(z) &= \frac{e^{2x}-1}{e^{2x}+1} && \text{Hyperbolic tangent (tanh)} \\ \sigma(z) &= \max(0, z) && \text{Rectifier (ReLU)} \end{aligned}$$

Using backpropagation we adjust the components in the weight matrices in all layers to minimize the error measure. For an in-depth discussion of backpropagation we refer the interested reader to [31, 34].

**Neural networks on text** A text can be considered as a sequence of data (words) from which we wish to approximate some labeling function (e.g. sensitive vs. non-sensitive). Below we discuss how and when it is beneficial to develop a richer structures for text than that of a sequence, and how this may help in learning more complex models over text. Here we consider sequences. When we work with sequences of data (of variable length) it is apparent that a limitation of the neural network is the need for fixed length input. The immediate solution which presents itself is to repeatedly re-apply the neural network at each location in the sequence. This is known as a *recurrent neural network (RNN)* [27, 34]. We briefly introduce notation and concepts for recurrent neural networks before introducing their generalization, the recursive neural network.

Let our input be a text document $d$ (i.e. a sequence) with words $d = (w_1, w_2, \ldots, w_{n_d})$. To encode the words in a way that allow for consumption of neural networks we apply an encoding to the words, i.e. map each word to a real valued vector. Common choices include bag-of-words [41] or word embeddings [71]. Traditional bag-of-words encode a collection of one-hot word vectors and thus do not encode word ordering and do not carry an inherent similarity for similar words [34]. Both bag-of-words and word embeddings can be seen as encoding some form of the Distributional Hypothesis [41], (see also the previous discussion in Section 3.1). Word embeddings have the notable properties that they directly encode a prediction of often occurring contexts and similar words have similar embeddings [34, 71]. Furthermore word embeddings are typically densely encoded vectors and such dense vectors are well suited as input to neural networks. Because of this, and because structure and word order are important in paraphrase detection and by extension in detecting sensitive information, we focus mainly on word embeddings in this work.

Assume we are given a neural network NN which takes as input a word $\mathbf{w}$ (we assume the word already is encoded with a word embedding) and a *state* $\mathbf{h}$. The state is often denoted with symbol $\mathbf{h}$ because the state acts as a kind of hidden layer which connects the evaluations over the sequence $(\mathbf{w}_1, \ldots, \mathbf{w}_{n_d})$. The input space for words is real valued vectors $\mathbb{R}^{N_e}$ for some embedding size $N_e$ and similarly for the hidden state $\mathbb{R}^{N_h}$. The dimensionality of the spaces $N_e, N_h$ are hyperparameters which can be fitted to optimize the performance on the given dataset $D$. Assume the initial state is the zero vector $\mathbf{h} = \mathbf{h}_0 = \mathbf{0}$ then the first application of the neural network is

$$\mathbf{h}_1 = \mathrm{NN}(\mathbf{w}_1, \mathbf{h}_0)$$

Thus we can define a recursive rule for applying the recurrent neural network given the result of the $t-1$ application

$$\mathbf{h}_t = \mathrm{NN}(\mathbf{w}_t, \mathbf{h}_{t-1}). \tag{4.3}$$

A visualization of a recurrent neural network following applications of this update rule can be seen in Figure 4.4. In the figure the input layer (the blue layer) is the words encoded via word embeddings. In this figure we have 4 input words and therefore 4 hidden states $\mathbf{h}_1, \mathbf{h}_2, \mathbf{h}_3, \mathbf{h}_4$, the initial hidden state $\mathbf{h}_0$ is not shown. In the figure there is a output layer (the green layer at the final step), this is simply another neural network layer which can be applied to the final hidden state (as shown in the figure) or, another option is to apply the output layer at all steps and thereby obtain a prediction for each step. We note that many other options for choosing when to generate output and on which hidden states exists [31, 34] but in this work we will focus on the former option described above, i.e. one prediction on the final state in the sequence. All evaluations in this work done on recurrent neural networks (mainly LSTM networks) are done only using the last final state for prediction. This is similar to how we predict based on the final state for recursive neural networks, see below.

The output neural network layer is denoted as $\mathrm{NN}_O$,

$$\mathbf{y}_t = \mathrm{NN}_O(\mathbf{h}_t).$$

I.e. if predicting sensitive (1) vs. non-sensitive (0) then $\mathbf{y}_t$ could be the predicted probability of the document containing sensitive information (in which case $\mathbf{y}_t$ would be a vector of dimension 1). We observe that for such recurrent approaches, the final hidden state $\mathbf{h}_{N_d}$ used for predicting if this document contains sensitive information can be viewed as encoding the most information about the document that our approach can obtain. I.e. recurrent neural networks are not keywords-based, instead they encode context as $\mathbf{h}_{N_d}$. We say that recurrent neural networks are *context-based* approaches and can detect sensitive information beyond context-less sensitive information.

**Recursive neural networks on text** Recurrent neural networks as presented above can only process input which have sequential form. While in the text data domain we can view a document as a sequential data structure, there is much structure in text such as compositional and constituency dependency parse trees which have been shown to accelerate and improve machine learning models for problems such as paraphrase detection [88]. In principle we could add such additional structure information by infusing the structure knowledge into our sequential data, simply by encoding these structures sequentially and then apply sequential machine learning models on this enhanced sequential structure. For instance, if our input is structured as a binary tree we can flatten this tree into an sequence with sub-trees enclosed by parentheses. However such an encoding have two main challenges

- The length of paths between nodes in the sequential encoded data may be much longer than in the original structure. For instance comparing a balanced binary tree with the corresponding sequential encoding; most nodes will be exponential further away from most other nodes when sequential encoded (if the tree is bi-directional then the longest path is $2 \log n$). Thus learning challenges such as vanishing gradient or exploding gradients [8, 44] becomes a (exponential) worse problem as we consider longer documents.

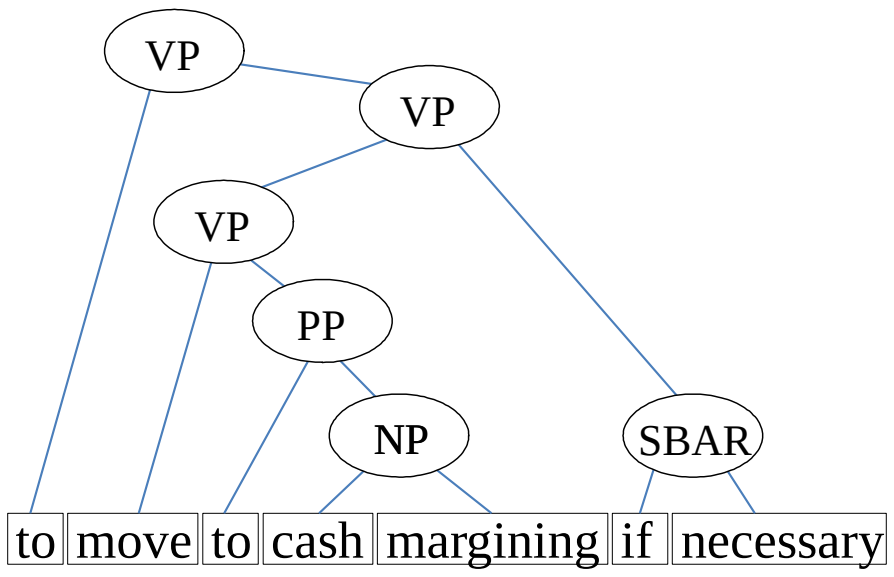

(a) Parse tree

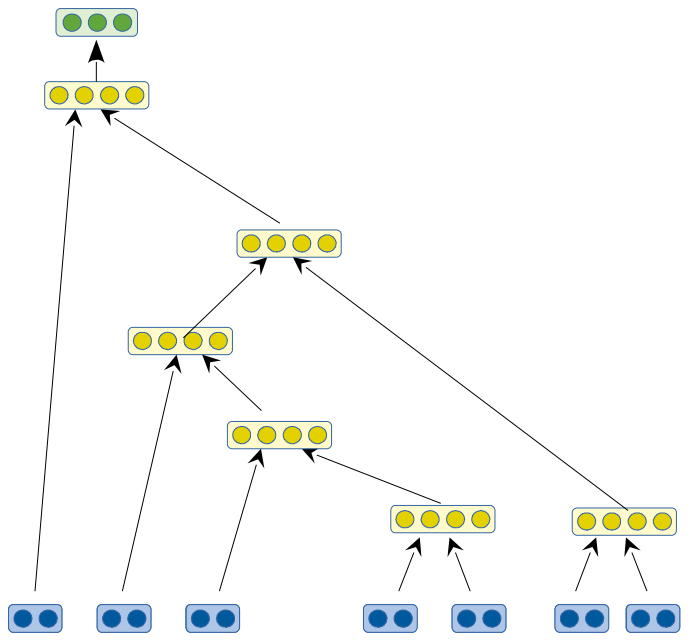

(b) Recursive neural network

Figure 4.5: The parse tree and the corresponding recursive neural network. Each node in the parse tree becomes an application of the neural network. The neural network thus generates a (hidden) representation for each node in the structure. Note arrows show forward evaluation direction, which corresponds to ancestor direction, e.g. the arrows point in direction of parents.

- Making the structure sequential will invariably break regularities and potential interesting relations in the original structure [31]. Close neighboring nodes in the structure may end up far from each other in the sequential encoding. This incur not only the vanishing gradient problem discussed above, but also harder learning since the structure becomes broken. In a nutshell; in sequential form we can only encode certain aspects of a rich structure in a localized way, the general aspects of the structure tend to be distributed across large distances when in sequential form, thus breaking structure.

These challenges make learning on sequential encoded structures harder or even in-tractable [31]. For paraphrase (and by extension sensitive information detection) where context and structure are important aspects to consider we need to ensure that our model can easily learn from structure, thus we propose to use *recursive neural network* which are capable of learning from advanced structures such as parse trees. To keep the distinction between recursive neural networks and recurrent neural networks clear we adopt the abbreviations RecNN for the former and RNN for the latter. Below we define the recursive neural network with particular emphasis on document datasets.

A *recursive* neural network can be applied to a sequence of data with some additional structure. E.g. a sentence and the parse tree for that sentence. In this work we are mainly interested in input structures which are binary trees. For a full introduction to general recursive structures see [31]. Given a document $d = (w_1, w_2, \ldots, w_{n_d})$, and a tree $\mathcal{T}_d$ defined recursively over the input. I.e. we consider the tree $\mathcal{T}_d$ as a function which returns a node from some set of nodes $U = \{n_1, \ldots, n_{n_d-1}\}$ and as input the tree function takes either the words from input $d$ or a node from the node set. We denote the input set as

$$V = U \cup \{w_i : w_i \in d\}$$

We have that the tree is a function from nodes and words to a node, which is considered the *parent* node in the tree structure. I.e. we have

$$\mathcal{T}_d : V \to U \cup \{\epsilon\}$$

Where $\epsilon$ is a special node value which is not considered part of the tree. E.g. $\epsilon$ is the empty or *null* node. Thus the tree function $\mathcal{T}_d(v)$ returns the parent node (in the tree) of input $v$, where $v$ can be either a node or a word. We focus on trees where each node is parent to two children from set $V$, i.e. the tree is binary (but not necessarily balanced). Note words are not in $U$ and has no children. Only one node may have no parent. We refer to that node as the *root* node of the tree.

Consider the tree in Figure 4.5a. This is a constituency parse tree generated over the sentence from the example in Chapter 1 (here shorten for clarity). The parse tree was generated from a probabilistic context-free grammar (pcfg) [56] and the node labels are from Stanford Penn Treebank [95]. The pcfg was trained on the Stanford Penn Treebank, i.e., the texts and node labels in the treebank. We followed Klein and Manning and used a CKY parser [56] to obtain rule probabilities. See Section 10.3 for further details on how we obtain the parse trees. All our parse trees, used in experiments and released as resources are obtained via the same parser.

We see the parse tree is a structure which fulfill the requirements just discussed. For instance, in the parse tree we have a node with the label "SBAR". This is the parent of two nodes, to the left it is parent to the input node of the word "if" and to the right it is parent of the input node of word "necessary". Furthermore all nodes (except the word leaf nodes) have two children and there is only one root node which has no parent.

To define a recursive neural network over documents $d$ and corresponding trees $\mathcal{T}_d$ we associate with each member of the input set $V$ a *representation*, i.e. some vector value from $\mathbb{R}^n$ for some $n$. For the words $w_i \in d$ this could be a word embedding vector and for nodes $n_i \in U$ this would be a hidden state value similar to the one calculated by a recurrent neural network in Equation (4.3). Let RecNN be some neural network which as input takes representations of two elements from $V$ and outputs a representation for the current node. I.e. let $t \in U$ be any node from $U$ and let $l(t)$ and $r(t)$ be (left and right) children of node $t$, i.e. $t = \mathcal{T}_d(l(t))$ and $t = \mathcal{T}_d(r(t))$, we have

$$\mathbf{h}_t = \text{RecNN}(\mathbf{h}_{l(t)}, \mathbf{h}_{r(t)})$$

where $\mathbf{h}_{l(t)}$, $\mathbf{h}_{r(t)}$, $\mathbf{h}_t$ are the representations of $l(t), r(t), t$ respectively. A recursive neural network which follows the step rule given above is shown in Figure 4.4 the structure used to define the recursive neural network is the one shown in Figure 4.5a.

Using generalized versions of the equation above it is possible to define recursive neural networks for trees that are not binary and also for directed acyclic graphs (DAG) or even for general recursive structures, see also [31].

Similarly to how we defined the output of a recurrent neural network we define the output of a recursive neural network as dependent on the hidden state at that node

$$\mathbf{y}_t = \text{RecNN}_O(\mathbf{h}_t), \quad t \in U$$

Initially the child node representations will be for the words in the sentence, but as we move up in the tree structure $\mathcal{T}$ one or both child representations will be for internal nodes where the representation has been generated by our network.

Please note that it follows from the above definition of recursive neural networks that we may view a sequential recurrent neural network as a special type of a recursive neural network where the structure is constructed such that the execution order for the recursive neural network is the same as for the recurrent neural network. Thus recursive neural networks are a generalization of recurrent neural networks, where we in the recursive neural network also consider structure. It follows that recursive neural networks are *context-based* approaches to sensitive information detection, as recurrent neural networks are.

Backpropagation over standard neural networks was formulated in the seminal work [83]. Applying recurrent neural networks over series is from [27], we follow here the presentation given in [34]. Discussion of problems with learning over many steps and the introduction of the Long Short-Term Memory (LSTM) cell is from [44]. Bengio et al. [8] also show that the gradient may grow or shrink exponentially over number of steps and thus demonstrate the learning problem of vanishing or exploding gradients. Backpropagation over structures was done as backpropagation-through-structure, which was introduced in [35] and further formalized as backpropagation over recursive execution graphs in [31].

In [88] authors do pre-training with an autoencoding neural network, i.e. where we train the network to predict the input using restricted access to information. These so-called autoencoders was trained to learn the representations in recursive structures. In [46] authors show that it is possible to obtain high accuracy and good recursive state representations without the need for the time consuming pre-training. In [46] there is access to handcrafted target values for each sentence and node in the parse tree.

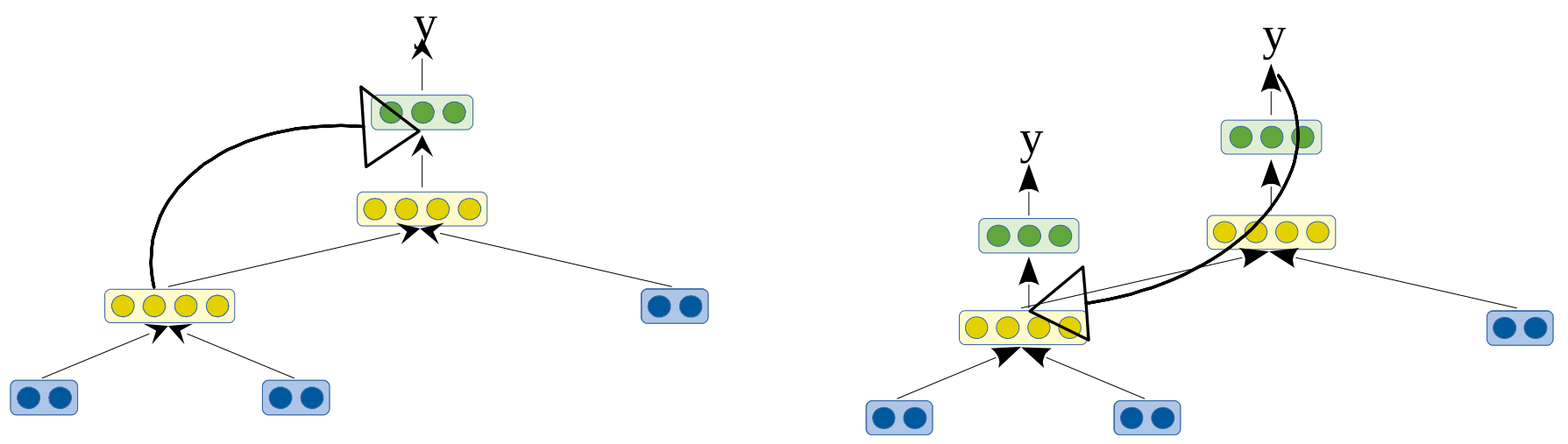

(a) The standard shortcut. (b) Inverse shortcut.

Figure 4.6: The difference between standard shortcuts and our inverse shortcuts. The standard shortcut is given by Equation (4.4). The inverse shortcut, with update gradient given by Equation (4.5).

### Inverse shortcuts

A main drawback of the recursive neural networks is the need for finely labeled training examples. Training from a single root node label over a large structure incurs the challenges of training over many steps such as vanishing and exploding gradients [8, 44]. Authors [46, 90] use finely labeled datasets with labels for each node in structure to obtain strong results with structured models without encountering vanishing and exploding gradients.

For deep convolutional networks the concept of *shortcuts* has been introduced to counter the challenge of vanishing gradients [43, 92]. Recall Equation (4.2) the output of standard layer in a neural network is given as

$$\mathbf{y}^{\ell}.$$

In the simplest form a shortcut alters the output of the layer to (see also Figure 4.6a)

$$\mathbf{y}^{\ell}_{\text{shortcut}} = \mathbf{y}^{\ell} + \mathbf{y}^{\ell-1}, \tag{4.4}$$

where $\mathbf{y}^{\ell-1}$ is the input to the unmodified $\ell$th layer. Thus the new output is weighted between the unaltered output and the original input, e.g. there is a "shortcut" from the input to output.

To see that this helps counter vanishing gradients observe that at backpropagation time the error signal will propagate both through the $\ell$th layer but also through the shortcut unaltered. Thus the error signal may travel a higher number of layers before vanishing.

Inspired by such shortcuts we propose *inverse shortcuts*, i.e. shortcuts from prediction to early layers. We referred to this in [74] as *propagating labels*. For each representation in our structure $\mathcal{T}$ we add a prediction node and try to predict the label for the complete tree from the partial information available at the current node. This is shown in Figure 4.6b.

Note the authors in [46, 90] used hand labeled sentences at each node in the parse tree. These were obtained using a resource heavy process on Amazon Turk [90]. The benefit of our proposed inverse shortcuts are two-fold. 1) We introduce shortcuts into the training which improves performance for deep learning [43, 92] and we avoid the costly fine labeling process by "shortcutting" the sentence label. Thus the concept of inverse shortcuts is an addition to deep neural networks in general and in particular to the approach for sentiment analysis proposed in [90].

Let the set of ancestors for node $v$ be

$$P_v = \{\mathcal{T}(v), \mathcal{T}(\mathcal{T}(v)), \mathcal{T}(\mathcal{T}(\mathcal{T}(v))), \ldots\}.$$

Let the backpropagation update gradient (e.g. the update signal for the model parameters as derived using backpropagation) from a parent node $u \in P_v$ be denote $g_v(u)$. In a regular network with only root node prediction and without inverse shortcuts we have that the update gradient is determined by error signal at the root node and the path between the root node and node $v$

$$g_v(\text{root}).$$

If we encounter the vanishing gradient problem then this update gradient will decay as a function of the distance between the root node and node $v$. When we apply inverse gradient the update gradient becomes

$$\sum_{p \in P_v} g_v(p). \tag{4.5}$$

I.e. a sum over update gradients from all ancestors in the tree. Thus we circumvent the vanishing gradient problem by introducing update gradients from close ancestors in the tree. Our strong results in [74–76] (see Part II) were obtained using inverse shortcuts.

## 4.2 Our Contributions

We introduce a new family of approaches for *context-based* sensitive information detection. We connect the field of paraphrase detection to that of sensitive information detection and show that sensitive information detection can be cast as a adapted form of paraphrase detection. The key concept for driving this connection is the observation that

> *A paraphrase of sensitive information is itself sensitive. Thus we want to be able to detect paraphrases of known sensitive information.*

Using this insight, our main contributions include

- Casting sensitive information detection as a special kind of paraphrase detection.

- Focusing on informational content rather than specific topic or keywords.
- A new family of *context-based* approaches for sensitive information detection.
- A encoding process of context through recursive neural networks.
- In our publications, Part II, we evaluate our approaches against state-of-the-art approaches for sensitive information detection. We demonstrate that we significantly outperform previous approaches.
- We propose inverse shortcuts to counter vanishing gradient. As a plus adopting inverse shortcuts reduces the need for finely labeled data samples. This reduces the requirements for annotators to label on phrase level which previous RecNN approaches require [90].
- The results of our evaluations demonstrate the strength and applicability of inverse shortcuts to sensitive information detection.
- In [75] we address the important problem of effective training of models over complex structures. We show that by clustering document representations we can do selective training with fewer updates to the model but where the final model performs at least as well as a traditional trained model.
- For Selective training we show that *Most Frequently Occurring* class label measure, $\Delta MFO$, allows us to train models using hard optimization conditions with a large speedup factor, see also Chapter 9.

# Chapter 5

# Evaluation, Data and Result Analysis

Evaluation of sensitive information detection is challenging in that no benchmarking dataset is available. Due to the nature of the problem most organizations are hesitant to release actual sensitive data for research and often any accidental released sensitive information will be followed by an redaction effort. As a consequence there are no publicly available datasets where sensitive information is retained, or even labeled as such. Therefore, researchers have proposed different methods to prepare document datasets such that the dataset can be used for evaluation purposes. We briefly review them in the following, and characterize their strengths and weaknesses.

Chow et al. in [16] develops the idea that a particular sensitive information type, characterized by a single seed keyword is sensitive. They develop two datasets; one extracted and developed from Wikipedia and the other from the Enron dataset. The authors define the task of finding sensitive information corresponding to that of a classification task and they evaluate the inference rules (see also Section 3.2) approach for this task definition. The first evaluation is using the seed word "HIV" and finding inference rules which predicts the seed word. Here Wikipedia is used as a dataset and inference rules are found where the seed word can be predicted high confidence. Evaluation is done on top 70 predicted rules by a human evaluator. A large fraction of the found inference rules are deemed correct by the evaluations.

For their 2nd dataset Chow et al. argues that the Enron dataset is more relevant to the task of detecting sensitive information than Wikipedia as it is a "private" dataset in the sense that it contains emails and other documents that are typically only available inside a company, but not shared with the outside world. Similarly to the first evaluation they choose a single seed word, "University of Wharton" to define the sensitive information to detect. As the Enron documents are not labeled, the authors apply the assumption is that *all* documents containing the keyword are sensitive. Precision is calculated as the fraction of documents predicted to be sensitive which also mentions the seed word. To calculate recall special care has to be made. I.e. the pathological rule with 100% recall is

$$\text{“University of Wharton”} \rightarrow \mathcal{L} = 1.$$

Thus to make a more fair comparison the authors compute recall of their approach by simply ignoring rules which uses the term “University of Wharton” as a condition to predict, i.e. where “University of Wharton” occurs on the left-hand-side of the rule.

We propose using the Enron dataset as well, as it is indeed a rare case of private data out in the public domain. However, as we will discuss in Section 5.1, we make use of labels that were generated based on human defined complex types of sensitive information and thus our evaluation reflect human judgment on types of information rather than seed word and rule based evaluation which we have shown only can detect context-less sensitive information types.

The authors Grechanik et al. [38] develops further the inference rules approach and obtain good results using change in entropy as a measure of *privacy loss*. The datasets they work on are software repositories, such as PCAnywhere by Symantec[1], Mozilla[2],and from Opentaps[3]. Where they search for sensitive information in function names and in code comments. Different types of sensitive information are defined by the authors using seed keywords sets containing up to 16 distinct words.

In this work we move beyond context-less sensitive information. Our public evaluation and benchmarking datasets have human labeled documents rather than seed word based definition such as the ones discussed above. Furthermore we consider sensitive information in a general NLP setting, i.e. where the sensitive content can be embedded in informational content rather than narrowing the dataset and labels to function names and named entities as above. A narrow dataset such as the one above may be less applicable for evaluating context-based approaches.

In [42] Hart et al. uses WikiLeaks[4], Enron and Google developer forums as sensitive information sources. From this the authors develop several pairs of datasets to evaluate sensitive information detection. In each pair, one document collection is assumed to be sensitive, while the other is considered non-sensitive:

- WikiLeaks (23 documents, military field manuals) vs. corporate web-site (174 documents) for DynCorp[5]

[1] Proprietary source code leaked by hackers http://www.computerworld.com/article/2472272/endpoint-security/antisec-leaks-symantec-pcanywhere-source-code-after--50k-extortion-not-paid.html

[2] https://www.mozilla.org

[3] http://www.opentab.org

[4] https://wikileaks.org

[5] http://www.dyn-intl.com/

- WikiLeaks (102 documents) vs. corporate web-site (120 documents) for Transcendental Mediation[6] (a religious organization)
- WikiLeaks (3, where one long document was split into several) vs. web-pages (277 documents) for Mormon[7] (religion)
- Enron dataset (0.6 million documents) vs. corporate web-site (581 documents)
- Developer blogs (open source projects at google, 1119 documents) vs. PR related blogs (1481 documents)

In addition, they also provide background non-sensitive documents which they refer to as *Non Enterprise (NE)*. I.e. in this setting we detect sensitive information in enterprise documents but we are also interested in performance on sets of non-sensitive examples such as NE documents. Let the set of sensitive documents be denoted $S$, the set of enterprise documents be denoted $E$ and the set of non enterprise documents be $NE$, we have

$$S \subseteq E \quad \text{and} \quad E \cap NE = \varnothing.$$

The NE documents are sampled into the datasets to provide further types of non-sensitive documents for training and testing. The authors use Wikipedia news article, Reuters news articles and the Brown dataset as NE sources.

While this approach makes it possible to define more general sensitive information types without having to rely on single keywords, it has its limitations in the dissimilarities of the document collection pairs. For example, sensitive military/religious manuals and the type of documents found on corporate web-sites, Wikipedia news and other news sites differ not only with respect to sensitive content but also with respect to other properties and style of language used.

The authors obtain high detection results however do not address if performance results might reflect the difference in documents origin more than the ability to isolate and detect sensitive information. Another potential issue lies in the large cardinality imbalance of document collections ranging from few hundreds to more than half a million documents.

In the domain of general sensitive information detection we expect our approaches also might encounter the more challenging situation where sensitive and non-sensitive documents have been generated by similar processes. To obtain strong results our approaches may have to rely on distinguishing properties such as the informational and semantic content rather then document structural properties such as length, language and domain.

---

[6] https://www.tm.org/
[7] https://www.lds.org

Sánchez and Batet [84] base their approach on pointwise mutual information (PMI, see Section 3.3) to define the relationship between sensitive words and other words in the document collection.

The authors evaluate on Wikipedia pages of individuals (like movie stars). The goal is to remove personal sensitive information on subjects such as diseases, city of birth and religion. These topical personal sensitive information can typically be defined using single seed keywords such as "AIDS". The authors use human manual generated labels for sentences from Wikipedia pages. Unfortunately, the dataset is not publicly available. On their focused dataset the authors obtain strong results, however more complex types of sensitive information are not studied.

In the following we present 8 different types of sensitive information. Our contribution is based on two corpora, the Enron corpus and the Monsanto corpus. For each corpus we obtain 4 publicly available different definitions of sensitive information on the document level. Furthermore we have human annotators provide labels on the sentence level for parts of the Monsanto corpus. We present each sensitive information type as its own dataset with text, parse trees and labels. In total we release more than $750,000$ labeled sentences in 12 datasets.

## 5.1 Enron Dataset

In this section we discuss how we may utilize the Enron corpus for sensitive information detection evaluation and what the challenges are and how these can be addressed.

We propose a public dataset and a labeling which contains different kinds of semantic relationships in the form of different sensitive information types. Similar to other authors in the sensitive information detection domain we make use of the Enron dataset as a way to access a dataset of real-life complex interactions. As noted by other authors in this field (see the discussion of previous work above), we find that the Enron dataset is a particularly useful one in that it was not originally intended to be published, but is still out in the public domain today, and can be used for studying sensitive information detection.

The Enron dataset was published in 2003 by FERC (Federal Energy Regulatory Commission) [58]. The dataset contains more than $600,000$ documents and emails from the communication of several employees of the Enron corporation. Some documents are internal coordination between employees, some are spreadsheets or contracts send around for feedback or to customers and some documents are emails containing advertisement from 3rd party vendors. Only minor redaction of private information has occurred and no redaction of the type of complex information that we are interested in has occurred.

A realistic evaluation and benchmarking dataset also needs access to several types of sensitive information, i.e. definitions which are different in nature and cover a wide variety of areas. To capture sensitive content which move beyond context-less sensitive information the types must also cover areas which are not necessarily clear cut, but which can be understood and searched by human users. Thus we must move beyond labels that are obtained through seed words or simple rules. We therefore propose to use human expert labels given to the Enron dataset as part of a competition of the TREC conference.

In the Legal track of the TREC conference [96] an effort has been undertaken labeling the Enron dataset according to a number of conceptually very different *requests*. Each request is defined as a *request for production of documents*. This is a legal construct where a person/lawyer is responsible for finding all relevant documents according to the particular request. The definitions of the requests can be seen in Table 5.1, taken from [22, 96]. Many of these requests are for information on a very specific type of information, e.g. request 201, Prepay Transactions, while some are more classical topical types, such as computer games, e.g. request 207, fantasy football. A subset of 2720 documents for each request of the Enron dataset has been labeled by law students and professionals [96]. For each request, the TREC annotators have annotated the 2720 documents with true-false labels. Each request is thus a binary decision problem with both positive and negative samples.

We processed and release datasets 201 (PPAY), 202 (FAS), 203 (FCAST) and 204 (EDENCE). See Chapter 9. PPAY deals with prepay transactions and FAS with transaction which may be "FAS" compliant. FCAST is discussion of financial forcasts and finally EDENCE is about tampering with evidence. All 4 types of information are examples of information that may be considered sensitive, and where both everyday wording and specialized notions (from finances, law, etc) are common. This is why we select these as representative of sensitive information types and we empirically evaluate that they contain more than context-less sensitive information in the sense of Definition 2. I.e., these datasets contain context-based sensitive information. See also Chapter 7 and Chapter 10. By contrast, for instance label 207 about sports is relatively easy for traditional classification [14] and therefore chosen as a definition of a complex sensitive information type.

If we wanted to do only standard text classification we could use some of the standard news wire classification datasets [14] however these categories are very broad in the sense that you do not actually need to understand for instance politics or baseball to be able to classify texts about these topics. I.e. we expect keyword-based approaches to perform well on such topical datasets. On the other hand classifying sub-parts of text which is about prepay transaction requires a much better model for detection as demonstrated by our experiments in [74, 76], see Chapter 7 and Chapter 10.

In Chapter 7 we work with request 201 (PPAY) and show that we obtain consistent results when approaching prepay as a form of sensitive information detection. In Chapter 9 we work with requests 201 (PPAY), 202 (FAS), 203 (FCAST) and 204 (EDENCE) and release a labeled dataset consisting of $747,801$ constituency parse trees over sentences.

| Topic number | Call text |
|---|---|
| 200 | All documents or communications that describe, discuss, refer to, report on, or relate to the Company's engagement in transactions concerning Real Estate. |
| 201 | All documents or communications that describe, discuss, refer to, report on, or relate to the Company's engagement in structured commodity transactions known as "prepay transactions." |
| 202 | All documents or communications that describe, discuss, refer to, report on, or relate to the Company's engagement in transactions that the Company characterized as compliant with FAS 140 (or its predecessor FAS 125). |
| 203 | All documents or communications that describe, discuss, refer to, report on, or relate to whether the Company had met, or could, would, or might meet its financial forecasts, models, projections, or plans at any time after January 1, 1999. |
| 204 | All documents or communications that describe, discuss, refer to, report on, or relate to any intentions, plans, efforts, or activities involving the alteration, destruction, retention, lack of retention, deletion, or shredding of documents or other evidence, whether in hard - copy or electronic form. |
| 205 | All documents or communications that describe, discuss, refer to, report on, or relate to energy schedules and bids, including but not limited to, estimates, forecasts, descriptions, characterizations, analyses, evaluations, projections, plans, and reports on the volume(s) or geographic location(s) of energy loads. |
| 206 | All documents or communications that describe, discuss, refer to, report on, or relate to any discussion(s),communication(s), or contact(s) with financial analyst(s), or with the firm(s) that employ them, regarding (i) the Company's financial condition, (ii) analysts' coverage of the Company and/or its financial condition, (iii) analysts' rating of the Company's stock, or (iv) the impact of an analyst's coverage of the Company on the business relationship between the Company and the firm that employs the analyst. |
| 207 | All documents or communications that describe, discuss, refer to, report on, or relate to fantasy football, gambling on football, and related activities, including but not limited to, football teams, football players, football games, football statistics, and football performance. |

Table 5.1: The list of Enron labeled requests for documentation [22, 96]. The requests have been slightly shorten from original form in order to fit here. Note that request 200 was only defined through an example set of documents, for that request we have created the textual definition given here. In our published datasets we have used labels $201, 202, 203$ and $204$. See discussion in text.

| Dataset | Descriptive text |
|---|---|
| *GHOST* | The documents concerning article writing and peer-reviewing by Monsanto paid people. Hiding the Monsanto connection, as well as concerning efforts in pressuring journals to retract damning studies in a way such that Monsanto is not seen as providing the critique. |
| *TOXIC* | The documents concerning discussions and testing of the chemical glyphosate which is part of Roundup. This includes in particular the toxicity of glyphosate and decisions not to fund further studies, decisions not to do requested studies or provide data to regulators. Discussions in which ways Roundup can and cannot be considered toxic. |
| *CHEMI* | Monsanto articles on Roundup chemistry when Roundup is used in nature. Internal email discussions on Monsanto studies as well as external studies and measurements together with findings, discussions on which questions are answered or need to be answered by Monsanto paid studies. Discussions on which paid studies are too “risky”, discussions on not publishing (hiding) bad (unexpected) results. |
| *REGUL* | The documents concerning discussions of rewarding people for science that protect Roundup business. Documents on active efforts to monitor and influence regulative bodies for possible problematic rulings related to Roundup. Documents on planning and discussions where specific people from regulatory bodies are being convinced to lower the rating or concerns for Roundup/glyphosate. |

Table 5.2: The list of Monsanto labeled information types. Reproduced from our paper [76]. See also Chapter 10 (page 118).

This is the first large-scale, human labeled sensitive information dataset available publicly. We provide 4 different types of sensitive information definitions in a large dataset and in an easy to access format. Links to data and code can be found on page 70 in Section 5.4.

## 5.2 Monsanto Dataset

As discussed in previous section the Enron dataset is an unique dataset, so far, with regards to real-world communication containing potential sensitive content. However just one dataset for benchmarking and developing true, general methods for sensitive information detection is not enough. In fact having only one dataset may introduce a bias in the models developed, such that these models may become tuned to particular structure only found in the Enron dataset and not in datasets with sensitive information in general. Also the Enron dataset is from 2002 and thus more than 15 years old and while the dataset provides unique access to data show-casing sensitive vs. non-sensitive information, the Enron dataset may no longer be as representative of internal present day human communication as it was 15 years ago.

To remedy these challenges we publish a new dataset providing expert labels and providing a up-to-date modern dataset with several different types of sensitive information. See Chapter 10. We propose using the recent publicized documents which are part of the Monsanto trial [7].

Early in the Monsanto trials the American legal system decided to lump different smaller trials into fewer larger trials. This is known as *centralization* and the legal document enforcing the centralization is known as a *transfer order*. The transfer order for the Monsanto trials succinctly summaries the overall topic of the trials[8]

> "`Roundup, a widely used glyphosate-based herbicide manufactured by Monsanto Company, can cause non-Hodgkin's lymphoma and that Monsanto failed to warn consumers and regulators about the alleged risks of Roundup`."

In the trial several individuals are suing Monsanto and the so-called *secret Monsanto papers* have been released to the public as part of the pre-trial discovery process where each party in the trial has access to relevant documents in order to build the evidence for the trial [7].

Redaction of these documents have been done to protect personal information for instance about people not related to the trial. Similar to the Enron documents this redaction is very light and as such the Monsanto documents contain several different types of information which could be consider sensitive. The documents do not come with predefined labels, again similar to the Enron documents. To remedy this we propose to use a labeling carried out by the law firm Baum, Hedlund, Aristei & Goldman where an effort has been made to labeled initial released documents into 4 different types of information relevant to the trial. The types of information are listed in Table 5.2. The information types can be viewed as sensitive in 2 ways;

[8] `https://www.cand.uscourts.gov/filelibrary/2886/JPML-transfer-order.pdf`

- We can choose to view this as information that is sensitive to the Monsanto company and potential information they wish to detect and protect, or
- We can view this as sensitive information for the suing partners and as important information which we need to be able to detect and bring attention to.

In both cases automated tools for detecting these types of sensitive information becomes important and motivates the use of the secret Monsanto papers with labels as a new dataset for sensitive information detection.

All 4 topic descriptions in Table 5.2 are considered potential complex sensitive information types. The dataset *GHOST* is communication about for instance ghostwriting where complex communication patterns in the e-mail correspondence is ongoing and which in the two perspectives in the bullet-list above can be consider as sensitive information. Similarly for *REGUL* there are discussions of how to influence discussions and choices in government regulatory bodies which are also sensitive in both perspectives. *TOXIC* deals with whether or not Monsanto (and the public as seen from internally within Monsanto) deems Roundup and similar products to be toxic, e.g. cancer producing. On the other hand *CHEMI* deals with experiments, chemical components and other fact based communication about how environment and usage affects the safety of Monsanto's products. All information types are clearly complex in composition and thus good candidates for complex sensitive information datasets. We empirically evaluate the types of sensitive information in these datasets in Chapter 10.

In the following we detail the extraction and labeling process of the Monsanto dataset.

### Extraction and labeling of Monsanto datasets

*Note: moved from chap 10. Only minor changes, except for the annotation guidelines.*

**Downloading.** We downloaded the raw documents from the website of lawfirm Baum, Hedlund, Aristei & Goldman[9]. There are 274 documents of various types. The lawfirm has made a human readable description of 120 links to documents[10]. We cleaned up the human readable list of documents and used the 4 aforementioned types of sensitive information, Table 5.2 to label the documents, based on the document labels provided by the lawfirm. The lawfirm has provided the documents ids in the form *MONGLYxxxxxxxx* or *ACQUAVELLAPRODxxxxxxxx* where the $x$s are replaced with actual numbers.

[9] `https://www.baumhedlundlaw.com/toxic-tort-law/monsanto-roundup-lawsuit/monsanto-secret-documents/`

[10] `https://www.baumhedlundlaw.com/pdf/monsanto-documents/monsanto-papers-chart-1009.pdf`

| dataset | #d | #url | # ids |
|---|---|---|---|
| *GHOST* | 29 | 29 | 53 |
| *TOXIC* | 40 | 39 | 95 |
| *CHEMI* | 17 | 17 | 23 |
| *REGUL* | 33 | 33 | 58 |
| Total | 119 | 118 | 229 |

Table 5.3: Some statistics over the Monsanto dataset. Number of documents are handcounted from the description document. The urls are matched to extracted/harvested metadata. For one document we were not able to match the link and the document. The ids are the number of unique ids (*MONGLY*...) used to refer by the law firm to the documents. See the discussion in the text for further information.

We removed the documents which are not directly related to the case (e.g. links pointing to *forbes.com*). We found 3 documents without previously assigned id and assigned these documents unique ids in the form *ASSIGNEDIDxxx*, where $x$s are a unique number. Several documents have several given ids, probably because the document can be or has been used in several different arguments in the court. We found one instance where two document links with different link text pointed to the same document. This document has only been included once in our dataset.

**Content.** Each document is annotated by a number, an id, a link, a link text and a description. An example is shown in Table 10.1 (Chapter 10).

All documents are pdf documents. Some are exported from emails, word documents, and so on. Some of the documents are or contain scanned images of their text without any optical character recognition. We extracted all text encoded in the documents but have not attempted to extract optical data (i.e. we have not applied any OCR) from the documents.

From the 118 documents we obtain 1.8 mega-bytes of raw unformatted text. There are $37,720$ punctuation (character '.') occurrences in the files in total, i.e. we expect in the order of, say, $20,000$ labeled sentences from the dataset. Note that we use the document labels to label the sentences. See table 5.3 for a summary of the number of documents across the 4 types of sensitive information.

**Tokenization.** We applied the NLTK toolkit [12] and tokenized sentences using the Punkt sentence boundary detection approach [55]. Given a sentence we generated probabilistic context-free grammars (pcfg) based constituent trees [56], where the pcfg was trained over the Penn Treebank [95].

Before tokenizing sentences, we removed email headers, except subject, and we removed ids (e.g. *MONGLY...*) and non-ascii characters. Our sentence tokenizer yields 10,774 sentences. The length distribution is shown in table 10.2 (Chapter 10).

**Sentence selection.** We cleaned the data further by removing very short sentences and very long sentences. We set the cutoff for minimum 5 words and maximum 200 words in a sentence. By doing so, we removed 3160 short sentences and 35 long sentences. We further cleaned ambiguous sentences (i.e. sentences which are exact matches and which have been labeled with 2 different classes). There are 82 such sentences, yielding a total of 7537 high quality informative sentences. The distribution over word length of the final, cleaned sentence set is shown in table 10.3 (Chapter 10).

**Labeling.** We employ two label approaches. One where we use human annotators. Datasets with this label approach we refer to as the *golden datasets.* Datasets from the second approach we refer to as *silver datasets.* We evaluate the silver datasets in Subsection *Silver Labels* (page 126) and the golden datasets in Subsection *Golden Labels* (page 127) using the same models and demonstrate that performance results and learning from the silver datasets carry over to the golden datasets.

**Labeling - Silver labels.** For the silver datasets we label each sentence with the information type assigned to the overall document by the lawyers at Baum, Hedlund, Aristei & Goldman. These datasets rely on very little human annotation effort and thus with same cost provide much more labeled data. We extract each of the 4 types of sensitive information from the full set of 7537 sentences. This constitutes the positive (e.g. sensitive) labeled sentences. From documents with different labels, we uniformly at random select sentences for negative sampling for each class. This resulted in the distribution of sentences across classes shown in table 10.4. The labeling approach to obtain the silver datasets provides a large number of labeled sentence using fewer resources. Clearly they may contain more noise than the golden datasets, however we demonstrate in our evaluation Subsection *Silver Labels* (page 126) that silver datasets carry intrinsic information on the sensitive information type. In particular we show that in Subsection *Golden Labels* (page 127) that models which *transfer* learning from silver to golden datasets are stronger than models relying on golden datasets alone.

**Labeling - Golden labels.** The silver labels discussed above are especially useful when sentence labels are missing or where only limited resources for la-beling are available. In particular for larger documents, though, a document which contains sensitive information may also contain non-sensitive information. E.g. an email may contain greetings or best wishes which is generally not sensitive.

**Labeling - Golden labels - Manual annotation.** For silver labels such documents may introduce noise. To study the impact of such noise in the Monsanto dataset we carry out an evaluation study where human annotators label sensitive vs non-sensitive sentences from the Monsanto dataset. These new labels are provided as an additional resource with the larger datasets containing the silver, document-based sentence labels.

We employ an iterative approach to developing annotation guidelines, with discussion with annotators when they encountered corner cases initially.

In our new golden dataset we have labels from 3 annotators. The annotators were given an introduction to the Monsanto case and the different types of sensitive information. This included an initial meeting where we presented the Monsanto Trial and discussed the various types of information which might be deemed sensitive. We discussed the 4 definitions of sensitive information (see Table 5.2). E.g. for the types of sensitive information the annotators used the descriptive text in Table 5.2 to guide the definition of sensitive information for the annotation process. For *GHOST* labels the lawyers were looking for papers where there was discussion on who to remove as authors in order to make an article appear as not being influenced by Monsanto. Also under the *GHOST* label was communication where Monsanto made effort to discredit experiments and scientists where the findings was deemed problematic by the Monsanto team. For *TOXIC* and *CHEMI* there was some discussion with the annotators how to distinguish between the two. Our response was that *TOXIC* included when Roundup is toxic. This includes when or in which circumstances Roundup could cause cellular change (e.g. cancer) and also how to treat or store Roundup to prevent (or cause) toxicity. The documents labeled *CHEMI* by Baum, Hedlund, Aristei & Goldman are focused on the chemical properties of Roundup. E.g. which components are active and how can they react with materials in the environment and which new reactive materials (molecules) might be generated under such reactions. Thus *TOXIC* is on the toxicity and *CHEMI* is on the chemistry of Roundup (and the environment in which Roundup is applied). The final type of sensitive information is *REGUL* is on the tracking and influencing of government investigations and regulation on the use of Roundup. Note that all types of sensitive information (*GHOST*, *TOXIC*, *CHEMI* and *REGUL*) has special interest if studies (controlled experiments) where properties of Roundup is under investigation, the four types differ on the aspect or property of studies with which they are concerned:

*GHOST* is concerned with ensuring support from scientist (to be signed au-thors) by funding or releasing studies.

*TOXIC* is concerned with the toxicity claims and consequences in studies

*CHEMI* is concerned with the chemical analysis and claims in studies

*REGUL* is concerned with the effect results or wordings in studies may have on governmental regulating organizations

All 4 types of sensitive information discuss whether to fund or continue studies. *GHOST* and *REGUL* are also concerned with trying to halt studies by outside researchers where the possible results are deemed problematic for Monsanto. Specifically this can happen if the outside researchers are using a different experimental protocol than Monsanto does internally. For *GHOST* trying to stop publication of a paper led Monsanto to gather support for discrediting the study in question and getting the study retracted[11].

Before the annotation process itself we also had a workshop session where we again presented the Monsanto trials and the documents from the trial. During the workshop we also presented the annotation tool (see next paragraph) developed to be used for the annotation process. Then the annotators tried out the tool and we provided feedback on expected workflow and also discussions when there was questions to the actual sentences. After this initial close collaboration the annotators left and annotated the remaining sentences offline and emailed results when done.

**Golden labels - Using the tool.** We developed a annotation tool which allow the annotator to see the description of why the current document is considered sensitive together with the current sentence and some sentences before and after. A screenshot of the tool can be found in Figure 5.1. We instructed the annotators using the tool as follows[12]:

```
1) Load the description file: click button "Labeling
 Guide" (lower left corner) and select the file:
 monsantoDataEntries.json
2) Load the data file: click the button "Load Data" (3rd
  button from the top, left side)
3) Select a user: Select dropdown "Select User" (2nd
 button from the bottom, left side)

You can click the arrow buttons left, right to move
 between sentences.
```

[11] https://retractionwatch.com/2017/08/10/unearthed-docs-monsanto-connected-campaign-retract-gmo-paper/

[12] See also https://github.com/neerbek/taboo-mon/blob/master/doc/AnnotationDescription.txt

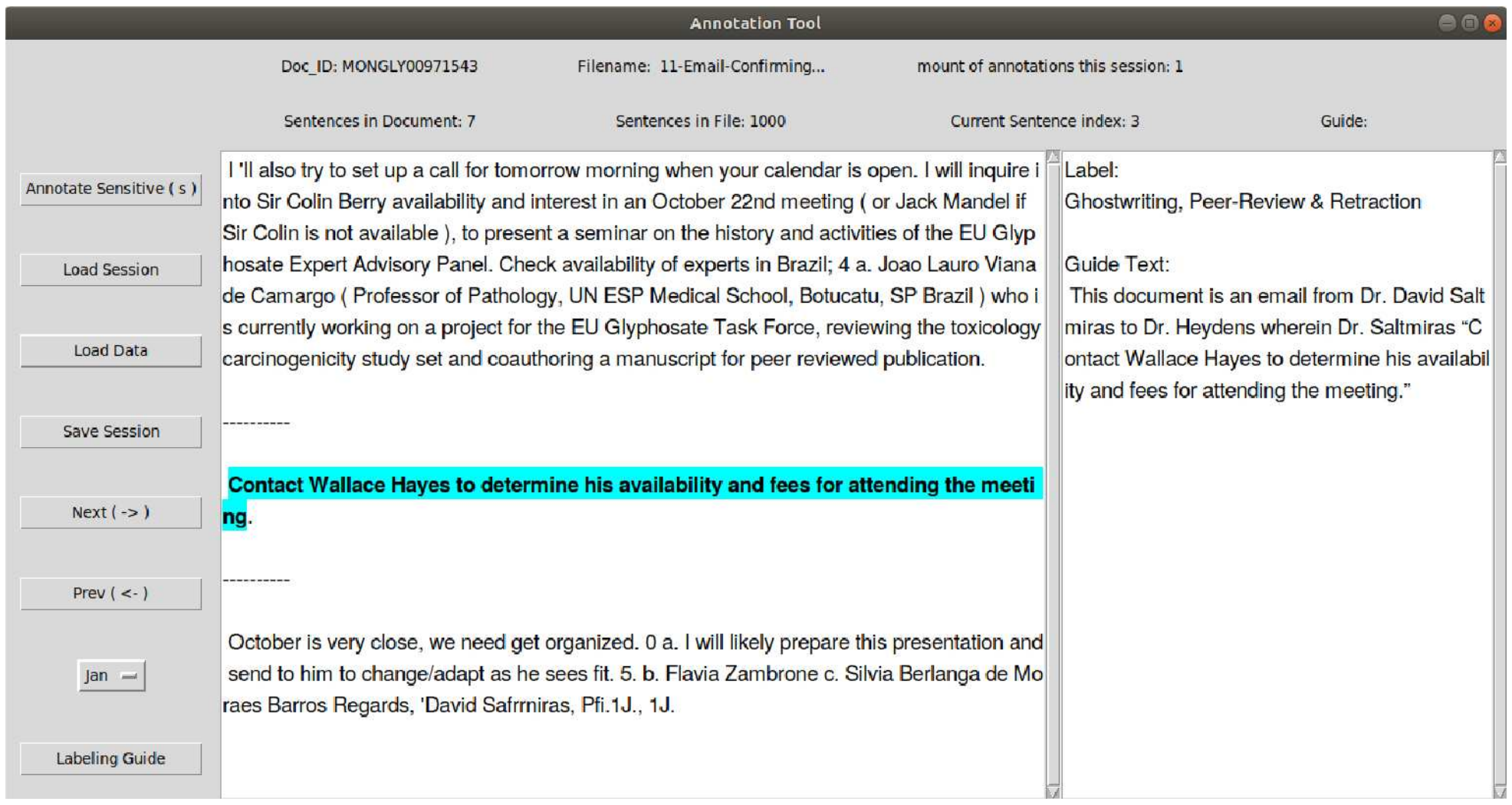

Figure 5.1: Screenshot of the annotation tool used for the golden labels. See text for a description of functionality.

```
A few previous sentences are shown before the current
 sentence to provide context together with the next few
 sentences. In the application there is a context bar to
  the right. This shows the current type of sensitive
 information which is contained in the full document and
  which we want to find sentences which contain. Below
 the label there is provided a small descriptive text
 which is written by the lawyer who labeled the document
  and describe why the document is labeled as it is.

To label sensitive:
You can highlight parts of the sentence and click the
 button "Annotate Sensitive" (or use shortcut key 's').
 You can highlight complete sentences or multiple places
  in a sentence where sensitive information is revealed.

To save annotations. Click "Save Session" (middle of
 left side) and select a filename to save to.
```

In addition, we instructed the annotators as follows:
For a given sentence to be annotated, the annotator is instructed to consider the type of sensitive information we are investigating, as defined in Table 5.2. The annotator should then consult the guide text (in the context bar in the right side of the application UI, see Figure 5.1). The annotator then has to decide whether this sentence contains expose any information about the sensitive information type given.

The annotators were allowed to use the internet/Google to provide further context. This was mainly used initially by some of the annotators to further build a background understanding of the Monsanto case and for recall the details of the case as introduced at the initial workshop.

Each annotator was given the same 1073 sentences taken at random from documents labeled by the lawyers at Baum, Hedlund, Aristei & Goldman. These 1073 sentences were distributed uniformly at random across each sensitive information type.

Each annotator then labels the sentence sensitive or not, based on the given sensitive information type.

**Labeling - Golden labels - Annotation Results.** We use majority of inter-annotator agreement i.e., assign sensitivity to sentences which at least 2 annotators have labeled sensitive.

We found that on 65.88% of all sentences, all 3 labeled the same (sensitive or not-sensitive). I.e., we applied the majority rule on the remaining 34.12% of sentences.

To judge the inter-annotator agreement a number of metrics can be used. There are the family of correlation coefficients metrics such as the Spearman rho [70], which given 2 random variables with rank values (samples of 2 annotator labels for our usecase) is calculated as the ratio between the covariance of the two variables as numerator and the product of the standard deviations of the variables as the denominator. Another family is the kappa family of metrics [6]. We use the well-known Fleiss Kappa [29] which works for more than two annotators and which tends to avoid over- or underestimation which can occur with correlation coefficients such as Spearman's rho [70].

Fleiss Kappa compares the sample probability with the probability of obtaining agreement from pure chance. The Fleiss Kappa metric takes values from negative to 1, where below zero indicates agreement no stronger than agreement by chance and 1 indicates perfect agreement. For our full set of golden labels with 3 annotators we obtain a Fleiss Kappa:

Golden labels, Fleiss Kappa: 0.33

A rule of thumb interpretation of the Fleiss Kappa score given in [62] maps our score of 0.33 to "fair agreement". While this indicates fair inter-annotator agreement, a higher score would be preferred. We note that our labels are skewed towards non-sensitive and the Fleiss Kappa score is generally lower on skewed data [6]. Calculating the Spearman rho to further characterize the inter-annotator agreement, we obtain the mean Spearman rho over all pairs of annotators to be 0.42. For Spearman rho a value of 1 indicates perfect correlation. The value for Spearman rho on our data is in line with the value for Fleiss Kappa.

### Outlook

We observe that out of the 274 released documents in the trial so-far[13] only 118 have until now been labeled by human annotators. E.g. so-far less than half of the Monsanto documents has been labeled, thus there is room for growing the Monsanto sensitive information dataset by obtaining labels for the rest of the documents. A particular interesting future research direction could be to generate an automatically labeled dataset by using the models developed in Chapter 10 and trained on the already labeled data to generate automatic labels on the unlabeled documents. These new auto-generated labeled sentences could constitute a secondary dataset with automatic inferred labels and while not of as high quality as the human labeled dataset this new proposed dataset could be helpful in generalizing and help prevent overfitting by providing additional data to train on. See also Section 6.1 for further discussion on potential future work.

## 5.3 Results Analysis

To motivate further the distinction between context-less and context-based approaches we provide here an overview of performance of the different approaches. We compare the approaches discussed in this chapter and the deep neural network approaches discussed in Chapter 4. In particular as an example of a *recurrent neural network (RNN)* we use a *Long short-term memory (LSTM)* networks and as an example of *recursive neural network* we use the RecNN approach. See Chapter 4 for details.

We provide a performance overview in two ways. One where we compare accuracy (fraction of correctly identified sentences) between the approaches and a second overview where we compare each approach to the RecNN approach. Here we consider the fraction of sentences only found by RecNN but not by other approaches. This metric tell us something about the effectiveness and overlap between RecNN and other approaches. Specifically we compare RecNN and the context-less approaches, as well as RecNN and LSTM approaches.

In Tables 5.4, 5.5, we show the accuracy percentage point increase when using context-based approaches vs context-less approaches. We compare performance on the datasets from Enron and Monsanto corpora. For an overview of these datasets see Chapter 4.2. For the Enron corpus we show performance on datasets PPAY, FAS, FCAST and EDENCE. In the row labeled "Enron (mean)" we calculate average performance over the different datasets. For the Monsanto corpus we show accuracy percentage point increases on both our silver and golden labels (see discussion Chapter 4.2). The rows GHOST (s), TOXIC (s), CHEMI (s) and REGUL (s) show results on the silver labels on datasets GHOST, TOXIC, CHEMI and REGUL, respectively. The row Monsanto (s, mean) is the mean performance gain (accuracy percentage point increase) over all silver labeled datasets. In a similar fashion, rows GHOST (g), TOXIC (g), CHEMI (g) and REGUL (g) show results on the golden labels on datasets GHOST, TOXIC, CHEMI and REGUL, respectively, and the row Monsanto (g, mean) is the mean accuracy percentage point increase over all golden labeled datasets.

[13] `http://baumhedlundlaw.com/pdf/monsanto-documents`

In the first table, Table 5.4, we show the different context-less approaches considered in this work. These are InfRule, C-san;$\alpha = 1$, C-san;$\alpha = 1.5$ and C-san;$\alpha = 2$. See Chapter 3 for further description of these approaches. Our performance measure is accuracy, i.e., the fraction of correctly identified instances by an approach. For each approach a development data subset was used to determine the ratio of sensitive to non-sensitive content which is used for setting hyper-parameters. We have followed the setting of hyper-parameters as done in [16, 84], see also Chapter 3. We have used a different development set than in Chapter 10 which yields slightly different values from Table 10.6 and Table 10.10.

Accuracy is given as usual (see also Table 1.1), for each dataset $D$ and approach the accuracy is

$$Acc = \frac{tp + tn}{|D|}$$

where $tp$ are the true-positive identified sentences, i.e., the sentences which have ground-truth sensitive and which the approach identified correctly as sensitive. Likewise, $tn$, are the true-negative sentences, i.e., the sentences which have ground-truth non-sensitive and which the approach identified correctly as non-sensitive. The size of the dataset is denoted $|D|$.

In Table 5.4, the column *Best* is, for each dataset, the best performing context-less approach among those considered here. We use the Best column to compare against our context-based approaches in Chapter 5.5.

In Table 5.5, the column RecNN contains the accuracy obtained by our best published RecNN model (Part II). The LSTM column contains the accuracies obtained with the best LSTM model. Note that these LSTM results for Enron are not found in Part II, but are added here to provide additional information on alternative machine learning models. Hyper-parameters from Monsanto silver datasets are reused. Both LSTM and RecNN (on both Enron and Monsanto) use early stopping to complete training. Here, we stop training when no increase in validation set accuracy is detected over 3 consecutive runs over the training set. The models obtained are close in performance to the more extensively trained models on the Monsanto datasets, reported in Chapter 10. The LSTM and Monsanto performance in Table 5.5 has been obtained using fixed hyper-parameters and stopping criteria to compare models given approximately the same training effort.

In Table 5.5, the column CL contains the best context-less approach (see Chapter 3) as obtained in Table 5.4.

The column $\Delta 1$ is the percentage point improvement between the RecNN model and the best CL model. The difference is measured as the number of correctly classified sensitive sentences on the test sets of the different datasets. The percentage point improvement by adopting RecNN over CL approaches, are 12.71% on Enron corpus, 9.77% on Monsanto, silver labels corpus and $-0.10\%$ on Monsanto, golden labels corpus. This indicates that RecNN performs well on complex sensitive information, in particular on our larger datasets Enron and Monsanto (silver). As discussed in Chapter 10, the RecNN and LSTM models seem to require more labeled data in order to learn more than the context-less approaches considered here. We also demonstrated in Chapter 10 that one way to obtain better accuracy for RecNN when only little labeled data is available is to consider sources of silver, semi-automated labels and using *transfer learning* to further increase the performance of RecNN.

The column $\Delta 2$ is a comparison of performance of RecNN vs LSTM. We compare the percentage point improvement between the best RecNN model and the best LSTM model in a similar fashion as for comparing RecNN and CL approaches. In both columns a positive number indicates that the RecNN approach correctly classifies more instances than the approach compared against. A negative number indicates that RecNN classifies fewer instances correctly. The mean percentage point improvement in accuracy on Enron datasets is $-0.76\%$, on Monsanto silver datasets it is 1.77%, and on Monsanto golden datasets the mean is 0.85%. We observe that where there was a large gain when moving from context-less approaches to RecNN on the larger datasets, the performance between RecNN and LSTM seems more at par, with RecNN having a small increase in accuracy on PPAY, $GHOST$ (s), $TOXIC$ (s) and $REGUL$ (s). Both RecNN and LSTM approaches are context-based and encode the sentence context as a vector of artificial network neurons outputs. The RecNN does this after a recursive structure (here constituent parse tree) whereas the LSTM does this following a recurrent pattern, see the discussion of artificial neural networks and recursive vs recurrent approaches in Section 4.1.

The second performance overview which we consider here is that of the fraction of sentences which the RecNN approach correctly identifies compared to other approaches. Take two approaches, denoted $M_1$ and $M_2$, and a particular dataset, denoted $D$. For example $M_1$ can be the RecNN approach and $M_2$ can be the InfRule approach and the dataset can be $D =$ FCAST. Then for sentences in the dataset, $s \in D$ we are interested in comparing the number of sentences where $M_1$ identifies the label correctly and where $M_2$ identifies the label correctly. I.e., the sentences on which $M_1$ (or $M_2$) predicts the correct label. We denote the sets of such sentences as follows

$$C_1 = \{s \mid s \in D \text{ and } M_1 \text{ identifies the label of } s \text{ correctly}\}$$

and

$$C_2 = \{s \mid s \in D \text{ and } M_2 \text{ identifies the label of } s \text{ correctly}\}$$

The sentences identified correctly by $M_1$ but not by $M_2$ are then given by subtracting set $C_2$ from $C_1$, i.e., $C_1 \setminus C_2$. To express this as a fraction we divide by the number of sentences identified correctly by approach $M_1$, e.g. we divide by the size of $C_1$

$$\frac{|C_1 \setminus C_2|}{|C_1|} \tag{5.1}$$

which we refer to as *the fraction of sentences only identified correctly by* $M_1$.

We start with the comparison between RecNN and the individual context-less approach in Table 5.6. Here we calculate the fraction against the context-less approaches InfRule and C-san (with various parameterizations, see Section 3.3) and RecNN compared with the intersection of correctly identified sentence over all context-less approaches (column Intersect). The fractions (given as percentages) are calculated using Equation (5.1). A high percentage means that a large fraction of the sentences identified by the RecNN approach are not identified correctly by the compared approach and thus is an indication that the RecNN approach performs well on sentences where the compared approach does not perform well. The higher the percentage the more the set of correctly identified sentences by the RecNN differs from the set of correctly identified sentences by the compared approach.

On the Enron dataset we observe that the context-less approaches seem to have similar performance. We see a mean fraction in the range 18.55% (C-san($\alpha = 1$)) to 26.47% (C-san($\alpha = 2$)). For the intersection of all approaches the fraction of sentences only identified correctly by RecNN reaches 32.05%, which indicates that the different context-less approaches correctly identify different sets of sentences. To put this in another way, the context-less approach which comes closest to identifying correctly all sentences which RecNN identifies correctly is C-san($\alpha = 1$) which identifies $100\% - 18.55 = 81.45\%$ (as a mean) of all sentences which RecNN identifies correctly, however we know that on $32.05\% - 18.55\% = 13.50\%$ of the set of correctly identified sentences (by C-san($\alpha = 1$)) at least one other context-less approach disagrees with the identification of the sentence. A similar observation can be made for Monsanto (silver), with the fraction of sentences only identified correctly by RecNN compared to the intersection of context-less approaches is 51.53%, and for Monsanto (golden) this fraction is 24.89%.

Two observations can be made based on these fairly large fractions; one, the RecNN approach finds a significant fraction of sentences which is not found by the intersection of context-less approaches, and two, we observe a high spread in the fractions over single datasets by various context-less approaches and parameterizations, which indicates that the set of sentences correctly identified by the context-less approaches is highly influenced by choice of approach and parameterizations. Furthermore, there does not seem to be a clear winner in terms of which context-less approach to use, for Enron C-san($\alpha = 1$) and InfRule are the two best, while for Monsanto (silver) the best are C-san($\alpha = 2$) and C-san($\alpha = 1.5$) and for Monsanto (golden) the best are C-san($\alpha = 1$) and C-san($\alpha = 1.5$). It appears that the best context-less approach for a given dataset needs to be determine using exhaustive search.

In Table 5.7 we list fractions of sentences only identified by RecNN when compared to LSTM and the intersection of context-less approaches (the column Intersect, copied from Table 5.6 for easy comparison). First, we observe that LSTM generally has much lower fractions than the context-less approaches. This is in line with in Table 5.5 where the two context-based approaches RecNN and LSTM both obtain high accuracies. However, we observe up to 20.08% of sentences only identified correctly by RecNN for dataset $TOXIC$ (s), and in general for Monsanto (silver) we observe higher fractions than for the other datasets (but still less than for the context-less based approaches). Comparing to Table 5.5 we observe that the mean difference in accuracy for RecNN and LSTM (column $\Delta 2$) is 1.77%. Thus, the (mean) fraction of sentences only identified by RecNN on Monsanto (silver) is 13.60% which seems to indicate that the context-based approaches correctly identifies different subsets of sentences.

In summary, we give an overview of the performance of InfRule, C-san (with 3 different parameterizations), LSTM and RecNN approaches on our 12 datasets. We consider both accuracy as the fraction of correctly identified sentences and, the fraction of sentences only correctly identified by RecNN as performance metrics. While there is some variance in values, there is a clear trend in that context-based approaches detect larger and different subsets of sentences correctly than the context-less approaches. These findings are also supported by the discussion in the coming chapters and the published works in Part II.

### Examples

Here we provide some examples to complement the analysis in the previous section. We begin with PPAY, which is a dataset with sensitive information type concerning financial so-call "prepay" transactions, see also Section 5.1 and Table 5.1. Here is an true positive example with sensitive content which both RecNN and LSTM identify correctly

```
Subject:  Re:  Expedited DASH Process for Prepaid and
other Embedded Financings in Commodity Transactions.
```

Note that this example contains the word "prepaid" which seems likely to be indicative of the PPAY sensitive information type.

A true negative example on PPAY without sensitive content and which both RecNN and LSTM identify correctly is the following:

```
The bulk of actual bid mitigation, since market
operations began, has taken place with respect to bids
from generators in transmission congested areas.
```

| Dataset | InfRule | C-san $\alpha = 1$ | C-san $\alpha = 1.5$ | C-san $\alpha = 2$ | Best |
|---|---|---|---|---|---|
| Enron | | | | | |
| PPAY | **90.37**% | 90.30% | 90.18% | 88.65% | **90.37**% |
| FAS | **59.75**% | **59.75**% | 59.59% | 59.59% | **59.75**% |
| FCAST | 75.00% | **79.10**% | 63.49% | 63.49% | **79.10**% |
| EDENCE | 82.22% | **83.71**% | 80.99% | 73.42% | **83.71**% |
| Enron (mean) | 76.84% | 78.22% | 73.56% | 71.29% | **78.23**% |
| Monsanto (silver) | | | | | |
| *GHOST* (s) | 61.09% | 52.26% | 65.60% | **71.99**% | **71.99**% |
| *TOXIC* (s) | 53.69% | 50.00% | 61.08% | **62.78**% | **62.78**% |
| *CHEMI* (s) | 59.93% | 56.95% | 68.54% | **72.52**% | **72.52**% |
| *REGUL* (s) | 60.74% | 61.07% | 69.46% | **72.82**% | **72.82**% |
| Monsanto (s, mean) | 58.86% | 55.07% | 66.17% | **70.03**% | **70.03**% |
| Monsanto (gold) | | | | | |
| *GHOST* (g) | 63.33% | **77.78**% | 75.56% | 75.56% | **77.78**% |
| *TOXIC* (g) | 64.18% | **73.58**% | 69.81% | 69.81% | **73.58**% |
| *CHEMI* (g) | 77.27% | 83.33% | **84.85**% | **84.85**% | **84.85**% |
| *REGUL* (g) | 80.60% | **91.04**% | **91.04**% | **91.04**% | **91.04**% |
| Monsanto (g, mean) | 71.34% | 81.43% | 80.32% | 80.32% | **81.81**% |

Table 5.4: Summary test accuracies of the context-less approaches.

While this sentence appears to be about financial transactions and terms such as “bid mitigation” are associated with financial markets, the context seems to suffice to determine the sentence as non-sensitive.

By contrast we here give an example from PPAY without sensitive content which RecNN identifies correctly,but which LSTM does not

```
(4) Other Charges for US Service and Canadian Service:
Buyer shall be liable for the payment of invoices from
the railroad for demurrage and hazardous materials
storage charges incurred by ECFC as the prepaid shipper
due to Buyers inability to receive a rail car and/or have
a rail car placed on Buyers siding.
```

This constitutes an interesting case because it contains the term “prepaid” as in the first example. Still, RecNN successfully determines from the context that this is not sensitive. Similar worded sentences are likely to be found in Enron contracts. It appears for this example having access to additional structure such as constituency parse trees, as the RecNN has, provides additional

| Dataset | RecNN | LSTM | CL | Δ1 | Δ2 |
|---|---|---|---|---|---|
| Enron | | | | | |
| PPAY | **96.30%** | 95.94% | 90.37% | 5.93% | 0.36% |
| FAS | 95.89% | **96.61%** | 59.75% | 36.14% | −0.72% |
| FCAST | 83.01% | **84.96%** | 79.10% | 3.91% | −1.95% |
| EDENCE | 88.56% | **89.30%** | 83.71% | 4.85% | −0.74% |
| Enron (mean) | 90.94% | **91.70%** | 78.23% | 12.71% | −0.76% |
| Monsanto (silver) | | | | | |
| *GHOST* (s) | **83.27%** | 82.14% | 71.99% | 11.28% | 1.13% |
| *TOXIC* (s) | **73.58%** | 73.01% | 62.78% | 10.80% | 0.57% |
| *CHEMI* (s) | **80.46%** | **80.46%** | 72.52% | 7.94% | 0.00% |
| *REGUL* (s) | **81.88%** | 76.51% | 72.82% | 9.06% | 5.37% |
| Monsanto (s, mean) | **79.80%** | 78.03% | 70.03% | 9.77% | 1.77% |
| Monsanto (gold) | | | | | |
| *GHOST* (g) | **77.78%** | **77.78%** | **77.78%** | 0.00% | 0.00% |
| *TOXIC* (g) | 71.70% | 69.81% | **73.58%** | −1.88% | 1.89% |
| *CHEMI* (g) | **84.85%** | **84.85%** | **84.85%** | 0.00% | 0.00% |
| *REGUL* (g) | **91.04%** | **91.04%** | **91.04%** | 1.50% | 1.50% |
| Monsanto (g, mean) | 81.34% | 80.87% | **81.81%** | −0.10% | 0.85% |

Table 5.5: Summary test accuracies of the best context-less (CL) vs the context-based approaches. Each column contains the best model obtained with that approach. RecNN on Monsanto golden datasets includes the transfer model RecNN-tf which was demonstrated to work well when we have combinations of silver and golden labels (see Subsection *Results - 4. Case: Overview on Golden* (page 131)). Column Δ1 is the difference in accuracy between RecNN and the best context-less (CL) approach. Column Δ2 is the difference in accuracy between RecNN and LSTM approach. In both columns a positive value indicates that RecNN has higher accuracy.

| Dataset | InfRule | C-san $\alpha = 1$ | C-san $\alpha = 1.5$ | C-san $\alpha = 2$ | Intersect |
|---|---|---|---|---|---|
| Enron | | | | | |
| PPAY | 7.93% | 7.93% | 8.14% | 9.73% | 13.44% |
| FAS | 40.18% | 40.18% | 40.35% | 40.35% | 40.35% |
| FCAST | 19.24% | 13.53% | 33.70% | 33.70% | 41.15% |
| EDENCE | 12.71% | 12.54% | 13.43% | 22.10% | 33.25% |
| Enron (mean) | 20.02% | 18.55% | 23.91% | 26.47% | 32.05% |
| Monsanto (silver) | | | | | |
| *GHOST* (s) | 35.67% | 46.28% | 29.12% | 22.35% | 50.79% |
| *TOXIC* (s) | 44.79% | 50.19% | 35.91% | 32.82% | 62.93% |
| *CHEMI* (s) | 37.45% | 38.68% | 25.93% | 20.99% | 48.56% |
| *REGUL* (s) | 34.84% | 33.61% | 22.95% | 20.08% | 43.85% |
| Monsanto (s, mean) | 38.18% | 42.19% | 28.48% | 24.06% | 51.53% |
| Monsanto (gold) | | | | | |
| *GHOST* (g) | 34.29% | 4.29% | 18.57% | 18.57% | 41.43% |
| *TOXIC* (g) | 26.32% | 5.26% | 2.63% | 18.42% | 28.95% |
| *CHEMI* (g) | 16.07% | 1.79% | 0.00% | 0.00% | 16.07% |
| *REGUL* (g) | 13.11% | 0.00% | 0.00% | 0.00% | 13.11% |
| Monsanto (g, mean) | 22.45% | 2.83% | 5.30% | 9.25% | 24.89% |

Table 5.6: The fraction of sentences only identified by RecNN, as given by Equation (5.1). Last column *Intersect* is the fraction of correctly identified sentences by RecNN when compared to the intersection of correctly identified sentences by all context-less approaches.

prediction power over the more linear structure considered by the LSTM.

From PPAY, an example with sensitive content which the intersection of context-less approaches *Intersect* all identify correctly is:

```
NN at Macleod Dixon sent us a fax requesing copies of the
documents on the prepay.
```

This sentence contains the word “prepay” and is identified correctly by RecNN and also the context-less approaches. Note the spelling error is from the original text.

From PPAY an example with sensitive content, identified by RecNN but not by Intersect is:

```
We intend to fit an actuarial model to predict option
exercises, and build some useful rules of thumb when
thinking about the future.
```

| Dataset | LSTM | Intersect |
|---|---|---|
| Enron | | |
| PPAY | 2.42% | 13.44% |
| FAS | 1.44% | 40.35% |
| FCAST | 7.87% | 41.15% |
| EDENCE | 5.51% | 33.25% |
| Enron (mean) | 4.31% | 32.05% |
| Monsanto (silver) | | |
| *GHOST* (s) | 9.71% | 50.79% |
| *TOXIC* (s) | 20.08% | 62.93% |
| *CHEMI* (s) | 11.11% | 48.56% |
| *REGUL* (s) | 13.52% | 43.85% |
| Monsanto (s, mean) | 13.60% | 51.53% |
| Monsanto (gold) | | |
| *GHOST* (g) | 10.00% | 41.43% |
| *TOXIC* (g) | 2.63% | 28.95% |
| *CHEMI* (g) | 0.00% | 16.07% |
| *REGUL* (g) | 0.00% | 13.11% |
| Monsanto (g, mean) | 3.16% | 24.89% |

Table 5.7: The fraction of sentences only identified by RecNN, as given by Equation (5.1). Here we compare against the LSTM approach (LSTM column) and by the intersection of the context-less approaches from Table 5.6 (Intersect column).

This sentence is about valuation (predicting value of something in the future), which makes it plausible as a discussion of prepay transactions. The Intersect approach identifies (incorrectly) this as non-sensitive which implies that $n$-grams in the sentence are assigned low sensitivity scores. This is an example of a paraphrasing that avoids "typical" keywords that would allow easy identification.

From the *CHEMI* dataset from the Monsanto corpus (see Section 5.2), with silver labels, an example is:

```
Ideally, he would be willing to discuss EPA assessment
of glyphosate and conclusion it is not carcinogenic b.
Minimally, he would participate and point out elements of
the IARC process that fall short of more in depth reviews
by regulators
```

Note that this text has been captured as a single sentence by the NLTK sentence splitter in our tool-chain (see Chapter 4.2). It appears to be an itemlist on the form "a. Ideally... b. Minimally..." where the first "a." has been split away by the sentence splitter. The sentence is not sensitive in terms of the CHEMI type of sensitive information, since it discusses EPA assessments and not the chemistry of glyphosate itself. Both context-based approaches RecNN and LSTM correctly identify this example as non-sensitive, however Intersect identifies this example as sensitive. By considering the context of the use of the term glyphosate, the RecNN and LSTM approaches can identify this sentence as being non-sensitive, whereas the $n$-grams in sentence are associated with the sensitive label.

From *GHOST* dataset with golden labels, where the use of ghost-writers and controlling publications is labeled sensitive, see Section 5.2, an example of a non-sensitive sentence identified as sensitive by both LSTM and Intersect, but where RecNN correctly identifies the sentence as non-sensitive:

```
I wanted to 'tee up ' an idea for a near term
communication, potentially via Forbes or another
appropriate media channel.
```

The sentence is about a meeting for discussing a publication, but it is not about assigning authorship or ghost-writing. So the domain is correct (writing publications) but this is not sensitive with regards to the *GHOST* (g) sensitive information type.

## 5.4 Our Contributions

A key challenge in evaluating and developing new approaches for sensitive information detection is the lack of good, public datasets with real-world examples of both sensitive and non-sensitive information. As discussed, while some datasets has been previously released, none of these datasets fulfill the criteria as a true evaluation dataset for sensitive information detection. Due to the inherent sensitive nature of the data publicly available real-world data including sensitive information is not easy to obtain.

To address this need we release publicly datasets with labels over the Enron and Monsanto document sets. These two datasets with total 8 different definitions of sensitive information types provide a novel and strong set of data for sensitive information detection. We highlight the following strong properties of our released datasets which form part of our contributions

- More than $750,000$ labeled texts.
- Both data and labels are humanly generated. I.e this is real-world data.
- Parse trees for all texts.
- 8 different types of sensitive information. The types represent very different kinds of information ranging from chemical texts to financial forecasts.

- Sensitive and non-sensitive texts come from same data and are very similar in structure
- Large datasets of unlabeled data is available which can be used for training unsupervised approaches and which potentially can be labeled in the future in order to grow the datasets even more.

All 8 different types of sensitive information are presented as binary labels: sensitive vs. non-sensitive. In the case of Enron, the original TREC labels are per type of information such that the same sentence can be labeled as sensitive (or non-sensitive) with respect to 2 different types of sensitive information. For Monsanto papers the original labels provided by lawyers are disjunctive and thus a sentence is only labeled sensitive with respect to one type of sensitive information. This should make it easier to evaluate and compare different sensitive information detection approaches. Our experiments show that the sensitive information types in both datasets contain at least some context-based sensitive information and the datasets may therefore serve as basis for future investigations into the differences of keyword-based and context-based approaches for sensitive information detection.

Furthermore we provide ready-to-use parse-trees for all sentences in the datasets. We found in our approaches that structure in the sentences helps in detecting sensitive information and the parse-trees makes it easy for others to evaluate structured approaches over the datasets.

It is our hope that these processed datasets and the provided labels will allow fellow researchers to advance the domain of sensitive information detection even further.

Resources available:

- General code including the parse tree parser: `https://github.com/neerbek/taboo-core`
- Monsanto specific code: `https://github.com/neerbek/taboo-mon`
- Selective Training code (included for completeness): `https://github.com/neerbek/taboo-selective`
- Enron labeled sentences with parse-trees: `https://dataverse.harvard.edu/dataverse/enron-w-trees`
- Monsanto labeled sentences with parse-trees: `https://dataverse.harvard.edu/dataverse/monsanto-w-trees`

# Chapter 6

# Outlook

The capabilities of computers to automatically extract information and content from unstructured text documents has increased rapidly in recent years. This is also true in the sensitive information detection domain. In this work we introduce new deep learning approaches such as recurrent and recursive neural networks to the problem of detecting sensitive information in natural language sentences. Experimentally we show in Part II there is an immediate jump in detection accuracy from the adaption of deep learning methods to the sensitive information detection problem. The new approaches allow us to detect new types of sensitive information, beyond the capabilities of the previous state-of-the-art keyword-based approaches.

We argue that considering the context of textual information is the key to detecting the more general and complex forms of sensitive information. We demonstrate that recursive and to some extend recurrent models do in fact capture some of the important context. However, there are still many future directions that might be fruitful to consider and which may lead to novel and competing approaches for detecting sensitive information. Directions and approaches which we just haven't discovered yet.

In this final chapter of Part I we consider the possible directions of future work in Section 6.1 and discuss the contributions in this work in Section 6.2.

## 6.1 Future work

Building on the work in this dissertation there are many possible areas and directions which present interesting research directions in themselves and also, potentially, completely new approaches with novel properties and potential for detection of new, advanced types of sensitive information. Better approaches for understanding context and informational content would of course benefit the sensitive information detection domain, but also the general domain of broad natural language processing as these are also important problems to tackle for the general field. Here we list some of the areas and directions which we find to be of particular interest and potential.

- We demonstrate good results by including structure in the form of a constituency tree rather than just viewing documents as sequences. An interesting direction to consider is if we can stack several structures on top of each other, i.e. similar to stacking LSTM to obtain a bi-directional LSTM. Stacked sequential and convolutional models has shown promising results for other problems such as universal sentence representations [21].

- Another interesting question would be to consider further and simpler models that require less training data or computation resources. While the focus in this work has been the creation of state-of-the-art performance for complex sensitive information detection using advanced machine learning models, another direction could be to conduct a broad comparative study into the differences among simpler models. Such a study could be used to characterize the differences between simple and complex sensitive information detection further.

- The selective algorithm for selective training presented in Chapter 9 is a good start but we might be able to create better selective algorithms. In particular it might be possible to create novel variants of the measure $\Delta MFO$, which might be able to take into account other cluster measures than the K-means one developed in Chapter 9. This in turn might lead to even faster training and better accuracies. One direction here could be to consider the spread or variance of a cluster as part of the improved measure. Intuitively a cluster with high spread of members from centroid contain at least some interesting information and should not be pruned away even if they have high $\Delta MFO$.

- The inverse shortcuts counter the vanishing gradient problem in our experiments [74–76], see also Part II. The inverse shortcuts does so by strengthening the error signal during the backpropagation step and thus helps our approaches learn on large structures. An interesting future direction could be to compare with the regular forward shortcuts approach and evaluate on several different types of problems, e.g. topic classification, sentiment classification and sensitive information detection. This could lead to insights of which approach is better on which problems. Combining the two approaches for stronger signals on very large structures would very interesting. This could lead to new general models which allow for longer sequences of neural layers before vanishing gradients become a problem.

- In a similar vein it might be possible to view recursive neural networks as a form of shortcuts themselves when the recursive structure is rele-vant to problem. Let the input be $d = (w_1, \ldots, w_{n_d})$ and a recursive structure over the input be $\mathcal{T}$. Assume the input has two words ($w_i$ and $w_j$) which are far apart $i \ll j$ in the document, while the words $w_i$ and $w_j$ are relevant together. Let the length of the shortest path between $w_i$ and $w_j$ in $\mathcal{T}$ be $\ell$. Then the structure $\mathcal{T}$ provides a shortcut between $w_i$ and $w_j$ if $\ell \ll |j - i|$. If a structure $\mathcal{T}$ (or maybe a set of structures) provides shortcuts for all relevant pairs of words then we might obtain an easier problem-encoding than the original input document. Whereas regular shortcuts or inverse shortcuts provides 1 shortcut per step, the structure $\mathcal{T}$ could be viewed as providing a distance-based set of shortcuts. Interesting topics to investigate could be; 1) Can we learn such a structure unsupervised?, 2) Does the multi-dimensional shortcuts stack with regular and inverse shortcuts, i.e. does it lead to better performance?

Some interesting directions to work on with respect to the datasets we publish

- Our released Monsanto datasets contain both silver and golden sensitive information definitions. This allows for extended empirically evaluations as in Chapter 10. It could be interesting to conduct a similar study for the Enron datasets also. Furthermore it would be interesting to compare with the performance of our current approaches on finely labeled datasets. Obtaining such a finely labeled dataset would require substantial domain expert manual labeling effort and it is not clear if it is always meaningful to carry domain specific definitions of sensitive information all the way to sub-phrase level. In such case we may require a set of human annotators and then the label may become a distribution of labels instead.

- Another interesting direction could be to consider large-scale additional data labeled semi-automatically over the Enron and Monsanto datasets. While we in this work presented innovative labeled versions of these datasets with gold labels provided by human annotators, an interesting future work would be to consider a set of silver labels based on current best models for sensitive information detection. These models are set to label all unlabeled data and the generated labels are then applied as (silver) ground-truth labels for this additional set of documents. This would increase publicly available datasets with labeled sensitive information by an order of magnitude, similarly what authors did in [61] for paraphrase detection on Twitter data. For sensitive information detection our models' performance on hold-out labeled tests data can be used to predict performance on unlabeled data and thus the quality of the silver labels generated. Interesting followup questions to consider are whether models trained on silver labeled data improves and how much the added noise from the silver labels impact performance.

## 6.2 Contributions

In this work we focus on the problem of sensitive information detection in unstructured data. That is, the challenge of finding sentences containing sensitive information. We discuss various challenges within this domain and capture the problem of sensitive information finding in a number of definitions. Below we list our contributions and we note the main research questions which contribution addresses. The research questions, [RQ1],..., [RQ3], were posed in Section 1.1. Our main contributions includes:

- The insight that paraphrase detection and sensitive information detection are related in terms of the types of informational or textual entailment that they consider. This new connection enable us to:

- Adapt state-of-the-art approaches for paraphrase detection to the domain of sensitive information detection. This was directly aligned with [RQ1] because this contribution motivated the use of constituency parse trees to encapsulate context for complex sensitive information detection

- Specifically we adapted recursive neural networks for paraphrase detection to sensitive information detection. This was directly aligned with [RQ1] because this contribution provided state-of-the-art ML (RecNN) methods for complex sensitive information detection

- We introduce the concept of inverse shortcuts where we use early prediction of label at each node in recursive structures. We demonstrate that these inverse shortcuts can be used for training recursive neural models in a meaningful and effective way. We demonstrate that we can train complex RecNN models using only sentence, rather than phrase based labels, thus this contributes to [RQ2] as it provided more efficient training methods.

- A novel approach to more effective training through selective training only on interesting examples. We show that our $\Delta MFO$ measure can be used to achieve several times faster model convergence without sacrificing predictive performance. This contributes to [RQ2] as this allow us to train deep neural networks using efficiently using fewer computational resources.

- Publication of 12, human labeled sensitive information detection problems over two distinct large document datasets In total we publish more than $750,000$ labeled sentences together with constituency parse trees. This contribute to [RQ3] as this is large set of available sensitive information types which can be used for benchmarking.

- Our published datasets are the first ever large scale, public sensitive information datasets. Our datasets are all text from real-world settings.

Furthermore our datasets contain both sensitive and non-sensitive information, whereas other previously published datasets have been artificially combined using very different types of documents in order to obtain a sensitive information dataset. See the discussion in Chapter 4.2. While our documents and the labels were previously available separately on the web, they had never been put together in this way before and the labeled sentences and parse trees had not been published before. Furthermore we provide 4 human annotated golden datasets with sentence level sensitive information annotation. We therefore provide the sensitive information detection community with access to much needed, new, readily available, real-world training and testing data to benchmark and further the algorithmic development and understanding for sensitive information detection. This contributed directly to [RQ3] as this provides the research community with publicly available benchmark data as required to be able to benchmark different sensitive information detection approaches.

# Part II

# Publications

# Chapter 7

# Detecting Complex Sensitive Information via Phrase Structure in Recursive Neural Networks

Jan Neerbek[1,2] Ira Assent[1] Peter Dolog[3]
[1] Department of Computer Science, Aarhus University, Aarhus, Denmark
{jan.neerbek,ira}@cs.au.dk
[2] Alexandra Institute, Aarhus, Denmark
[3] Department of Computer Science, Aalborg University, Aalborg, Denmark
dolog@cs.aau.dk

**Abstract:** State-of-the-art sensitive information detection in unstructured data relies on the frequency of co-occurrence of keywords with sensitive seed words. In practice, however, this may fail to detect more complex patterns of sensitive information. In this work, we propose learning phrase structures that separate sensitive from non-sensitive documents in recursive neural networks. Our evaluation on real data with human labeled sensitive content shows that our new approach outperforms existing keyword based strategies.[1]

[1]Minor changes to the published version. In particular, an error in the formula for $F1$ score has been fixed. Non-pretrained embeddings of size 10 have been removed. All reported trends and conclusions remain the same.

## 7.1 Introduction

Detecting sensitive information in unstructured data is crucial for data leak prevention. State-of-the-art approaches are based on *defining keywords* [10, 16, 38, 42, 84], i.e., assume that the sensitive topic is described in full by a small set of keywords. While effective for simple sensitive topics as in named entity recognition [16] and personal identifiable information [84], they ignore context, i.e., the way in which people describe sensitive topics in natural language phrases. As a result, they may fail to report *complex sensitive information* or report false positives.

Concretely, complex sensitive information is characterized by the fact that words are sensitive or not sensitive depending on their context. For example, describing sensitive financial transactions might use the same vocabulary as in the non-sensitive case, but using different expressions in natural language.

In this work, we therefore propose to extract the phrase structure from sensitive information to learn these expressions, and create a recursive neural network model (RNN) [46, 88] that uses phrases to predict the sensitivity of documents. We suggest a training approach for the RNN that requires only document level sensitivity information and thereby does not require labeling individual sentences, phrases, or even words. Such fine-grained labels are required for existing backpropagation-through-structure training, but are generally not available in practice. Our evaluation on real sensitive content with humanly curated labels demonstrates superior detection accuracy compared to state-of-the-art keyword-based approaches. We furthermore show that by boosting relative importance of incorrectly predicted samples we can increase sensitive detection accuracy - in the extreme at the expense of an increase in false prediction rate. This adds flexibility to our model and allows for domain-based adjustment between prediction accuracy and end-user confidence in detected samples.
Our contributions include:

- Introducing and analyzing complex sensitive information detection
- A new RNN model based on representations of multiword structured phrases
- Training of our RNN model on document labels alone

## 7.2 Complex Sensitive Information Detection

We assume a corpus of documents $D = \{d_1, d_2, \ldots, d_m\}$, where each document is a sequence of words $d = (w_1, w_2, \ldots, w_{n_d})$ such that $w_i \in V$ and $d \in D$, and training labels $L : D \rightarrow \{0, 1\}$, where 0 means non-sensitive, and 1 means there is (some) sensitive information.

Note that the problem is asymmetric in that non-sensitive documents are known to be completely non-sensitive, whereas sensitive documents may only contain very little sensitive information. Thus, the problem is recall-oriented, focusing on finding the pieces of sensitive information [10].

In existing work, each word $w$ is assigned a sensitivity score $sen(w)$, without considering its context of use. In this paper, however, we differentiate based on how the word is used, i.e., the sensitivity of a word is conditional on its context $d$, the sequence of words in which it occurs: $sen(w|d)$.

**Definition 4** *If for all words $w$ and pairs of documents $d, d'$, we have*

$$sen(w|d) = sen(w|d')$$

*then sensitivity is* context-less*.*

*Conversely, if there exists a word $w'$ and a pair of documents $d, d'$ such that $sen(w'|d) \neq sen(w'|d')$, then sensitivity is* context-based*.*

Please note that the definition reflects the asymmetry in sensitive information detection as discussed above.

## 7.3 SPR - Sensitive Phrase based RNN model

Existing sensitive information detection approaches count co-occurrence of a keyword, or small set of keywords, with other words in the text. Co-occurrence is then taken as an indication of sensitivity. This works well e.g. for topics like HIV where co-occurrence with terms such as AIDS is easily detected.

However, as we argue here, more complex topics, as in intricate financial transactions, require models that can capture context. Sentences can have arbitrary length and structure, which our model should be able to process. We therefore encode the context of *phrase structures*, which are semantic substructures in the text extracted through *constituency parse-trees*. Constituency parse-trees structure a text into constituents, i.e., compositionality through sub-phrases [90]. Creating phrase structure embeddings in a recursive manner, we generate an encoding of the entire context. Varying sizes of context are encoded by iterating in a large structure. Our SPR (Sensitive Phrase RNN) model thereby captures the complexity in sensitive information detection in natural language.

### Phrase Structure

While sentences obviously can be interpreted as sequences, this does not reflect the way in which humans understand them. Consider the sentence

`We may have to move to cash margining if necessary.`

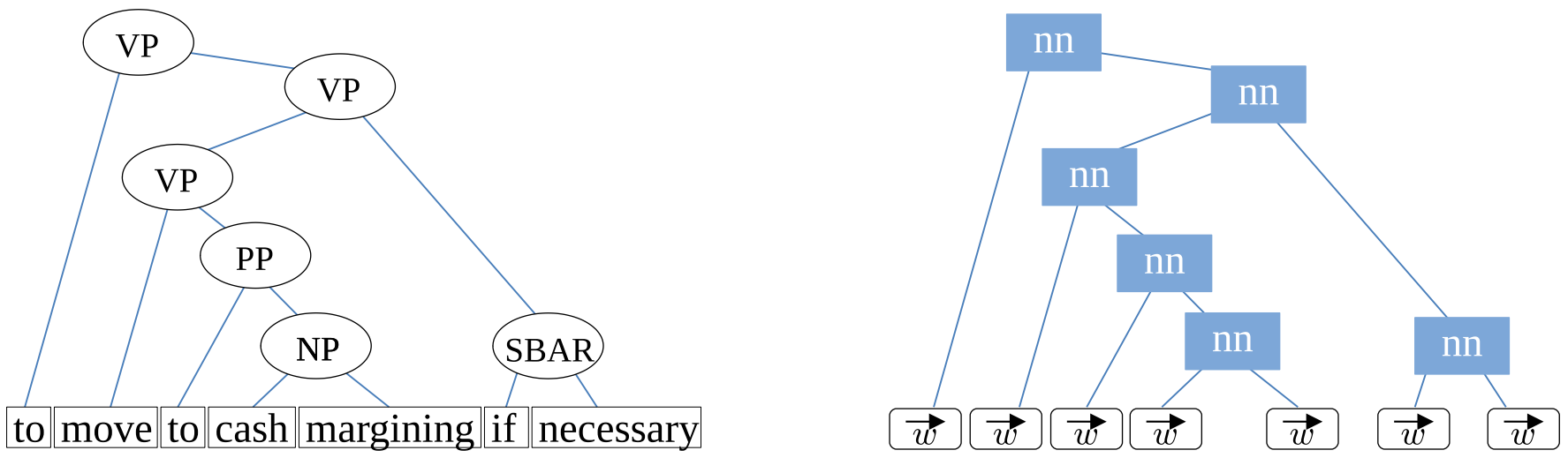

Figure 7.1: Constituency parse-tree (left). Node labels are syntactic tags, e.g., *NP* Noun Phrase, *VP* Verb Phrase; cf. Penn Treebank [95]; note *SPR* relies on phrase structure alone, not the particular label; corresponding RNN (right).

It begins with a sequence of common words, "We may have to move to", that could appear in many different contexts. The particular context for this sentence becomes clearer from the words "cash margining". In sequence order, the state after processing "We may have to move to" would be a general state, whereas given "cash margining" first, we expect a more specific hidden state. The latter order also reflects the grammatical structure in natural language.

We therefore propose to learn from sentences following their grammatical structure instead. In Natural Language Processing, this structure is captured in constituency parse-trees. Their leaves correspond to words, and nodes to sub-phrases, as illustrated in Fig. 7.1. A depth-first descent from the root node visits the words of the sentence in order. Constituency parse-trees can be automatically generated in high quality for most languages [87]. Using phrase structures as features we can successfully capture the context of words in the way they are used in natural language. On the other hand, however, phrase trees provide features which can be of arbitrary size and structure. We therefore propose to build a model that can handle this variable input by recursively taking in parts of the input. Concretely, we build a recursive neural network structure.

### Recursive Neural Networks with Phrase Structure

Our approach is to create embeddings of the complex structure in phrase trees using neural networks. To handle input of arbitrary size we propose building a *recursive neural network (RNN)* [89]. In a nutshell, a RNN recursively takes in a new part of the input from the input structure. It includes output from the previous step in the current step by applying the same architecture repeatedly. In this manner, recursive neural networks are capable of transferring knowledge between the steps, and of taking in complex input structures.

Given a constituency parse-tree we process sentences in grammatical order using the RNN recursively in a bottom-up fashion, where in each step we process a new node in the parse-tree, ending at the root node. Processing phrase

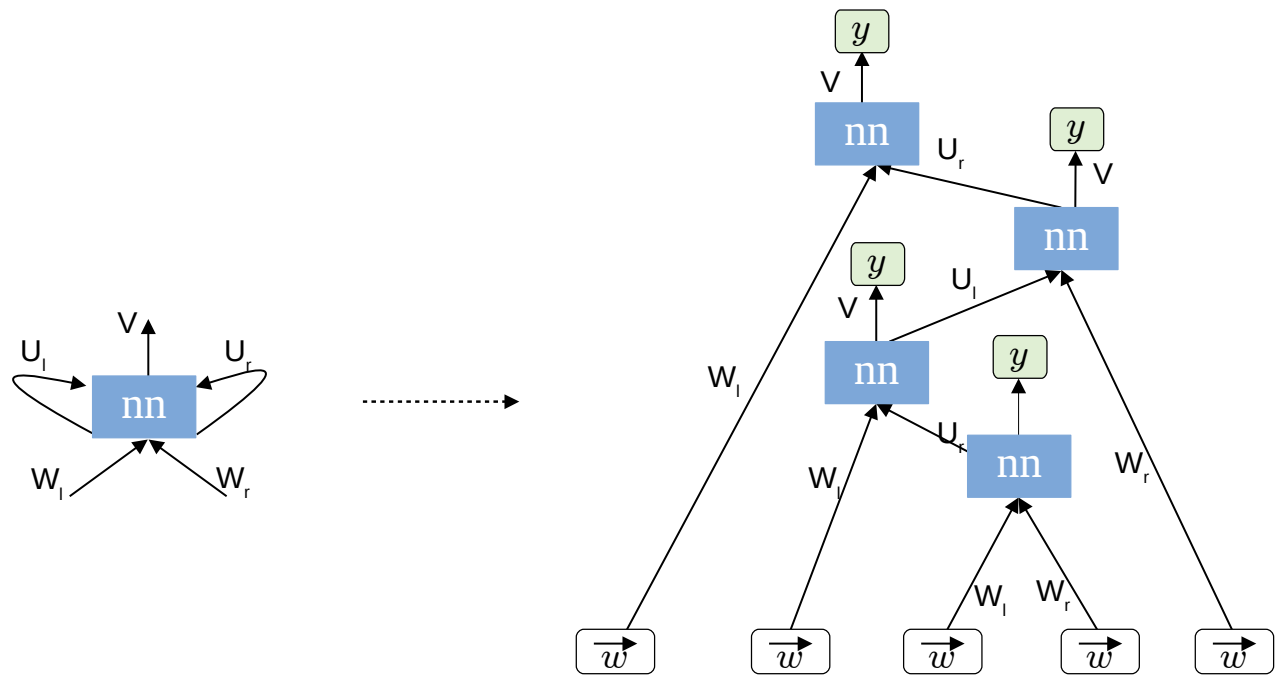

Figure 7.2: RNN; on the right unfolded in structure. A node has 4 possible inputs of which 2 are active at a time. For readability, not all output edges are labeled $V$

trees through RNNs, we automatically learn what the relevant structures in language are for the complex sensitive information detection problem. We do not need to define the size or structure of the context a-priori, which makes this a flexible and easily applicable model.

As illustrated in Fig. 7.2 (left), the recursive neural network is a function that for each node in the constituency parse-tree is evaluated with the representations of its children as input to generate its output representation (as shown in the unfolded in structure illustration in Fig. 7.2 (right)). For a node $n$ with child nodes $n_\ell$ (left) and $n_r$ (right)

$$rep_n = RNN(rep_\ell; rep_r) = \sigma[W_\ell rep_\ell + W_r rep_r + b] \tag{7.1}$$

where $\sigma$ is some non-linear activation function and $rep_x$ is the generated representation of node $n_x$ (for $x \in \{n, \ell, r\}$). Matrices $W_\ell, W_r$ weigh the different inputs, $b$ is an additive bias vector.

Representations for words are usually word embeddings [46], however, (7.1) assumes that word representations (leaves) have same number of dimensions as structure representations (nodes). Thus, it might be beneficial to consider words and structure as two different learning problems. Fortunately, this can be done with a minor adjustment to the equation above. We define a transformation function $t$ for child nodes $n_x$ as

$$t(rep_x) = \begin{cases} U_x rep_x + b_U & n_x \text{ is node} \\ W_x rep_x + b_W & n_x \text{ is leaf} \end{cases} \tag{7.2}$$

with $n_x \in \{n_\ell, n_r\}$. We can then rewrite (7.1) as

$$rep_n = RNN(rep_\ell; rep_r) = \sigma[(t(rep_\ell) + t(rep_r)]$$

The decoupling in (7.2) gives us freedom to select optimal word embedding size and hidden representation size [46].

The output of a node in the tree is

$$o_n = \sigma(Vrep_n + b_p),$$

a prediction (sensitive/non-sensitive) based on the current node representation. Thus for internal nodes in the constituency parse-tree, we obtain our prediction having only seen part of the structure so far.

### Training SPR

For training, we use the *unfolded* network, i.e., copies of the RNN for each node in the constituency parse-tree as illustrated to the right in Fig. 7.2. We may backpropagate errors through the tree in a top-down order and aggregate the errors across all copies of the neural network. This can be viewed as a *backpropagation-through-structure (BPTS)* [35] approach.

Our SPR makes use of the well established *softmax* activation function $\sigma(x)[i] = e^{x[i]}/(\sum_j e^{x[j]})$ which provides a differentiable *soft* max of the output. We expect only one answer to be true and therefore maximize (softly) the best probability as the output. Using the softmax function allows us to interpret the outcome as probabilities.

The prediction $y_n$ of SPR is given as $y_n = \text{argmax}_{x\in\{0,1\}}\, o_n[x]$ where $o_n[x]$ can be seen as probability that the input is $x$, with $x \in \{0, 1\}$.

Different sensitive domains require different focus on the model's ability to correctly predict sensitive vs. non-sensitive information. We model this using relative weighing of the two types of information in the loss function (Different weights are studied in the experiments). Adding everything together and using cross-entropy for error function, we obtain our loss function

$$L = -\sum_{n=1}^{N} \left[ w t_n \log(p_n) + (1 - t_n)\log(1 - p_n) \right] \tag{7.3}$$

where $w$ is the weighting hyper-parameter. If $w > 1$ then we weigh sensitive loss higher than non-sensitive loss.

Using the ground truth labels and BPTS we can learn our parameters $\theta$ for maximal likelihood given our data. However, supervised BPTS as defined in [35], requires a label for each node in the tree, which is not available and would be difficult to obtain, as this would require assigning sensitivity scores to phrases of increasing complexity. Still we can solve this challenge by propagating labels from the root to the internal nodes of the tree. We can then view $SPR$ as learning to assign probabilities to each node wrt. being sensitive or not. That is, if we have the same sentence $s$ occurring in, say, 3 different documents with label assignment $l_1$=0, $l_2$=1, $l_3$=0, SPR will minimize the loss function by assigning probability $SPR(s) = 1/3$. A similar argument can be made for internal nodes in the constituency parse-trees. Thus sentences $n$ that contain actual sensitive information have MLE of $SPR(n) = 1$ because they never occur in a non-sensitive document. On the other hand, if there are sentences that never occur in a sensitive document, their nodes have MLE of 0. We find that this approach tends to provide a self-regularizing effect and we found no improvement of the accuracy of the models when adding further L1 or L2 regularizing terms to the loss function, i.e. (7.3).

## 7.4 Evaluation Methodology based on public available data

Working towards our goal of an representative empirical evaluation of sensitive information detection approaches a key requirement is the availability of a publicly available dataset. This data should contain representative examples of sensitive information from "private" document collections. Once a dataset has been modified or redacted it will in most cases no longer contain representative examples of private sensitive information. Thus detection methods trained and evaluated only on public datasets tend to target specifics and structure found in the public data rather than tuned to the sensitive information directly. These methods will often not accurately capture the specific characteristics of sensitive information and therefore the evaluation will not be representative of actual accuracy/performance. It is a further challenge that on one hand even with access to a private dataset containing sensitive information these data collections are obviously not available for publication as this would entail publicizing information which for one reason or another has been deemed sensitive. But on the other hand, public collections (such as public available corporate web pages) has, as discussed above, no inherent sensitive content to be detected and public data collections are therefore not suited for evaluation of sensitive detection approaches.

For general or complex sensitive information detection, i.e. where the definition of sensitive is complex and domain dependent, no public benchmark data exists. In this work we provide access to a public real-world dataset. The dataset contains much of the same types of structure as internal corporate communications and we use this dataset to obtain a sensitive labeling which contains different kinds of semantic sensitive relationships.

The dataset that we use is the Enron dataset [58]. The Enron data comprises of real-world complex interactions and where we can benchmark our sensitive information detection approach. The Enron dataset was not originally intended to be published, but is now available in the public domain. Therefore, we find that the dataset is uniquely suited as a benchmark of real-world actual complex communication with possible sensitive information embedded. The Enron data is a rare example of private data out in the public domain.

There have been previous attempts at utilizing the Enron dataset as a public dataset for sensitive information detection, e.g. [16]. However for real-istic evaluation we also need access to representative labeled data. I.e. relying on automatically labeled data works well for examples with low inherent complexity [16] and does not in general capture complex sensitive information well.

So, to move beyond clear cut simple sensitive information captured with automatic label generation from seed words and similar constructions, we instead rely human expert labels given to the Enron corpus as part of a competition of the TREC conference. The labels were generated by human experts. Specifically we take the annotations for documents containing information on *prepay transaction* which were developed as part of the 2010 TREC conference, legal track [22] [96].

## 7.5 Evaluation

### Evaluation Methodology and Data

Complex sensitive information detection is not only a challenging task, but also challenging to evaluate. Existing work has created evaluation ground truth for sensitivity using three main strategies. The first strategy uses word co-occurrence with seed words to semi-automatically label sensitive information as ground truth [16, 38]. This ground truth uses the same assumption as keyword-based detection that sensitive information co-occurs with other sensitive terms, but does not reveal performance on more complex sensitive information. Another strategy use sources of actual sensitive information, such as WikiLeaks, and insensitive information from other sources. The major disadvantage of this strategy is that insensitive and sensitive information exhibit major differences in structure and content [42]. Thus, evaluation may actually measure how well the differences in structure and content are learned and not necessarily how well sensitive content is detected. The final strategy uses human labeled data. However, the sensitive information in these approaches is typically simple like named entities (e.g. names of cities) [10], or in general Personal Identifiable Information detection, such as person names, sicknesses [84].

In this work we propose evaluation on the Enron dataset [58], an actual private dataset which contains both sensitive and non-sensitive data, and which is labeled by human experts. Our dataset contains examples of complex sensitive information that cannot be characterized by a few keywords. It shows the varied structure as seen in internal corporate communications and there are different sensitive issues.

While this dataset has been used before for sensitive information detection, e.g. [16], there were no complex sensitive information labels in the ground truth. We here propose to exploit the human expert labels given to the Enron corpus as part of a competition of the 2010 TREC conference, legal track [22] [96]. Specifically, we study the case of *prepay transactions*.

Table 7.1: Overview over our labeled extract of the ENRON/TREC data: number of sentences and constituency parse-trees for each split of the dataset.

| set | sensitive | non-sen | total | non-sen/total |
|---|---|---|---|---|
| train | 2985 | 6015 | 9000 | 0.6683 |
| validation | 462 | 968 | 1430 | 0.6769 |
| test | 322 | 638 | 960 | 0.6646 |
| total | 3769 | 7621 | 11390 | 0.6691 |

The Enron dataset has $1.2M+$ documents [58], and there are 2720 documents labeled by human experts.

For each sentence in any document, we obtain constituency parse-trees, 11390 in total, which we split into training, validation and test sets, see Table 7.1. All performance measures use the same test set, with no hyperparameter tuning beyond what is reported in these tables. All word vectors come from the Stanford Glove word vector set[79]. To evaluate performance, we study accuracy of detection of complex sensitive information, with a particular focus on finding sensitive data. Formally, if ground truth label $l_s = 1$ on sentence $s$ and prediction of our model $SPR(s) = 1$ then $s$ is *true positive*, if $l_s = 0$ and $SPR(s) = 1$ it is *false positive*. Similarly for negatives. $C_{tp}, C_{fp}, C_{tn}, C_{fn}$ denote the counts of true positives, false positives, true negatives and false negatives respectively. Accuracy of a data set $D$ is then $acc_D = \frac{C_{tp}+C_{tn}}{|D|}$.

## Performance evaluation for complex sensitive information

We begin by studying the impact of input dimension of the word embeddings, varying from 50 to 300. The hidden state internally is fixed at size 200. The results are shown in Table 7.2, and as we can see, there is a clear increase in accuracy as the dimension of the word embeddings is increased, but this trend diminishes. Going from 100 to 200 the increase in accuracy score is 0.028, whereas going from 200 to 300 the increase is negligible at only tiny 0.007.

Next, we investigate the impact of the size of the hidden state, here we fixed word embedding size to 300. Similarly, to input word embedding size which positively affects performance up to a certain point, we observe similar performance increase when increasing internal hidden representation size, as shown Table 7.3. There is a drop between using 200 neurons with another increase for 300 neurons, after which the performance decreases again at 500 neurons. This suggests that we reach a level which provides a good size for the internal hidden representation after which we most likely see effects of worse performance due to overfitting as the model complexity exceeds what is useful for a well generalizing model. We also evaluate the use of several layers at each node (Table 7.4). In all experiments the sum of all neurons in the models is fixed to allow for the same potential expressive power, and the number of layers varies between 1 and 3. Here we find that the best hidden representation is a single, wide layer rather than stacked thinner layers. In [46], the best performance for sentiment analysis is observed for a stacked architecture with 3 layers. While a throughout examination of these differences falls outside the scope of this work, we make the following observation; Sentiment analysis has a more *Boolean* behavior than sensitive information detection. In sentiment a single "not" in the sentence can flip the expected label as in "good" vs. "not good". This Boolean property means that the model can make localized decisions which suggest thin and high models, whereas complex sensitive information can be viewed as a measure on the complete context. I.e., sensitive information is a property of what is communicated and is not as easily flipped by a single word, which then suggests that a wide layered model will perform better.

Table 7.2: $SPR$-input; Performance for varying size of input word embeddings

| vector size | $Acc_{val}$ |
|---|---|
| 50 | 0.7531 |
| 100 | 0.7580 |
| 200 | 0.7608 |
| **300** | **0.7615** |

Table 7.3: $SPR$-hidden; Varying the size of the hidden state

| hidden state | $Acc_{val}$ |
|---|---|
| 50 | 0.7497 |
| 100 | 0.7650 |
| 200 | 0.7615 |
| **300** | **0.7678** |
| 500 | 0.7643 |

Table 7.4: Comparison of $SPR$-layer models as the number of layers is varied, keeping the total number of neurons fixed

| Hidden state | layers | $Acc_{val}$ |
|---|---|---|
| **300** | **1** | **0.7678** |
| 150 | 2 | 0.7554 |
| 100 | 3 | 0.7594 |

We compare our final model with state-of-the-art in the field of sensitive information detection both in terms of overall accuracy, and in terms of the performance when focusing mostly on the sensitive information, as opposed to correctly predicting non-sensitive documents. As discussed in more detail in Related Work in Sect. 7.6, state-of-the-art is based on word counting, i.e., word co-occurrences ($n$-gram), inference rules and mutual information models are all keyword-based. In Table 7.5, Assoc Rules denotes [16] with default parameters for "Email" corpus, C-sanitized [84] with $\alpha$ values as reported in their experiments. Keyword-Based denotes a generic keyword based approach, given optimal keyword set and is an upper bound for Assoc Rules and C-sanitized. $SPR$-input: best model of experiments in Table 7.2, $SPR$-hidden: best model of experiments in Table 7.3, $SPR$-layer: best model of experiments in Table 7.4.

Table 7.5: Overall accuracy comparison; Assoc Rules [16], C-sanitized [84], Keyword-Based generic keyword based, given optimal keyword set, $SPR$-input best of Table 7.2, $SPR$-hidden best of Table 7.3, $SPR$-layer best of Table 7.4

| Approach | parameter | $Acc$ |
|---|---|---|
| baseline | | 0.6646 |
| Assoc Rules | supp= 2, conf= 0.6 | 0.7104 |
| C-sanitized | $p_{sen} = 0.3354$, $\alpha = 2.0$ | 0.6479 |
| C-sanitized | $p_{sen} = 0.3354$, $\alpha = 1.5$ | 0.6479 |
| C-sanitized | $p_{sen} = 0.3354$, $\alpha = 1.0$ | 0.7240 |
| Keyword-Based | | 0.7476 |
| $SPR$-input | 300 | 0.7615 |
| **$SPR$-hidden** | **300** | **0.7678** |
| **$SPR$-layer** | **1** | **0.7678** |

Table 7.6: Weighing sensitive examples in loss function; accuracy for sensitive information; $F1$ measure

| Approach | $Acc_{Sen}$ | $F1$ |
|---|---|---|
| baseline | 0 | $N/A$ |
| keyword-based | 0.2795 | 0.4009 |
| $SPR_{w=1}$ | 0.3540 | 0.4720 |
| $SPR_{w=2}$ | 0.3540 | 0.4800 |
| $SPR_{w=3}$ | 0.7236 | **0.5143** |
| $SPR_{w=4}$ | **0.9224** | 0.5073 |

As we can see in Table 7.5, all approaches improve upon the base line accuracy value of 0.6646. Our models, as studied in the experiments above, outperform this baseline, and also existing keyword based approaches with respect to overall accuracy. However, as discussed before, in sensitive information detection, we are typically much more interested in successfully identifying sensitive information than we are in correctly predicting non-sensitive information. We therefore conduct an experiment that investigates this case in depth. Overall, $SPR$ successfully identifies complex sensitive information and outperforms state-of-the-art particularly when focusing on documents containing sensitive information as opposed to identifying non-sensitive information. $SPR$ shows value in capturing the phrase structure used to describe complex sensitive information that might go unnoticed when relying on keywords alone.

In Table 7.6 we weigh errors on false negatives (unidentified sensitive examples) higher in the loss function. We report class-based accuracy $Acc_{Sen}$ and also the $F1$ score which weights false negatives and false positives in a single score according to $F1 = \frac{2C_{tp}}{2C_{tp}+C_{fn}+C_{fp}}$. We obtain close to 100% accuracy on sensitive example detection by increasing this weight, but for weights greater than 3 the $F1$ score starts decreasing, reflecting that the number of false positives now is so high that overall performance across both sensitive and non-sensitive information degrades unreasonably. We observe that all of the *SPR* models in Table 7.6 have higher performance than the best result we can obtain using previous state-of-the-art algorithms. The weight parameter allows our *SPR* model to be adjusted to the domain in a natural manner; for some domains false negatives may be associated with high cost in which case a corresponding high weight can be used to ensure a relatively low number of false negatives. In other domains, on the other hand, too many false positives may be undesirable and thus here it would intuitively make sense to have a lower weight for false negatives.

Table 7.7: *SPR* semantics in related single words (top) and sentences (bottom)

| Label | Close *SPR* embeddings |
|---|---|
| Dates | May-02, Feb-02, Oct-03, Apr-01, Jun-99 |
| Names | Stacy Dickson, Martha Braddy, James Westgate, GEA Rainey, Citibank ISDA |
| Good-byes | "Yours sincerely, EnronEntityName" |
| | "Signature of Company Officer" |
| Prepay | "However, he did not attempt to calculate a VaR statistic for the daily cash requirements for the exchange traded positions" |
| | "Li identified several of these and they are given on the flowchart (gas settlements, merchant assets, etc.)" |
| Oil & Gas | "The oil flows through the orifice and into the bearings and forms a film that cools and lubricates the journal" |
| | "In accordance with NFPA, the fire and gas detection controls will be powered by a dedicated 24V DC battery system" |

### Qualitative Analysis

For our qualitative analysis (see Table 7.7) we investigate the semantics that *SPR* learns, i.e., how close selected groups of phrases are in terms of Euclidean distance in *SPR* embedding space. Table 7.7 lists single words such as "May-02" as being close to similar other dates. Please note that we did not do any preprocessing for dates and "May-02" is not in our input vocabulary. We can see that all our phrases in Fig. 7.3 seem far away from unknown input words, which suggests that our model learns semantic meaning for all phrases. The *SPR* model extracts these semantics automatically, as exemplified also for *Names* and *Goodbyes*. We observe that the space in Fig. 7.3 also contains structure where dates seem almost localized to a distinct point, where-as *Prepay* sentences form a curly line-like structure. The final listing in Table 7.7, which we termed *Oil & Gas* due to its apparent semantics, is from a region of the space completely devoid of sensitive phrases. *Prepay* on the other hand, has 99.2% sensitive sentences, and is thereby a strong indicator of how sensitive information is captured by *SPR*.

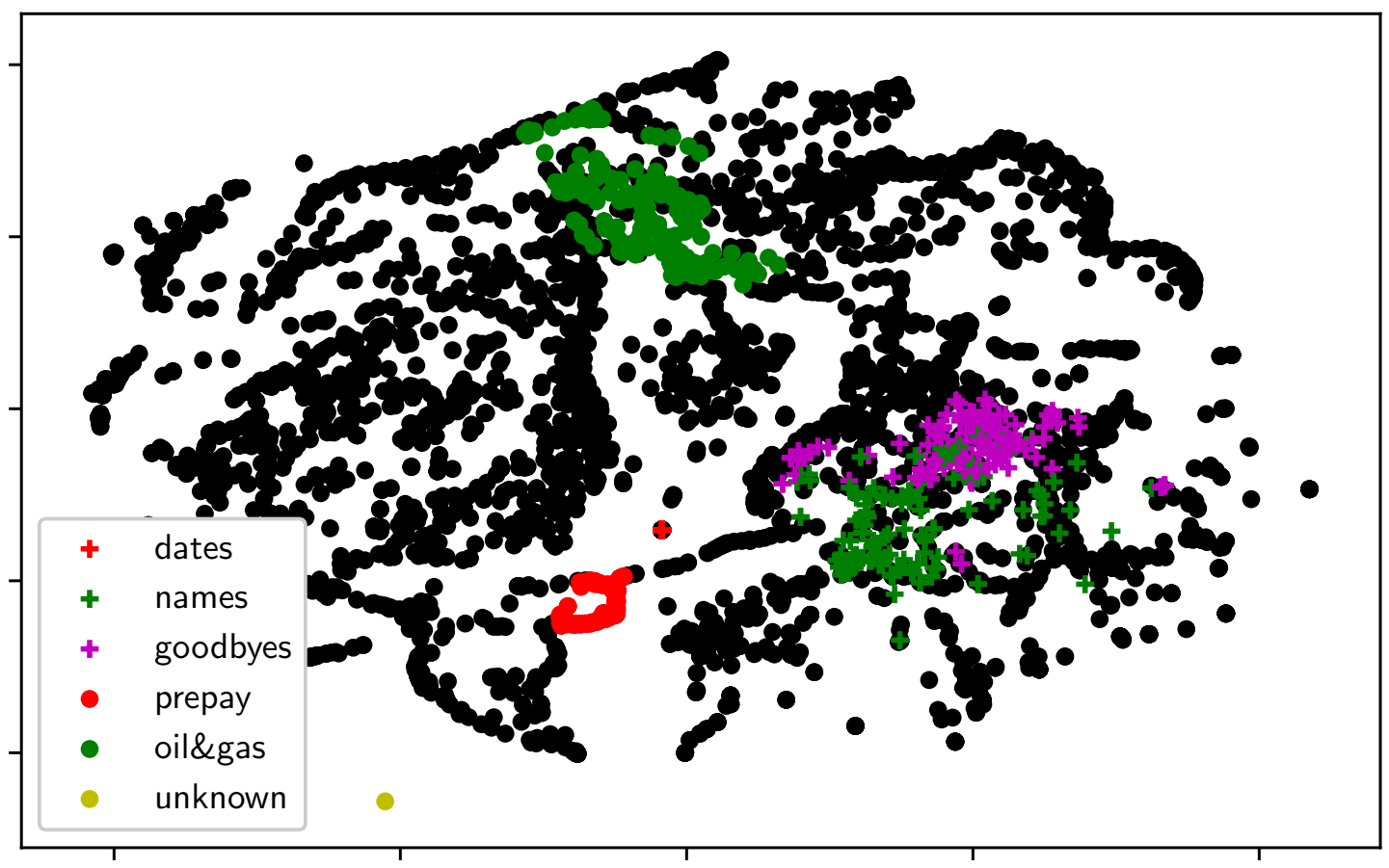

Figure 7.3: 2-dimensional view of *SPR* phrase model with colors highlighting 200 closest phrases of label groups listed in Table 7.7

Our qualitative analysis shows that the *SPR* model indeed learns semantic meaning from the documents. This suggests that the successful identification of sensitive information in *SPR* is based on its ability to identify distinct compositional semantic information.

## 7.6 Related Work

Early work in sensitive information detection is inspired by inference rule discovery [16]. Here, sensitive information is defined from an initial seed set of words and new inferred sensitive words are found using inference rules. The inference rules are created based on word co-occurrence counting such that words with high *confidence* are used as predictors. A word has high confidence if the probability of the document containing sensitive information is high given that the document contains the word. The authors argue that in contrast to other domains, such as market basket recommendations, high *support* (the word occurs often) is not required for interesting sensitive information rules as we are interested in *any* word with high confidence, regardless of support. In their qualitative evaluation, [16] use confidence for words as a sensitivity score and detect sensitive information where any word in a paragraph has high sensitivity score.

In [38], the inference rule approach is applied to detect sensitive information that is to be redacted in software code-bases. The paper presents examples where a company has released software to third parties or to the public, but where comments or function names have leaked internal sensitive information such as insights into internal business processes or names of customers. The goal of the work is to detect and redact sensitive information in a way that allows new programmers to the code-base to retain the "source code" understanding, i.e. the utility of the code-base, and at the same time minimize the inferred sensitive information. The approach of [16] is used with seed words and inferred rules to define what the sensitive information is in this particular application case. An external ontology data such as WordNet is used to count word co-occurrences for the inference rules and to obtain synonyms for actually redacting the sensitive information.

In [42], direct word-to-word (bi-grams) co-occurrence counts are used to determine which words are sensitive. Rather than considering precision or recall, the *false-discovery-rate (FDR)*, i.e., the inverse of the precision, is used to ensure low number of false positives. The authors argue that too many false positives lead to users ignoring potential leaks of sensitive information. The word sensitivity scores obtained from bi-grams are evaluated as well as a number of other black box methods. Two models are developed, one which is good at removing non-sensitive information $SA_{public}$ and another one, $SA_{secret}$, which targets detection of sensitive information. By applying the models in sequence, the number of false positives is reduced.

In [10], an assistive approach is chosen to semi-automatically rank documents in order to best detect documents containing sensitive information. The documents are ranked by maximizing *total utility*, which is a measure of the model's probability that the word is sensitive together with the *gain* associated with the word. The gain is a measure of the expected increased accuracy from learning the true label of the word, i.e., the weighted increase in $F1$ score as more false-positives or false-negatives are corrected. Experiments are performed investigating the average increase in accuracy when top 5%, 10% and 20% of the data is probabilistically given correct labels.

[84] develops a pointwise mutual information (PMI) approach. PMI is in this context the co-occurrence count of bi-grams, where one word from a seed set of sensitive words is divided by the occurrence count of the candidate word together with the count of the sensitive word. PMI is compared to a threshold of the information theoretic Information Content (IC). IC is the logarithmic value of the ratio of the occurrence count of the sensitive seed word and the size of the corpus. An additional hyper-parameter $\alpha$ can be used to control how strict the sensitive information detection should be, i.e., varying the $\alpha$ parameter allows for an increase of recall of sensitive information at the expense of higher number of false positives.

All approaches above uses single word definitions for sensitive content. While they differ in the precise sensitive information definition, they all rely on seed-word and word pair co-occurrence counting. Such co-occurrence counting of pairs ignores context as discussed above. In this work, our proposed

*SPR* model uses the structure extracted from natural language, namely the phrase trees, which feed into a recursive neural network that can handle arbitrary length structures, and thereby learn automaticlaly what characterizes the sensitive examples.

## 7.7 Conclusion

We introduce the *complex sensitive information* detection problem where context and structure in language has to be taken into account. Our *SPR* model extracts phrase structure to learn a recursive neural network model. Our experimental evaluation, which is the first to use a real document corpus containing both sensitive and non-sensitive documents with human expert labels, shows that we outperform state-of-the-art keyword based models.

**Acknowledgments**

This project has received funding from the European Union's Horizon 2020 research and innovation programme under grant agreement No. 645198 (Organicity Project) and No. 732240 (Synchronicity Project).

# Chapter 8

# TABOO: Detecting unstructured sensitive information using recursive neural networks

Jan Neerbek[1,2] Ira Assent[1] Peter Dolog[3]
[1] Department of Computer Science, Aarhus University, Aarhus, Denmark
{jan.neerbek,ira}@cs.au.dk
[2] Alexandra Institute, Aarhus, Denmark
[3] Department of Computer Science, Aalborg University, Aalborg, Denmark
dolog@cs.aau.dk

**Abstract:** Leak of sensitive information from unstructured text documents is a costly problem both for government and for industrial institutions. Traditional approaches for data leak prevention are commonly based on the hypothesis that sensitive information is reflected in the presence of distinct sensitive words. However, for complex sensitive information, this hypothesis may not hold.

Our TABOO system detects complex sensitive information in text documents by learning the semantic and syntactic structure of text documents. Our approach is based on natural language processing methods for paraphrase detection, and uses recursive neural networks to assign sensitivity scores to the semantic components of the sentence structure.

The demonstration of TABOO focuses on interactive detection of sensitive information with the TABOO system. Users may work with real documents, alter documents or prepare free text, and subject it to information detection. TABOO allows users to work with our TABOO engine or with traditional approaches, and to compare results. Users may verify that single words can change sensitivity according to context, thereby giving hands-on experience with complex cases of sensitive information.

Video: https://youtu.be/tqQ4BP0wqs8

## 8.1 Detecting Complex Sensitive Information

Detecting and redacting sensitive information in documents prior to publication, *Data Leak Prevention* (DLP), is increasingly important in industry and government, as more and more documents are made available publicly [16, 42].

Traditional approaches for DLP such as $n$-grams [42] or inference rules [16], assign sensitivity scores to words directly without considering context. These traditional approaches, including NLP inspired sentiment analysis, topic modeling (e.g. Latent Dirichlet Allocation) or Named Entity analysis [47], generally perform well for *private* information; usually entities such as location or personally identifiable information [47].

However, a core challenge in DLP is that the definition of *sensitive* information is often human specified and complex in nature. An entity, such as a company name, may be sensitive in one context and non-sensitive in another context, or sensitive information may not even be captured by a single name or term alone.

Consider the real case that we use in our demo, namely the internal and external communication of Enron, that was published when the company was prosecuted for fraud [57]. The documents contain both sensitive and non-sensitive content (manually labeled by experts with respect to complex issues such as “prepay transactions” [96]).

An example for such *complex* sensitive information are so-called “prepay transactions”, where “letters of credit” may be sensitive, but not if discussed in the context of e.g. possible loan of money. We observe that sensitive information may be embedded in the semantic meaning of the text, even when the text contains only “non-sensitive” words.

Our *TABOO* system extracts compositional sub-structures of the sentences to learn sensitivity scores. It builds on successful NLP methods including paraphrasing, sentiment analysis and image-sentence ranking [54, 88]. TABOO is motivated by the observation that complex sensitive information bears some similarity to paraphrasing. Consider the (real) examples “We need letters of credit, with approved collateral in order to approve the prepay transaction” containing the sensitive keyword “prepay transaction”. It is a paraphrase of “we have proposed letters of credit for the approved form of collateral pending further discussion" (shortened in the interest of space). Also this example is sensitive even in the absence of the keyword.

In TABOO, a Recursive Neural Network (RNN) processes the sentences in a structured way. The same neural network is applied recursively according to the structure given by the syntax tree of the text. At each node in the syntax tree the RNN generates a *representation* of the particular compositional substructure captured by that node. These representations have been shown to successfully predict whether one sentence *paraphrases* another [88].

## 8.2 The TABOO Engine

The TABOO system takes a set of documents containing sensitive information as training input. Using the trained model the system then detects sensitive content in new documents. TABOO consist of a number of steps to process the input. As a learning system TABOO has two different modes of operation; training mode and predicting mode. Training mode is used when training the RNN model for improved predictions. When TABOO is introduced to a new domain a training set must be provided and a RNN model must be trained. The user can also choose to retrain an existing model, if, say, the definition of sensitive changes over time. Once a model is obtained the system can be used in predicting mode. Here, TABOO takes a document and detects sensitive information in the document according to the model.

TABOO input documents are either loaded by scanning an input directory or through the TABOO graphical user interface where the user can load or write a document directly. Each document is then subjected to 1) Sentence splitting, 2) Syntax Parsing and 3) RNN. 1) Splits the document into sentences for further processing, using NLTK [11], the natural language toolkit for Python. 2) Extracts substructures of sentences in the form of a parse tree reflecting the structure of the sentence using the Stanford NLP parser [66]. 3) Trains a model (or detects using an existing model) on the syntax trees using Deep Recursive Neural Networks as developed by İrsoy and Cardie [46]. Training employs backpropagation-through-structure with dropout, evaluating recursively in feed-forward manner [46] and persists the model for future use. Detection outputs the highest scoring prediction per sentence under the model.

## 8.3 The TABOO System and Demo

Our interactive demo allows the user to load any document or devise any text, and subject them to a number of sensitive information detection approaches using different selectable definitions of sensitivity. The TABOO system also allows the user to load a set of documents for training a new model under a new definition of sensitivity.

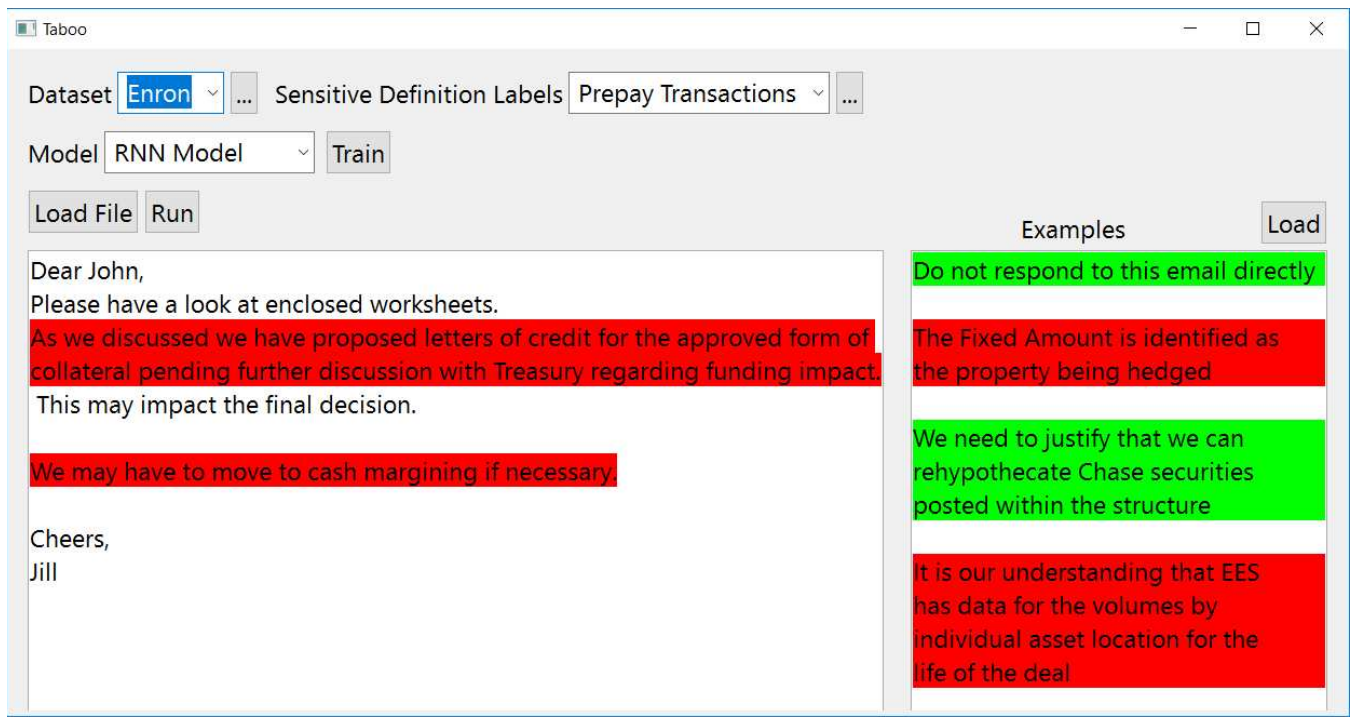

Figure 8.1: Main screen of the TABOO system. The user can load a document or edit the text directly. Clicking "Run" subjects the document to sensitive information detection under the selected model and definition of sensitive information ("Dataset"). On the right, the user has access to samples of sensitive and non-sensitive content.

For the demo, we additionally prepare real case text samples from the Enron case that users can embed in documents for testing purposes or may alter them freely to test the detection capabilities.

The TABOO demo comes with 3 different prediction engines, The RNN model (our approach), $n$-gram model [42] and inference rules model [16].

In Figure 8.1 the TABOO system interface is shown. The left frame is the main frame that holds the document for detection which is either loaded from file, created or edited directly. The right side of the screen contains samples of real sensitive and non-sensitive text for validation purposes. These samples are not used in training to avoid any bias, and are color-coded red and green to indicate sensitive and non-sensitive text snippets, respectively. Clicking allows copying them to the document in the main frame. These samples can be used as-is (in particular when working with a new domain or new definition of sensitivity) or altered to challenge the detection capabilities of the different models.

The underlying corpus (*Dataset*), as well as the sensitive definition given through labeled training documents (*Sensitive Definition Labels*), can be selected. The system comes with pre-generated models for 8 different definitions of sensitive information. The user can also choose which particular approach to use for detecting sensitive content; "RNN model", "2-gram" or "inference rules". The latter two are traditional approaches. Clicking the *Run* button executes the model on the document in the main window and highlights sentences with sensitive content (in red).

In interacting with the demo, conference attendees will gain hands-on experience with the differences in detection power of different engines, and insights into different complexities of sensitive information and the challenges associated with them. The Enron case contains accessible sensitive informa-tion that is difficult to characterize, allowing users to develop an intuition of when keyword-based approaches suffice and when the definition of sensitive information is so complex that it requires a structure-based model such as the RNN model we propose in our TABOO engine.

TABOO is designed as an analysis tool for determining the best approach for a given application and for validating documents used for training. I.e., in interacting with the system, the analyst can verify whether particular types of sensitive information are successfully captured or whether more or different training data is required. Different detection models can be persisted and used for comparison and deployment in practice, making TABOO a tool to manage different redaction requirements prior to document publication.

## Acknowledgment

This project has received funding from the European Union's Horizon 2020 research and innovation programme under grant agreement No. 645198 (Organicity Project)

# Chapter 9

# A Selective Training Approach for Very Fast Backpropagation on Sentence Embeddings

Jan Neerbek[1,2] Peter Dolog[3] Ira Assent[1]
[1] Department of Computer Science, Aarhus University, Aarhus, Denmark
{jan.neerbek,ira}@cs.au.dk
[2] Alexandra Institute, Aarhus, Denmark
[3] Department of Computer Science, Aalborg University, Aalborg, Denmark
dolog@cs.aau.dk

**Abstract:** Distributed word embeddings and related models have been shown to successfully learn natural language representations. However, for complex models on large datasets training time can be extensive, approaching weeks, which is often infeasible in practice.

In this work, we present a novel method to reduce training time substantially by selecting training instances that provide relevant information for training. Selection is based on the similarity of the learned representations over input instances, thus allowing for learning a non-trivial weighting scheme from multi-dimensional representations.

We demonstrate the efficiency and effectivity of our approach in several text classification tasks using recursive neural networks. Our empirical evaluation shows that the objective function converges up to 6 times faster without sacrificing accuracy.

## 9.1 Introduction

In computational linguistics, distributed word embeddings and related models have been shown to learn successfully natural language representations [9, 20, 50]. However, their training time can be prohibitive, in particular for large training corpora and complex models.

As a consequence, training time may be a critical factor in the deployment and advancement of more powerful, expressive machine learning models. This is certainly true for deep neural network models where training times of weeks are reported [19] and where the quest for stronger and better neural models drives doubling of models sizes (number of neurons) approximately every 2.4 years [37].

In this work, we present a novel highly effective training approach to train Artificial Neural Network models much more efficiently. The core idea is a selection strategy that focuses training to cluster-based structures in the space of representations. In a nutshell, by focusing on instances with relevant information for training, our approach requires fewer training iterations to converge to a stable and effective model.

While in this paper we focus on the deep recursive training approach called *backpropagation-through-structure (BPTS)* [35, 46] for text classification tasks, our results are applicable to most classifier models which generate distributed instance representations, e.g. all deep neural network models.

We demonstrate in our empirical study on several document corpora that the gains in training time do not come at the cost of accuracy, but may even bring a slight improved accuracy score[1].

Unlike existing work on adaptive training, such as boosting (e.g. AdaBoost [32]), where training methodology is modified, our focus is on modifying the data samples used for training in order to speed up training times. Also, as opposed to strategies that change the actual training methodology (e.g. [93]), for example by updating the model in parallel using several mini-batches, or by automatically adjusting the learning rate depending on the loss ([51]), our work selects training samples that are then subjected to the original training methodology. In this manner, our proposed method can be used in a large variety of existing training approaches, speeding up training while showing high model accuracy.

Our contributions include a novel selection strategy for training with substantial speed up while maintaining accuracy, an empirical study on four real world datasets that demonstrates the effectiveness and efficiency of our approach as well as robustness of our approach with respect to parametrization, and a detailed error analysis.

[1]Please note that the source code and the data used in our empirical study will be made available online at publication time of this article

## 9.2 Background

Distributed word embeddings [9] have been immensely successful in NLP for a wide range of task including sentiment analysis [90], POS-tagging [20] and text classification [50]. Many approaches learn unsupervised word embeddings on large corpora such as word2vec [71] and GloVe [79].

Research in generating sentence embeddings from word embeddings includes VecAvg [63] which defines them as the average of all word embeddings in the sentence. VecAvg has since been superseded by stronger models such as recurrent neural networks (RNN) using gated memory cells such as the LSTM [44].

A generalization of the recurrent model has been proposed in [87] as *recursive neural networks (TreeNN)*[2]. In TreeNN models we may incorporate semantic knowledge about the sentence in a tree or graph like structure. To evaluate a TreeNN a walk is required from node to node through the entire tree. This may negatively impact performance, and the walk may be hard to parallelize, therefore various restricted version of the TreeNN have been proposed such as Hierarchical ConvNet [21] and Graph Convolutional Networks (GCN) [52] where the number of steps in the graph is restricted to a fixed constant number. In our experimental study we consider embeddings generated by the full TreeNN model over the constituency parse trees as proposed in [88], but our approach is more general and only assumes embeddings and thus can be applied to any text/sentence embedding generating approach as well.

## 9.3 Problem Definition

Given a dataset $D$ of $n$ texts $D = \{d_1, d_2, \ldots, d_n\}$, a ground-truth labeling $L : D \to \{1, 2, \ldots C\}$ with $C$ classes, and label $L(x)$ for $x \in D$, train a given model $m$ using as few training cycles as possible. Model $m$ is parametrized by a set of parameters $\theta \in \Theta$, where $\Theta$ denotes all possible model parametrizations. The approach used to find a good model is referred to as the learning approach, which we denote $T_m$, e.g. $T_m$ could be the application of backpropagation on model $m$.

Given (mini-)batches of $b$ texts per cycle the learning approach updates the set of parameters $T_m : (\Theta, D^b) \to \Theta$, i.e., given a parametrized model $m_\theta$ and a subset $D' \subseteq D$ of $b$ texts the training function returns a new set of parameters $\theta'$: $\theta' = T_m(\theta, D')$.

Efficiency of the training approach is then the expected number of random batches of training texts that needs to be processed by $T_m$ before the performance of the model $m_\theta$ converges, i.e., further batches do not bring

[2]In this work we refer to recursive neural networks as *TreeNN* to avoid name clash with RNNs.

$m_\theta$ closer to $L$, as indicated by an error measure, such as the squared error $(L(x) - m_\theta(x))^2$. We wish to minimize the objective function $obj(m_\theta) = \frac{1}{|D|}\sum_{x\in D}(L(x) - m_\theta(x))^2$, . Given a threshold $\epsilon$, model $m_{\theta'}$ as the model after training $m_\theta$ on additional $\delta$ batches, then we say that the model $m$ has *converged* iff $obj(m_\theta) - obj(m_{\theta'}) < \epsilon$ , i.e., further batches do not bring $m_\theta$ closer to $L$.

In this paper, the expected number of batches to produce a good model is used as a measure of the training time needed, denoted $t(T_m)$ or just $t_m$ for short. We define the training time optimization problem for a dataset $D$ with labeling function $L$, a model architecture $m$ and training method $T_m$ as the minimization of $t(T_m)$ under the constraint that model accuracy be preserved.

We consider the hard constraint that given a model trained on the full training data $m_\theta^f$ and the model trained using our selection strategy $m_\theta^s$, then selection must perform at least as well as the full model, i.e. $obj(m_\theta^s) < obj(m_\theta^f)$. It would also be possible to consider soft, approximate versions, sacrificing some accuracy in the interest of training time, by demanding $obj(m_\theta^s) - obj(m_\theta^f) < \Delta$ for some small $\Delta$.

For embedding based approaches we use a distributed representation of the input to predict the correct label. Thus, our model $m_\theta$ can be split into a predictive part $m_\theta^p$ and an embedding generating part $m_\theta^e$ as follows: $m_\theta = m_\theta^p m_\theta^e$. In complex models the embedding layer may in fact consist of several layers $m_\theta^e = m_\theta^L \ldots m_\theta^2 m_\theta^1$. For our method, we only use the most informative embedding which is the final embedding produced just before a prediction is made.

Evaluating model $m_\theta$ on training instance $x$ yields $m_\theta(x) = m_\theta^p m_\theta^e(x)$. We refer to the output of the embedding layer as a *representation* of $x$: $repr(x) = m_\theta^e(x)$.

The key idea in this work is to exploit this representation as a means to identify a subset of the training instances that does not contain information for training. Excluding it from further training reduces training time $t(T_m)$ while not hurting accuracy.

## 9.4 Our Approach

To minimize the number of training cycles $t(T_m)$ our goal is to select the most informative training examples to be provided in training batches. Intuitively, we want to choose the batches which increase performance of our model the most.

Given a model $m_\theta$ and input $x$ we consider the generated representations $repr(x)$. If our model is strong then two inputs, $x$ and $x'$, with different labels $L(x) \neq L(x')$ should have different representations - because if the representations were very similar then it would be challenging for the model prediction layer to distinguish between the two representations. Thus, we expect the two representations to be far apart. Furthermore, for a successful (strong) model this behaviour can be expected across all instances. We informally refer to this as the *structure of the representation space.*

We observe that the structure of the representation space allows us to select training instances of interest, namely groups of instances with different labels in close vicinity in representation space. Clearly, training on instances with similar representations and identical label provides less information for distinguishing between classes.

We therefore pose the problem of finding subsets of interesting training instances as a clustering problem. While in general any hard clustering method can be used, we use the well established K-means [17, 30, 64] as a simple and efficient choice. In brief, K-means assigns representations into $k$ clusters such that the total sum (TS) of $L2$ distances between cluster centroid and cluster members is minimized: $TS = \sum_{x_i \in D} \|repr(x_i) - c_j(x_i)\|^2$, where $c_j(x_i)$ is the centroid of the cluster that input $x_i$ is assigned to.

Let $C = \{C_1, C_2, \ldots, C_k\}$ be the resulting clusters, grouped exclusively based on the similarity in representation space. We now analyze them with respect to label purity to identify areas in the representation embedding that have already been learned and can be omitted from further training. We formalize this analysis as the ratio of the *most frequently occurring (MFO)* class label ratio in a cluster, $MFO_i$, as $MFO_i = \max_{\ell \in L} \frac{\sum_{x \in C_i : L(x) = \ell} 1}{|C_i|}$ i.e., the ratio of the most frequently occurring class label in a cluster $C_i$ is the maximum (over possible labels) of the ratio between the number of instances with that label and the cardinality of the cluster. Clusters with low $MFO_i$ are valuable for training, whereas those where $MFO_i$ is close to 1 are uninteresting as little more can be learned. A strong model has $MFO_i$ for all clusters close to 1 (otherwise accuracy is low, see above). If there are only two classes $\{0, 1\}$, we simplify using the ratio for only one class $f_i = \frac{\sum_{x \in C_i : L(x) = 1} 1}{|C_i|}$ and obtain $MFO_i = \max(f_i, 1 - f_i)$.

Our goal is to separate interesting from uninteresting clusters by finding a suitable threshold for the $MFO$ ratio to filter uninteresting training instances away and focus training on interesting ones only. More formally we wish to choose the lowest possible $MFO$ threshold such that models trained with our approach $m_\theta^s$ satisfies the hard optimization constraint, i.e. that the objective function value over the data on our obtained model is at least as low as the objective value obtained with standard training approach.

This can be seen as a balance between two forces: First, we want as high a $MFO$ threshold as possible. Higher $MFO$ means we are filtering only data where we are very sure of the label, and do not mistakenly dismiss information from training. Setting the $MFO$ threshold too low might lead to loss of information that decreases the overall model accuracy. Second, we want to remove as much training data as possible. Removing more data, i.e., setting the $MFO$ threshold lower means we have reduced training to fewer instances, i.e., we expect to use fewer minibatches for learning.

Obviously, the maximum $MFO$ threshold is 1, but removes nothing and obtains no training time improvement. If, on the other hand, we set the threshold very low to remove as much training data as possible, clusters with low $MFO$ are dismissed from which we could potentially learn more. Clearly, the best filtering cutoff is a trade-off. We propose to study the decrease in $MFO$ in the log-scale, and define $\Delta MFO = -\log_{10}(1 - MFO)$. Intuitively, we want remove as much training data as possible while keeping $\Delta MFO$ as high as possible. Our empirical study suggests values in the range $2 \geq \Delta MFO \geq 1.5$.

**Algorithm 1** Proposed fast training approach

**procedure**<br>
    $D \leftarrow$ corpus of labeled documents<br>
    $k \leftarrow$ Number of clusters to generate<br>
    $MFO_{cut} \leftarrow$ cutoff for filtering<br>
    **while** *pretraining* **do**<br>
        $\Delta Acc \leftarrow$ rate of acc. improv.<br>
        **if** $\Delta Acc$ starts dropping **then**<br>
            break pretraining<br>
    Cluster using K-means<br>
    $MFO_i \leftarrow MFO$ for cluster $i$<br>
    $D' \leftarrow$ clusters with $MFO_i \leq MFO_{cut}$<br>
    **while** *main-training* **do**<br>
        train on reduced dataset $D'$<br>
        **if** convergence is achieved **then**<br>
            break main-training

Algorithm 1 outlines our training approach. At prediction time we match new data to clusters. For data in removed clusters we use its dominating class label. For other clusters use the model trained on the reduced set. In the experiments, we demonstrate that our approach indeed converges faster, i.e., uses fewer training epochs and converges to an accuracy that is as high as the one for full time training on all data.

## 9.5 Evaluation

### Data and experimental setup

We use the raw text document corpus Enron [58], comprised of actual email and company documents that vary greatly in style, language and length and thus provide excellent insights into performance on different types of text.

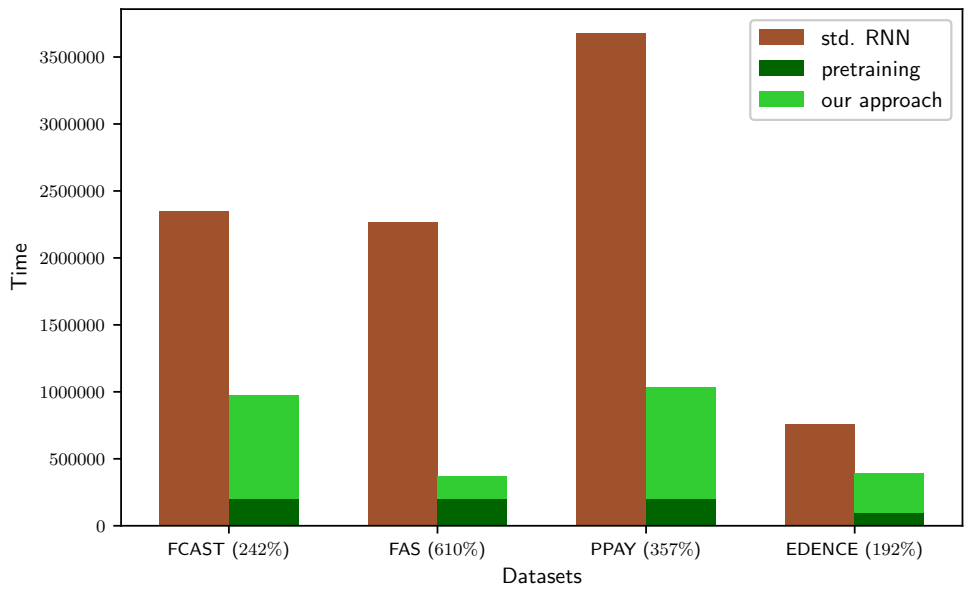

Figure 9.1: Training time (in mini-batches) on 4 datasets

All documents are split into sentences and then constituency parse trees (i.e., splitting sentences into phrases). Each phrase is to be placed in the embedding space.

We use labels from ongoing efforts of the TREC conference [22, 96], where different topics were labeled by at least 3 human annotators: $FCAST$**:** 267366 sentences regarding Enron's financial state. We use 40000 sentences for validation, 40000 for testing, and the rest for training. The ratio of sentences in class 1 is approximately 31%.

$FAS$**:** 178266 sentences where Enron claims compliance with Financial Accounting Standards[3]. We use 27000 sentences for validation, 27000 for training. The ratio of sentences in class 1 is approximately 59%.

$PPAY$**:** 134256 sentences about financial *prepay transactions*. We use 15000 sentences for validation, 15000 for testing. Approximately 13% labeled class 1.

$EDENCE$**:** 167913 sentences discussing tampering with evidence. We use 25000 sentences for validation, 25000 for testing. Approximately 23% in class 1. For further details see [22, 96]. Our processed datasets will be made available online for the research community.

## Empirical study and discussion

Fig. 9.1 on 4 different datasets of varying complexities shows that standard backpropagation converges on the $EDENCE$ dataset in $758,000$ minibatches, and on the $PPAY$ dataset, $3,674,000$ minibatches are needed for convergence even though the datasets are of comparable sizes. Our approach reduces the training time by a factor of approximately 2 to 6. The actual reduction factor seems to depend on intrinsic properties of the dataset. Consider $FCAST$ and $FAS$ which have approximately the same runtime on the full dataset. Our approach reduces this for $FCAST$ by factor 2.42, while for $FAS$ we obtain an impressive factor of 6.1. It seems that $FAS$ can be learned from fewer training instances, which our method picks up on. Our approach can also be used to improve model accuracy by training using more minibatches than the minimal required (not shown due to space constraints; small accuracy improvements of $0.1\% - 0.4\%$).

---

[3] `http://www.fasb.org/jsp/FASB/Document_C/DocumentPage?cid=1218220124871`

| Pretraining | Training | Total |
|---|---|---|
| 0 | $3,674,000$ | $3,674,000$ |
| $50,000$ | $2,128,000$ | $2,178,000$ |
| $100,000$ | $2,365,000$ | $2,465,000$ |
| $150,000$ | $913,000$ | $1,063,000$ |
| **$200,000$** | **$830,000$** | **$1,030,000$** |
| $250,000$ | $1,036,000$ | $1,286,000$ |
| $500,000$ | $1,818,000$ | $2,318,000$ |

Table 9.1: Pretraining cutoff on $PPAY$

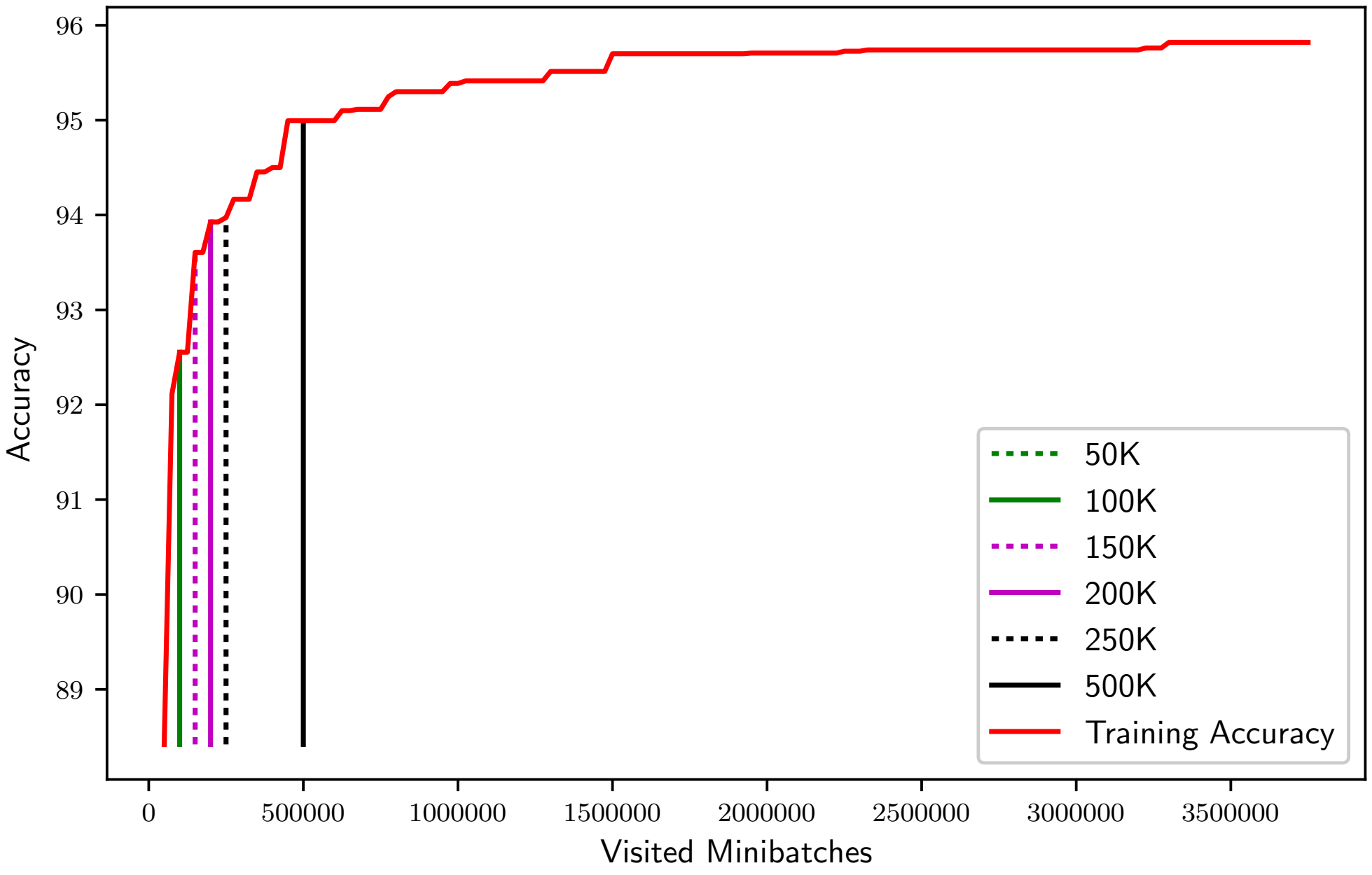

Figure 9.2: Pre-train cutoffs, $PPAY$ training curve

Stopping pre-training, i.e., when to cluster and remove instances, can be determined from the graph over accuracy as a function of training time (see Fig. 9.2 for accuracy on $PPAY$ using backpropagation; other datasets show similar behaviour; omitted here).

Vertical cuts on the training graph show pretraining stopping points and the number of additional mini-batch visits required after filtering.The minimum (Total column) is at $200,000$ mini-batches. Comparing with cut-lines in Figure 9.2 we note that the best stopping point for pretraining is where the curve “bends”, i.e., where the rate of improvement starts to plateau. At this

| $MFO$ | $\Delta MFO$ | Cluster count | Percentage | Total Training |
|---|---|---|---|---|
| - | (0) | - | - | $3,674,000$ |
| 0.9970 | 2.52 | 1 | 3.6% | $3,362,000$ |
| 0.9945 | 2.26 | 3 | 11.9% | $2,574,000$ |
| 0.9940 | 2.22 | 4 | 19.3% | $1,194,000$ |
| 0.9867 | 1.88 | 5 | 23.0% | $1,030,000$ |
| 0.9863 | 1.85 | 6 | 26.9% | $891,000$ |
| 0.9858 | 1.85 | 8 | 34.3% | $1,480,000$ |
| 0.9800 | 1.70 | 12 | 50.6% | $1,739,000$ |
| 0.9720 | 1.55 | 15 | 59.9% | $> 5,000,000$ |

Table 9.2: Filter percentage, $PPAY$, $200K$ minibatches

.

point we have gained most (further gains are more expensive) and thus have the best potential for out-performing full training. Further, our approach relies on meaningful sentence embeddings and the "bend" occurs when large-scale statistical properties of the data are encoded.

To determine how many clusters should be filtered out, we take the model after $200,000$ minibatches and generate sentence representations. These representations are clustered using K-means and we filter out all clusters with a higher $MFO$ ratio than a filter cutoff. Figure 9.3 shows clusters sorted by the ratio of sentences with class 1 (using the simplified approach for 2-class problems as described above). Clusters to be filtered (i.e., high $MFO$ ratios) are at the far left and right and clusters to keep are at the center. Note that the $MFO$ ratio for the far left clusters is much higher than the $MFO$ ratio for the far right clusters. For dataset $PPAY$ it seems easier finding pure clusters of label 0, whereas clusters with a high ratio of label 1 tend to be mixed with many examples of label 0 occurrences. Thus, here we only filter clusters on the far left.

We test different filtering cutoffs from 1 cluster (3.6% of data filtered) to 15 clusters (59.9%) and show converge times in Table 9.2. We observe marked improvements for filter cutoffs in the range 19.3% to 26.9%. Cutoffs are shown as vertical lines against class 1 ratio in Fig. 9.3.

Table 9.2 shows a large drop in $\Delta MFO$ from 2.22 to 1.88, where we thus should place our cut to obtain a runtime of $1,194,000$, which is significantly less than the standard runtime of $3,674,000$.

Table 9.3 shows the effect of different values of $k$ is small on runtime. We test different $k$ values between 15 and 70 and compare runtime of 3 different filtering cuts for each. Each cut is a row in Table 9.3 ("Small (S)", "Medium (M)", "Large (L)"). "Medium (M)" corresponds to the optimal size cut and the others are smaller and larger, respectively. We set the $MFO$ cutoff such that we filter approximately the same amount of examples across $k$ values (ratio of filtered examples given in column "Size"). Ratio of examples filtered

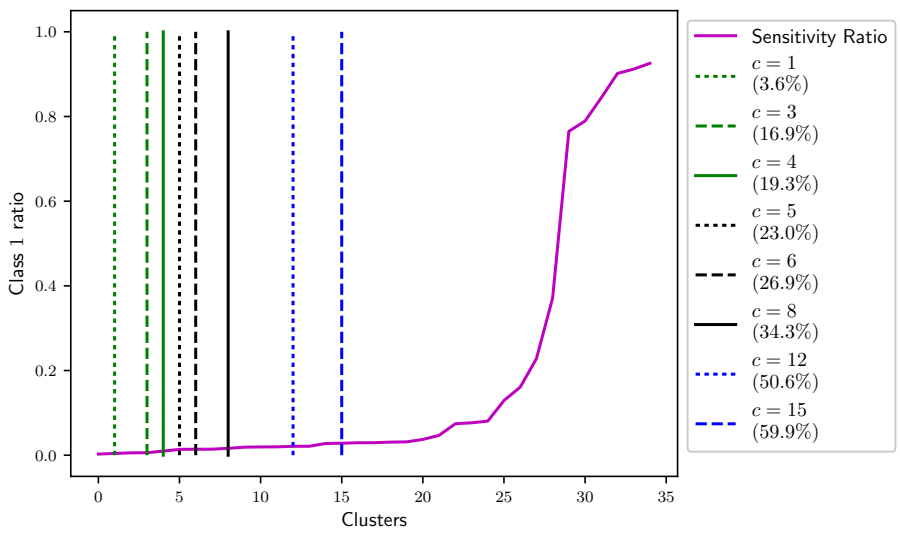

Figure 9.3: Filter cutoffs (Table 9.2) over class 1 ratio.

| | $k = 15$ | | $k = 35$ | | $k = 70$ | | |
|---|---|---|---|---|---|---|---|
| | Size | Time | Size | Time | Size | Time | Avg. Diff |
| S | 11.92% | $1,486K$ | 11.91% | $1,683K$ | 11.87% | $1,756K$ | 6.32% |
| M | 27.84% | $924K$ | 26.92% | $891K$ | 26.92% | $1,040K$ | 6.19% |
| L | 33.66% | $1,551K$ | 34.32% | $1,480K$ | 33.07% | $1,933K$ | 11.21% |

Table 9.3: Runtimes over $k$ and filtering cutoffs (Small (S), Medium (M) and Large (L)). Size is filtered data ratio. Time is measured as runtime of 1000s of minibatches, i.e., $1,486K$ means that the training took $1,486,000$ minibatches

varies only slightly due to filtering an integral number of clusters, but still allowing a comparison of runtimes across different values of $k$.

Across $k$ values we experiment with filtering a "Small (S)" amount ($\approx$ 12%) of data, a "Medium (M)" amount ($\approx$ 27%) and a "Large (L)" amount ($\approx$ 33%). Table 9.3 shows that runtimes only vary slightly, i.e., an average difference from mean of 6.32%, 6.19%, 11.21% for Small, Medium and Large, respectively. Thus, $k$ does not impact reduction in training time significantly.

## 9.6 Detailed Error Analysis

We study whether removing training samples leads to models that generalize as well as full training. We start by comparing prediction accuracy for data used by our approach and full training in $PPAY$ (76067 sentences with embeddings from both).

Table 9.4 shows the standard approach misclassifies 1926 instances, whereas our approach only misclassifies 1565. A more detailed inspection reveals that out of these 1565, 869 are made by the full training approach as well, i.e., our approach makes 18.74% fewer errors and out of the remaining errors 55% are identical to those of full training. We conclude that our speed up does not jeopardize accuracy, but even leads to slightly improved results.

We analyze errors using K-means clustering on the embeddings (Table 9.5) and observe that except for group 7 with 310 elements, all other clusters have size around $100 - 200$. This uniform size distribution (especially compared to Fig. 9.3) suggests that the errors are not concentrated in any particular part of the embedding space. Our approach thus is expected to generalize well, at least there is no particular part of the embedding space where our approach is weak.

| Total number of shared embeddings | 76067 |
|---|---|
| Errors of the standard training | 1926 |
| Accuracy of the standard training | 97.47% |
| Errors of our training method | 1565 |
| Accuracy of our training method | 97.94% |

Table 9.4: Shared embeddings in $PPAY$

| Group id | Cardinality | Perc. of class label 1 |
|---|---|---|
| 1 | 97 | 100.0% |
| 2 | 113 | 100.0% |
| 3 | 112 | 99.1071% |
| 4 | 188 | 98.9362% |
| 5 | 134 | 89.5522% |
| 6 | 202 | 78.7229% |
| 7 | 310 | 74.8387% |
| 8 | 78 | 71.0526% |
| 9 | 185 | 48.6486% |
| 10 | 148 | 41.8919% |

Table 9.5: Error group statistics

Before studying actual examples, we note that the first four groups in Table 9.5 contain 510 out of 1565 of our errors (32.59%) with high MFO, all for label 1, which makes these groups interesting for future work on improving our approach. Groups $6, 7, 8$ have a medium $MFO$ score, with group 5 showing a slightly higher score. Finally, groups $\{9, 10\}$ show lower MFO. From Table 9.6 with error examples from each group, we observe the following types of errors:

**Soft errors.** Clusters $2, 3, 4$ contain examples of prepay transactions (label 1) where wording and structure seem to be good indicators of the class. These clusters are candidates for future improvements. Cluster 7 seems similar, but shows greater diversity in label distribution, which is more challenging to resolve.

**Poor filtering.** Cluster 8 contains examples of sentences which should have been filtered out in pre-processing, and can therefore be used to inform the pre-processing step.

**Short emails and headers.** Clusters $6, 9$ contain examples of short

| Group Id | |
|---|---|
| 1 | Type of error: Document headlines<br>Example: Section 12.1 Duration |
| 2 | Type of error: Generic sentence<br>Example: I spoke to Tim today regarding the prepayment transactions with terminated counterparties |
| 3 | Type of error: Stating facts<br>Example: Let me know, but as I understand it, the Chase agreement should be a good format for this master agreement |
| 4 | Type of error: Prepay technicalities<br>Example: A similar swap was entered into between the American Public Energy Agency and Chase on the same date |
| 5 | Type of error: Specific location<br>Example: New York, New York 10043 |
| 6 | Type of error: Short standard email<br>Example: If there is anything else you need please feel free to call me on 0207 783 5404 |
| 7 | Type of error: General question wrt. entity<br>Example: Please let me know the nature of the transaction with National Steel |
| 8 | Type of error: Bad text/filtering<br>Example: [...] – – – – – - Enron-6.11.00.ppt |
| 9 | Type of error: Preambles, email header<br>Example: RE: Swap Transaction; Deal No M180816 |
| 10 | Type of error: Abbreviation and numbers<br>Example: The PSCO Project will be sold into TurboPark on 1/19/01 |

Table 9.6: Samples from each of the 10 error groups with description of type of error. The Id column is over error Group Id and the second column contains a description/headline of the type of error and an example from the dataset of each type of error.

email sentences. Again, this can be used to inform pre-processing e.g. combining them with email subject, to/from or the like.

**Hard errors.** Clusters $1, 5, 10$ contains headlines, locations and sentences with little information, which we consider unlikely targets for improvement based on sentences alone.

In summary, some groups can be used directly to improve performance further in pre-processing, others offer potential for future work, and few contain too little information to be correctly predicted. Models trained using our faster selection strategy seem to generalize just as well as full training, providing even slightly better accuracy.

## 9.7 Related Work

There is a large body of research on different training strategies, often with the aim of improving the accuracy or stability of a classifier. The classical approach is AdaBoost [32] which adapts training depending on the error observed by weighing "difficult" examples higher. At a very abstract level, boosting also refocuses training.

However, there are two core differences between our approach and boosting: first, we aim to reduce training time (i.e., speeding up convergence of the model parameters) by removing samples that are not expected to benefit. And secondly, we base the selection of samples on their multi-dimensional representation rather than on the one-dimensional difference in ground truth label and predicted label.

For an extensive survey of bagging and boosting in classification we refer the interested reader to [59].

Adaptive learning methods, such as the one presented in [51] automatically adjust the learning rate based on adaptive estimates of lower-order moments in the loss function. [51] observe faster convergence of the model during training to a particular accuracy (cost) value compared to other popular learning optimizers (AdaGrad, RMSProp, SGDNesterov, AdaDelta). Adaptive learning methods thus essentially adjust the learning rate, whereas our approach takes a fundamentally different approach that adapts the training data being used to learn accurate models much more efficiently.

A slightly different angle to the problem is taken in [93], which presents a demonstration from COLING 2016 along with an analytical discussion that model parameters can be updated in parallel during training of several minibatches. [93] observe parallel updates lead to read-write conflicts and following potential errors in model parameters, but demonstrate one can still obtain consistent results. Please note that our work again does not change the training methodology as such, but instead the data to improve runtimes.

[99] combine local word context (neighboring words) with global context (which texts or clusters does the word appear in) for learning text embeddings.

They use K-means to identify different cluster centroids as global contexts. Please note that while we also use K-means clusters in our approach, our goals is not to distinguish between local and global context nor to improve the embeddings, but instead to identify those clusters that should be removed from training. The clusters that we are interested in actually contain little (relevant) information such that dismissing them from further training speeds up learning without hurting accuracy.

## 9.8 Conclusion

We present a novel approach for reducing training time on large text corpora. We propose studying groups in representation space to identify where learning from training data seems to be complete, and where more training is expected to improve model accuracy. Discarding clusters with pure label distribution, we refocus training to those samples that lead to high accuracy models with less training time.

We show how to easily infer the parameters for selecting clusters using rate of improvement on training graphs and our proposed $\Delta MFO$ measures.

In thorough experiments, we demonstrate up to 6 times faster training without loss of accuracy on a number of datasets based on human labels. Overall, our method allows training complex models on large data volumes by identifying those samples that are most important to learning, while achieving even slight improvements in training accuracy.

# Chapter 10

# A real-world data resource of complex sensitive sentences based on documents from the Monsanto trial

Jan Neerbek[1,2] Morten Eskildsen[1] Peter Dolog[3] Ira Assent[1]
[1] Department of Computer Science, Aarhus University, Aarhus, Denmark
jan.neerbek@cs.au.dk,morten@moddi.dk,ira@cs.au.dk
[2] Alexandra Institute, Aarhus, Denmark
[3] Department of Computer Science, Aalborg University, Aalborg, Denmark
dolog@cs.aau.dk

**Abstract:** In this work we present a corpus for the evaluation of sensitive information detection approaches that addresses the need for real world sensitive information for empirical studies. Our sentence corpus contains different notions of complex sensitive information that correspond to different aspects of concern in a current trial of the Monsanto company.

This paper describes the annotations process, where we both employ human annotators and furthermore create automatically inferred labels regarding technical, legal and informal communication within and with employees of Monsanto, drawing on a classification of documents by lawyers involved in the Monsanto court case. We release corpus of high quality sentences and parse trees with these two types of labels on sentence level.

We characterize the sensitive information via several representative sensitive information detection models, in particular both keyword-based (n-gram) approaches and recent deep learning models, namely, recurrent neural networks (LSTM) and recursive neural networks (RecNN).

Data and code are made publicly available.

## 10.1 Introduction

*Sensitive information detection* addresses the problem of identifying (parts of) text documents that are considered sensitive in a particular application context. Sensitive information detection is of great importance in a number of applications, where unintended leak of sensitive information may incur severe negative consequences for individuals, businesses or authorities. In a study from 2017 Poneman Institute and IBM find that the average cost of a breach of sensitive information is $3.6 million in total, for detection, escalation, notification, and after-the-fact response [81, 82].

Sensitive information detection has been studied in the Natural Language Processing and Machine Learning research communities [10, 16, 36, 38, 42, 74, 84, 85]. In this work we focus on sensitive information detection considered as a form of text classification, where the goal is to predict whether a sentence contains sensitive information. Here we consider sensitive information to be domain specific and the definition in 4 different datasets in this work is given by labels from domain experts.

A distinguishing feature of sensitive information detection with respect to traditional text classification is that we are interested not only in the content (topics and entities) but also in context [36]. In this work we follow [36] and focus on context within a sentence.

A limiting factor in research and evaluation of sensitive information detection methods is the lack of high quality corpora, which at least in part can be attributed to the very nature of sensitive data. Given the lack of publicly available real-world data, existing work has resorted to the creation of ad hoc evaluation data by defining particular seed keywords as sensitive [16, 38, 84], or by using two distinct data sources for sensitive and non-sensitive data [42]. Evaluations using these data sources are thus limited to comparatively simple sensitive information that is captured by keyword co-occurrence alone, or may account for structural differences in data sources rather than for actual accuracy of sensitive information detection.

As an example of the types of sentences we encounter in the Monsanto corpus we provide here an example of a sentence from the *GHOST* sensitive information type:

```
But I suspect that is wishful thinking Are you interested
in writing a column on this topic?
```

This sentence has been labeled sensitive by the annotators, and indeed the sentence discusses writing for Monsanto. Note that the sentence does not explicitly talk about ghost writing or even authorship of the written material. This is an example of where context influences sensitivity.

So far, a single corpus provides a language resource with real sensitive information, namely, the Enron corpus [22, 58]. It contains corporate documents with a variety of information content and structure, and has been used extensively to evaluate sensitive information detection [16, 42, 84]. However, the corpus is unlabeled and more than 15 years old. Thus, the need for an up-to-date labeled corpus for state-of-the-art empirical evaluation.

We here present a new real-world sentence resource with complex sensitive information. We process, label, analyze, and characterize recently released documents that are part of the Monsanto trial [7] as a source of great value to the research community. We provide two sets; the first contains inferred sentence level datasets based on expert labels at document level by lawyers involved in the trial ("silver" labels); the second contains labels directly annotated manually at the sentence level ("golden" labels). Following the 4 different sensitive information definitions extracted from the trial documents, we in total provide 8 classification datasets and $15,000$ labeled sentences. We release this language resource into the ELRA Catalogue of Language Resources[1] together with source code for loading of corpus and building of new models[2].

Furthermore, we characterize the complexity of the datasets in terms of the sensitive information content in the sentences. In particular, we study traditional sensitive detection methods such as n-gram or inference rule based approaches and more recent deep LSTM models and recursive neural networks. We find that all datasets present *complex* sensitive information which is not fully captured by traditional models. Complex models that consider phrase-like context capture more of the complexity of the sensitive information. Still, some sensitive information is not detected using existing methods, which provides interesting open problems for research and evaluation of future methods.

## 10.2 Related Sensitive Information Corpora and Related Work

Sensitive information corpora are scarce due to the inherently private nature of the data. This poses a challenge to research in sensitive information detection. We here review document collections used for evaluation purposes.

Open datasets such as Wikipedia has been used for detecting well-defined types of sensitive information, e.g. Personal Identifiable Information (PII); *HIV* (Health) [85] or *Catholicism* (Religion) [84]. As discussed e.g. in [74], such forms of PII are often defined as a seed set of named entities which are comparatively easy to detect, and such resources are thus not sufficiently challenging for realistic sensitive information detection benchmarking.

[1] http://catalogue.elra.info

[2] https://github.com/neerbek/taboo-mon

WikiLeaks is used as a sensitive information source in [42], where other webpages are considered non-sensitive. That is two very different sources for content (internal secret documents versus public webpages) and successful distinction may be due to differences in data source rather than sensitivity of content. See also discussion in [74].

[10] uses a corpus of 1111 historical records from UK government on *Personal Information* and *International Relations*. The corpus is not publicly available due to its sensitive nature. [67] shows that both types of information can be modeled using features such as entity (person or country) and sentiment towards this entity, and thus does not capture aspects of sensitivity beyond such entity (sentiments).

The Enron corpus [16, 42] has been partially labeled by law students as part of the TREC legal track NLP tasks [22].

The corpus we present in this work contains real documents with complex sensitive and recent content. It complements the Enron corpus which concerns mostly finances with further complex sensitivity notions as discussed below.

## 10.3 Curation of the Monsanto Datasets

The Monsanto papers and the series of trials from which they originate are still ongoing. The trial(s) was begun in 2017 where a group sued Monsanto for claiming Roundup[3] to be safe, while Monsanto allegedly knew that Roundup could cause cancer. The Monsanto papers are internal papers from Monsanto, relevant to the trials and released due to effort by Baum, Hedlund, Aristei & Goldman law firm during the trials[4] [7, 69].

As part of this trial [7], law firm Baum, Hedlund, Aristei & Goldman categorizes Monsanto corporate documents into four categories (see below). No formal definition is provided, but a headline and description of the (sensitive) content in each document. Below we list the headlines and our informally derived descriptions of sensitivity notions (manually created based on content of sampled documents):

- *GHOST*, Ghostwriting, Peer-Review & Retraction. Concerns article writing and peer-reviewing by Monsanto salaried people as well as efforts in pressuring journals to retract damning studies without revealing Monsanto connection

- *TOXIC*, Surfactants, Carcinogenicity & Testing. Concerns chemical glyphosate (part of Monsanto product Roundup), in particular toxic-

[3] Roundup is a herbicide which is developed by Monsanto

[4] `https://www.baumhedlundlaw.com/toxic-tort-law/monsanto-roundup-lawsuit/`

| | |
|---|---|
| Number | 21 |
| Id | $MONGLY03934897$ |
| Link | http://baumhedlundlaw.com/pdf/monsanto-documents/25-Invoice-Showing-Monsanto-Paid-$20000-to-Expert-Panel-Member-Dr-John-Acquavella.pdf |
| Link text | Invoice Showing Monsanto Paid $20,000 to Expert Panel Member Dr. John Acquavella |
| Description | This document is an invoice dated August 31, 2015 from Monsanto to Dr. John Acquavella in the sum of $20,700 for "consulting hours in August 2015 related to the glyphosate expert epidemiology panel." at *1. |
| Number | 27 |
| Id | $MONGLY02085862$ |
| Link | http://baumhedlundlaw.com/pdf/monsanto-documents/4-Internal-Email-Further-Demonstrating-Heydens-Involvement-Drafting-Expert-Panel-Manuscript.pdf |
| Link text | Internal Email Further Demonstrating Heydens' Involvement in Drafting Expert Panel Manuscript |
| Description | This document contains an email from Dr. Heydens to Ashley Roberts regarding the introduction to the Expert Panel Manuscript. Among other features, Dr. Heydens' draft attempts to convey "that glyphosate is really expansively used." at *1. |

Table 10.1: Example metadata harvested in human readable form: for each document number and id, a link to the source, a link text and a brief description are provided.

ity; declining funding for further studies; declining requested studies or declining data to regulators

- *CHEMI*, Absorption, Distribution, Metabolism & Excretion. Discussion on studies and results with regards to how animals and human react/absorb ingredients found when using Monsanto products. Discussions on starting studies deemed "risky". (Note, while *TOXIC* is concerned with when and if Monsanto's products might cause cancer, *CHEMI* is more concerned with the actual chemical reactions with ingredients found in Monsanto products.)

- *REGUL*, Regulatory & Government. Concerns rewards for scientists that protect Roundup business; efforts to monitor and influence regulative bodies for possible negative rulings or ratings related to Roundup / glyphosate.

| Length (characters) | Count |
|---|---|
| $[0; 4]$ | 1175 |
| $[5; 19]$ | 1122 |
| $[20; 74]$ | 2428 |
| $[75; 124]$ | 2062 |
| $[125; 299]$ | 3165 |
| $[300; 499]$ | 573 |
| $[500; 1000]$ | 195 |
| $[1000; 3165]$ | 54 |
| $[0; 3165]$ | 10774 |

Table 10.2: Raw sentence length distribution (in characters)

We downloaded all 274 documents (emails, doc, excel, scans, etc) from lawfirm Baum, Hedlund, Aristei & Goldman[5] with human readable description of 120 links to documents[6]. We matched the documents with the human readable description. We resolved minor issues with matching document ids, links and descriptive texts. We use the four different types of sensitive information that Baum, Hedlund, Aristei & Goldman identified at the document level to label the documents (as discussed above).

Each document is annotated by with number, an id, a link, a link text and a description. An example is shown in Table 10.1. All documents are pdf documents. Some are exported from emails, word documents, and so on. Some of the documents are or contain scanned images of their text without any optical character recognition. We extracted all text encoded in the documents, but have not used OCR to transform non-text content.

Before tokenizing sentences, we removed email headers, except for the subject. We used the NLTK toolkit [12] and tokenized sentences using the Punkt sentence boundary detection approach [55], yielding $10,774$ sentences. The length distribution is shown in Table 10.2.

We cleaned the data further by removing very short sentences (4 words or less) and very long sentences (200 words or more). By doing so, we removed 3160 short sentences and 35 long sentences. We used label majority as the label for the dataset. We obtain a total of 7537 high quality sentences (see also Table 10.3).

We employ two labeling approaches to curate two sets of labels for each Monsanto datasets, to create *silver datasets* and *golden datasets*.

For the silver datasets, we assign the document label (sensitive or not with respect to each of the 4 datasets) to all sentences of that document, as provided by the lawyers at Baum, Hedlund, Aristei & Goldman. Such silver datasets thus require little human annotation effort (if we were to add more documents), as the legal experts only need to label at the document level. We thus have sensitive labels for all 7537 sentences. From documents with different labels, we uniformly at random select sentences for negative sampling for each dataset, resulting in the distribution of sentences shown in Table 10.4.

[5] https://www.baumhedlundlaw.com/toxic-tort-law/monsanto-roundup-lawsuit/monsanto-secret-documents/

[6] https://www.baumhedlundlaw.com/pdf/monsanto-documents/monsanto-papers-chart-1009.pdf

| Length (words) | Count |
|---|---|
| [5; 9] | 705 |
| [10; 19] | 2339 |
| [20; 29] | 1881 |
| [30; 39] | 1076 |
| [40; 49] | 572 |
| [50; 74] | 605 |
| [75; 99] | 203 |
| [100; 149] | 114 |
| [150; 200] | 42 |
| [5; 200] | 7537 |

Table 10.3: Final sentence length distribution (in words)

| Dataset | Total | Train | Dev | Test |
|---|---|---|---|---|
| *GHOST* | 6932 | 5900 | 500 | 532 |
| | 3466 | 2949 | 245 | 272 |
| | 50.00% | 49.98% | 49.00% | 51.13% |
| *TOXIC* | 2892 | 2200 | 340 | 352 |
| | 1446 | 1099 | 176 | 171 |
| | 50.00% | 49.95% | 51.76% | 48.58% |
| *CHEMI* | 2702 | 2100 | 300 | 302 |
| | 1351 | 1048 | 154 | 149 |
| | 50.00% | 49.90% | 51.33% | 49.34% |
| *REGUL* | 2548 | 1950 | 300 | 298 |
| | 1274 | 951 | 170 | 153 |
| | 50.00% | 48.77% | 56.67% | 51.34% |
| Total | 15074 | 12150 | 1440 | 1484 |
| | 7537 | 6047 | 745 | 745 |
| | 50.00% | 49.77% | 51.74% | 50.20% |

Table 10.4: Silver data (row 1: sentence count; row 2: sensitive sentence count; row 3: ratio of sensitive sentences)

The silver labels are representative of application scenarios where sentence labels are not available or (too) costly to obtain. In some applications, and in particular for larger documents, though, a document which contains sensitive information may also contain non-sensitive information. E.g. an email may contain greetings or best wishes which is generally not sensitive. For silver labels such documents may introduce noise. To study the impact of such noise, we also create golden labels where assignment of sensitivity is manually conducted at the sentence level. In the evaluation, we compare models constructed and tested on datasets following either labeling approach.

The golden labels are provided by 3 annotators for each sentence in a subset of about 1000 sentences. For annotation guidelines the annotators were given an introduction to the Monsanto case and the different types of sensitive information (the list introduced in the beginning of this section above), and participated in a kick-off workshop[7]. Each annotator was given the same 1073 sentences taken at random from documents labeled by the lawyers at Baum, Hedlund, Aristei & Goldman. These 1073 sentences were distributed uniformly at random across each sensitive information type. Each annotator then labels the sentence sensitive or not according to any of the sensitive information types given. We use majority of inter-annotator agreement i.e., assign sensitivity to sentences which at least 2 annotators have labeled sensitive. In our data all 3 annotators agreed on label for 65.88% of the sentences. The inter-annotator agreement can be assessed with the Fleiss Kappa [29] which takes values below or equal to 1, with 1 indicating perfect agreement and less than 0 indicating agreement by chance. Our Fleiss Kappa is 0.33 which in the rule of thumb by [62] can be considered a "fair agreement".

Distribution of labels in this golden annotated dataset is shown in Table 10.5.

We observe that $GHOST$ and $TOXIC$ have sensitive ratio around 25%, where $CHEMI$ and $REGUL$ are more skewed with a sensitive ratio around 15%.

## 10.4 Empirical Characterization

We characterize the sensitivity of information in sentences in our data resource by an empirical study of existing approaches in the field. We place particular focus on comparing silver and golden labels.

### Detection Models

Broadly speaking, the models for sensitive information detection can be divided into *keyword*-based and *context*-based [74]. Keyword-based approaches assign probabilities to words (or rather, $n$-grams) occurring in sensitive (or non-sensitive) sentences. They differ in how they utilize these probabilities [10, 16, 38, 42, 84]. Context-based approaches consider the context (beyond $n$-grams) of a word occurrence for assigning probability of a sentence being sensitive. Dense embedding approaches can be seen as a prototypical way of encoding context for a word (e.g. [71, 79]). In a context-based approach, the probability of a particular word or phrase being sensitive is allowed to vary with the context (sentence, paragraph, document) in which the word appears, allowing them to detect more complex types of sensitive information that are not characterized by (co-)occurrence of keywords alone. In this evaluation we focus on sentence level sensitive information.

[7]See also
`https://github.com/neerbek/taboo-mon/blob/master/doc/AnnotationDescription.txt`

| Dataset | Total | Train | Dev | Test |
|---|---|---|---|---|
| *GHOST* | 296 | 144 | 62 | 90 |
| | 77 | 41 | 14 | 22 |
| | 26.01% | 28.47% | 22.58% | 24.44% |
| *TOXIC* | 252 | 134 | 65 | 53 |
| | 57 | 26 | 15 | 16 |
| | 22.62% | 19.40% | 23.08% | 30.19% |
| *CHEMI* | 250 | 123 | 61 | 66 |
| | 32 | 17 | 5 | 10 |
| | 12.80% | 13.82% | 8.20% | 15.15% |
| *REGUL* | 275 | 139 | 69 | 67 |
| | 34 | 19 | 9 | 6 |
| | 12.36% | 13.67% | 13.04% | 8.96% |
| Total | 1073 | 540 | 257 | 276 |
| | 200 | 103 | 43 | 54 |
| | 18.64% | 19.07% | 16.73% | 19.57% |

Table 10.5: Golden data (row 1: sentence count; row 2: sensitive sentence count; row 3: ratio of sensitive sentences)

We quantify the complexity of our corpus by making use of these characteristic differences in keyword-based and context-based approaches, respectively. Simply put, datasets where the performance gap between the two is large, contain more complex sensitive information. We use recurrent memory cell neural networks, LSTM[44] and recursive neural networks, RecNN[27, 35, 90] as examples of context-based approaches. Both generate an embedding for each context and predict based on this context embedding.

Keyword-based approaches used are InfRule [16], C-san (C-sanitized) [84] and an empirical upper bound on keyword-based approaches we term *Keyword-Max*.

**InfRule.** One of the earliest works in the sensitive information detection domain [16] is inspired by association rule mining [2]. It considers words in a sentence as events in a probabilistic process and discovers rules which can either be simple: $w \rightarrow s$ (*word $w$ implies sensitive information $s$*) or complex combinations using conjunction, disjunction and logical not $(w_1 \wedge w_2 \wedge \neg w_3 \wedge (w_4 \vee w_5)) \rightarrow s$. The confidence of a rule is the fraction of times it occurs and predicts correctly in the training set. We follow the setup in [16] which uses InfRule on the Enron corpus using simple rules and a constant confidence cutoff.

**C-san.** [84] use point-wise mutual information (PMI) between a word $w$ and a type of sensitive information $s$ ($s$ can be a known sensitive word or inferred some other way) $PMI(s;w) = \log \frac{P(s \wedge w)}{P(s)P(w)}$, i.e., logarithm of the probability of the joint occurrence of word $w$ and sensitive information $s$, normalized by the probability of occurrences of sensitive information $s$ multiplied by the probability of occurrences of the word $w$. A sentence is considered sensitive if its PMI exceeds a sensitivity threshold. The threshold is determined using the *information content* (IC) of the sensitive information $s$, defined as the logarithm of the fraction of occurrences of sensitive information $s$: $IC(s) = -\frac{1}{\alpha}\log(P(s))$, where $\alpha$ is a user defined constant which reflects the cost of false negatives. A text is sensitive if for any word we have $PMI(s;w) \geq IC(s)$. The intuition behind this definition is that (for $\alpha = 1$) PMI is maximal if $PMI(s;w) = IC(s)$ and word $w$ will predict/disclose the information $s$ with probability 1, thus $w$ is a good predictor. By dividing $IC$ by $\alpha > 1$ we detect keyword-based predictors with lower than 1 probability and thus will be able to predict sensitive information even when perfect predictors do not exist.

**Keyword-Max.** To identify how much of the sensitive information potentially could be captured by keyword-based approaches, we include a form of (upper) empirical baseline. We allow it to set hyperparameters based on the test set, which means it is given access to additional information that in reality is not available. It is still interesting as it denotes the limit of keyword based approaches, and thereby provides a further indication of the complexity of sensitive information that cannot be captured by keyword-based approaches.

**LSTM.** The sequential LSTM approach builds a neural network model and for each word in a sentence applies the neural network in sequence. For a given text $t = (w_1, w_2, \ldots, w_n)$ and for each step consider a new word $w_i$ and apply the neural network to obtain both a new memory cell state and a hidden state. Whereas the hidden state is mainly used to parse information from one step to another, the memory cell is "protected" by several gated states which allows the LSTM to carry information across longer step counts than what is generally possible using vanilla recurrent neural networks. In our previous work [75] we built LSTM models for sensitive information detection. Prediction is based on the state arrived at after sequentially processing every word in the sentence by adding a fully connected layer. In our evaluation we apply these models on the Monsanto datasets developed here.

Please note that the LSTM could be augmented with structural information similar to the RecNN below. In our dataset characterization, we use the LSTM as a representative of unstructured sequential deep methods, and the RecNN as a structured one. Both approaches use GloVe word embeddings [79].

**RecNN.** As discussed in [74, 75] the recursive neural network, RecNN, approach has been used successfully for sensitive information detection. The use of RecNN for context dependent tasks is motivated by the previous RecNN models for e.g. sentiment analysis [90] and paraphrase detection [88]. In a RecNN we are given both the text $t$ and a structure over the text $S$. As structure here we generate probabilistic context-free grammars (pcfg) based constituent trees [56], where the pcfg was trained over the Penn Treebank [95]. Let $Y$ be the set of all nodes in the structure and all words in $t$, then the structure $S$ is a mapping from each element in $Y$ to a list of parents also in $Y$. The structure can be a directed acyclic graph (DAG). In this study we follow [88, 90] and restrict the approach to only tree-like structures. In this case the length of the list of parents is at most 1, and the list of parents is empty for the root node in the structure. As described in Section 10.3, our data resource contains constituency parse trees for each sentence (text) $t$. We follow [74] where given a sentence, the root state is the last state of the evaluation of the neural network on that sentence which may carry most information about the sentence. As for the LSTM, we add a fully connected layer for predicting sensitivity. Compared to our previous work we develop transfer learning for the RecNN model between our silver and golden dataset and show improved performance of the RecNN model.

**Experimental setup.** Both InfRule and C-san use a cutoff of minimum confidence that a keyword must have. These cutoffs are set using the dev dataset. In contrast, Keyword-Max is allowed to set that cutoff based on the data in the test set, even though that is not available in a real application. As we observe in our study, there is a limit to the sensitive information that keyword-based approaches can successfully detect, which makes it possible for us to reliably characterize complex sensitivity in our datasets. InfRule uses default parameters on Enron data as in [16], C-san $\alpha$ values used in [84], namely, $\alpha \in \{1, 1.5, 2\}$. LSTM and RecNN approaches use GloVe embeddings [79], with embedding size 100 given the relatively low number of labeled sentences. Dropout rate 0.5 was found to work well for LSTM, while lowering dropout rate for RecNN to 0.1 yielded the best results. For LSTM we obtain

| Approach | *GHOST* | *TOXIC* | *CHEMI* | *REGUL* |
|---|---|---|---|---|
| InfRule | 57.80% | 59.71% | 60.33% | 67.33% |
| C-san; $\alpha = 1$ | 49.60% | 52.94% | 54.00% | 61.33% |
| C-san; $\alpha = 1.5$ | 62.40% | 65.29% | 67.67% | 71.33% |
| C-san; $\alpha = 2$ | 72.60% | 70.29% | 71.33% | 74.33% |
| LSTM | 83.60% | **77.33**% | **86.67**% | 82.33% |
| RecNN | **86.60**% | 75.00% | 83.67% | **87.00**% |

Table 10.6: Characterizing complexity of silver label data using accuracy of keyword-based (top) and context-based approaches (bottom): keyword-based approaches can successfully capture the majority of sensitive content; more complex sensitive information is captured by deep learning methods; no existing method can fully recover all sensitive content

the best results using AdaDelta optimizer for learning rate optimization. For RecNN the best results were found using stochastic gradient decent (SGD) with learning rate determined through line search. Please note that we are mainly interested in obtaining optimal performance for each approach such that the complexity of the datasets is accurately characterized.

### Silver Labels

In the following, we characterize our data resource with the above models using silver labels for training and evaluation. For each sensitive information type we train a specific model for each of the approaches.

In Table 10.6, we characterize sensitive information complexity using silver sentence labels on reported accuracy scores[8]. We observe that InfRule generally finds more complex sensitive information than C-san when $\alpha$ is set to 1, but if this parameter is optimized, C-san captures additional sensitive information beyond InfRule results. We observe that InfRule and C-san generally perform better on *REGUL*, where differences between all models are smaller. This indicates less complex sensitive information compared to the other datasets. Additionally, we find that by giving keyword-based approaches access to test set information, in the Keyword-Max model as described above, we obtain an empirical upper limit on the less complex sensitive information as follows:

| *GHOST* | *TOXIC* | *CHEMI* | *REGUL* |
|---|---|---|---|
| 78.60% | 73.24% | 80.67% | 75.00% |

[8]More details on experiments parameters can be found in `https://github.com/neerbek/taboo-mon/blob/master/doc/ExperimentParameters.txt`

The context-based approaches LSTM and RecNN are capable of capturing more complex sensitive information beyond the keyword-based approaches. We observe that on silver labels LSTM has best performance on $TOXIC$ and $CHEMI$. These datasets both deal with discussions on cause and effect of chemical compounds and experimental design. Likely, this follows a more sequential buildup, presentation-wise, which the LSTM is particularly designed for capturing. Conversely, we observe that the structured approach RecNN which has access to the constituency parse tree for each text shows best performance for datasets $GHOST$ and $REGUL$. Both datasets contain many emails and are thus more conversational in nature. Accordingly, we observe that the RecNN outperforms LSTM here. This shows that complex sensitive information may show different structures in these datasets.

Overall, we conclude that all datasets contain sensitive information that can be captured by keyword-based approaches, but also more complex types that require advanced methods that exploit the context. We also note that none of the approaches achieves close to perfect accuracy, i.e., these datasets still provide potential for research on methods that can capture aspects of sensitivity that are not currently detected.

### Golden Labels

We now turn to the characterization of the data with respect to the golden labels. We subdivide this study into four cases and due to space considerations we restrict our characterization experiments to our most expressive model family, the RecNN. While small differences occur, the overall conclusions remain the same.

Furthermore in our 3. case we motivate the use of transfer learning between our larger silver dataset and the smaller golden dataset as a way to characterize the level of sensitive information learnable from the silver dataset. Such characterization is based on the concept of transfer learning discussed in [98] for embedding based model families and thus not as such applicable to the keyword-based approaches.

In our 4. case we return to characterization using all models, including the transfer learning models and summaries the characterizations learned over the datasets.

**1. Case: Silver-to-Golden** In this evaluation, we build silver label based models and study how well they predict golden labels. This allows an understanding of how valuable the relatively easily obtainable document-based silver labels are when compared to human labels on sentence level. Note that in the silver dataset all the labels of the golden subset are sensitive. If the models have learned to distinguish sentences containing sensitive information from noisy, falsely labeled non-sensitive sentences then the model should predict some of the sentences correctly as non-sensitive in the golden dataset simply because the model has learned the sensitive information type. Put differently, noisy sentences which are incorrectly labeled sensitive in the silver dataset may be similar to non-sensitive sentences in the silver dataset. Consider our previous example with sensitive emails. The initial greeting may be very similar to other greetings from non-sensitive emails. Thus a model may still learn to correctly label greetings as non-sensitive even though they appear in a sensitive email. When this is the case, we say that the model has successfully learned the sensitive information type, and it is an indication of the usefulness of sensitive labels for training of sensitive information detection models.

**2. Case: Golden-to-Golden** Here, models trained on golden labels are evaluated against golden label test sets. This provides insight into accuracy using sentence level human labels. A major challenge with the golden dataset is its smaller size as it is based on manual effort, which may make the models prone to overfitting. We train with different types of regularization to combat overfitting.

**3. Case: Silver-Transfer-to-Golden** The third case outlines how transfer learning models may combine both silver and golden labels to counter both issues with noise in silver labels and issues with limited training data in golden labels. It further provides an indication about the relationship between the silver and golden labels beyond Case 1. Our study is based on transfer learning for deep neural models as discussed in [98] for convolutional models (CNNs). They train a layered model on one task and then *transfer* the weights to a second task that benefits if sufficiently similar. In our study, we transfer all layers except 1 from silver models and train the final layer using the golden training set.

**4. case: Overview on Golden** We provide an overall comparison of all models to characterize the golden dataset as we did with the silver dataset in Subsection *Silver Labels* (page 126)

### Results - 1. Case: Silver-to-Golden

In Table 10.7 we show precision for each class (sensitive vs non-sensitive) as well as accuracy against the golden labels. Consider a correctly predicted sensitive label as true-positive ($tp$), a sensitive label predicted incorrectly as non-sensitive as false-negative ($fn$), a correctly predicted non-sensitive label as true-negative ($tn$) and a non-sensitive label predicted incorrectly as sensitive as false-positive ($fp$), then precision sensitive is Prec-sen $= \frac{tp}{tp+fn}$, and precision non-sensitive is Prec-Non-sen $= \frac{tn}{tn+fp}$.

Due to space limitations, we here show results only for RecNN models that capture most sensitive information in our previous evaluation. The focus in this characterization is on the relationship between silver and golden labels; a final overview also on the golden labels is provided in the final characterization.

| Dataset | Prec-Sen | Prec-Non-sen | Acc |
|---|---|---|---|
| *GHOST* | 31.82% | 61.76% | 54.44% |
| *TOXIC* | 37.50% | 83.78% | 69.81% |
| *CHEMI* | 70.00% | 51.79% | 54.55% |
| *REGUL* | 16.67% | 73.77% | 68.66% |

Table 10.7: Dive in on performance of RecNN model; Precision and accuracy on golden label test set for models using silver labels for training.

From the results in Table 10.7 we observe that models trained on silver labels do learn to correctly predict sensitive sentences vs non-sensitive sentence, even though all non-sensitive sentences in the golden test sets are labeled sensitive in the silver datasets. This demonstrates that datasets with silver noisy labels indeed provide useful training data for sensitive information detection models.

### Results - 2. Case: Golden-to-Golden

In the interest of space, we only present RecNN characterization as before (results for all models are summarized in the final overview). We train models on the training data with golden labels and evaluate on the golden test sets. As the golden datasets are relatively small due to the efforts in manually labeling on sentence level, we particularly study overfitting. For this, we show performance results on training, development (validation) and test sets, separately (Table 10.8). As expected, the models that perform well on the training data fail to generalize well to the development and test set, i.e., experience overfitting. Concretely, the models reach almost 100% accuracy on the training set, but much lower accuracy on the development and tests sets. Model hyperparameters was found through a line search on development set[9].

Except on *TOXIC* where we observe higher test score than just always predicting "non-sensitive", we observe that the overfitting results in worse generalization (i.e., test scores being lower than major class fraction). *TOXIC* seems to have a high ratio of sensitive information in the test set. The data was sampled uniformly and thus the distribution is expected to be uniform, but for small size datasets small variance in actual numbers can lead to a biases which can contribute to the score on *TOXIC*.

The overfitting is a sign that the sensitive information types are difficult to detect and require larger samples of labeled data to detect properly. If the information types could be characterized using a simple set of keywords, then we would expect our RecNN model to be able to obtain better performance. Our results in Table 10.8 implies that our sensitive information types extend beyond simple keyword based definitions and in fact contain some complex information.

[9]`https://github.com/neerbek/taboo-mon/blob/master/doc/ExperimentParameters.txt`

| Dataset | Train | Dev | Test |
|---|---|---|---|
| *GHOST* | 71.53% | 77.42% | 75.56% |
| | 100.00% | 79.03% | 75.56% |
| *TOXIC* | 80.60% | 76.92% | 69.81% |
| | 100.00% | 76.92% | 71.70% |
| *CHEMI* | 86.18% | 91.80% | 84.85% |
| | 100.00% | 83.61% | 80.30% |
| *REGUL* | 86.33% | 86.96% | 91.04% |
| | 100.00% | 82.61% | 88.06% |

Table 10.8: Accuracies on golden test set by training using golden label training set only. For each dataset, row 1 is accuracy if always predicting “non-sensitive”, row 2 RecNN accuracy. Note: 100% accuracy on training set and poor test results mean overfitting due to small training sets.

| Dataset | Acc | Non-sen |
|---|---|---|
| *GHOST* | 77.78% | 75.56% |
| *TOXIC* | 71.70% | 69.81% |
| *CHEMI* | 84.85% | 84.85% |
| *REGUL* | 92.54% | 91.04% |

Table 10.9: Accuracy obtained on golden label test set using transfer learning, i.e., trained first on silver label training set, then all but one layers fixed and finetuning the final layer using the golden label training sets.

| Approach | *GHOST* | *TOXIC* | *CHEMI* | *REGUL* |
|---|---|---|---|---|
| InfRule | 76.67% | **73.58**% | **84.85**% | 92.04% |
| C-san; $\alpha = 1$ | **77.78**% | **73.58**% | 83.33% | 91.04% |
| C-san; $\alpha = 1.5$ | 75.56% | 69.81% | **84.85**% | 91.04% |
| C-san; $\alpha = 2$ | 75.56% | 69.81% | **84.85**% | 91.04% |
| LSTM | **77.78**% | 69.81% | **84.85**% | 91.04% |
| RecNN | 75.56% | 71.70% | 80.30% | 88.06% |
| RecNN-tf | **77.78**% | 71.70% | **84.85**% | **92.54**% |

Table 10.10: Characterizing complexity of sensitive information on golden test sets using keyword-based approaches (top) and context-based approaches (bottom): keyword-based approaches capture more sensitive content on less noisy golden data as compared to silver data; across almost all models and datasets performance increases slightly; in particular, REGUL golden labels seem easiest to recover; transfer learning captures most sensitive content as it makes use of both silver and golden labels; no existing method can fully recover all sensitive content

In the next section we address the need for additional data (using transfer learning) and show increased performance for our models when we can combine golden datasets with transferred learning from the silver datasets. This indicate a key characteristics of our sensitive information datasets, namely that they do indeed carry complex sensitive information which cannot be captured by simple keyword-based approaches alone.

### Results - 3. Case: Silver-Transfer-to-Golden

We now turn to the combination of silver labels and golden labels using transfer learning. As discussed above, transfer learning allows making use of both silver and golden labels, thereby potentially counteracting noise and limited training data. We used the same models trained on the silver datasets as above for transfer learning with golden labels. We then trained a single layer model on top of these (fixed) representations. We fine-tuned hyper-parameters using line search and found adding data augmentation in the form of small amounts of random noise to the input embeddings worked well as regularization. We obtained the test accuracies shown in Table 10.9.

With transfer learning we are able to extract the most learning from the datasets, i.e., obtain the highest accuracies across the datasets. Following [98] we know that transfer learning performs well if the two tasks share similarities, which means that silver and golden labels are sufficiently related, and can thus be used for training and evaluation sensitive information detection models.

We have successfully transferred learning from the original models (the silver labels) to the golden labels. This furthermore implies that our document based silver labels actually provide knowledge which with relatively little effort can be utilized for sensitive information detection, even at the sentence level. Noise in the silver labels can thus be successfully ignored by the models used in our characterization.

Similar performance even in the face of noise in the silver labels furthermore implies that, all things being equal, a larger dataset with silver labels may be more valuable than a smaller golden label dataset. If available, the combination of the two labels in learning seems a promising approach indeed, both with respect to training and with respect to evaluation of approaches.

### Results - 4. Case: Overview on Golden

We conclude the characterization by comparing all models on the golden datasets. In Table 10.10, we provide a complete overview over results of all the models used to characterize the golden datasets. *RecNN-tf* here denotes the transfer model discussed in the previous section.

We note that the golden dataset, as seen before, provides limited training data, which means that RecNN does not perform well. The performance of the different keyword-based methods is similar in trend as we saw in Table 10.6, C-san performing better than InfRule when the $\alpha$ parameter is chosen to match the dataset. On $REGUL$ InfRule is slightly better than C-san, both worse than RecNN-tf. The RecNN-tf model performs better than the keyword based models, except for $TOXIC$ where the small dataset sizes makes the keyword based methods better.

Overall, the transfer model RecNN-tf provides the best performance and thereby indicates how much of the sensitive information can be successfully captured by the models in our study using both silver and golden labels. It thus also provides an indication of the potential for further improvement of sensitive information detection models using this data resource.

## 10.5 Conclusion

In this work, we present new, real-world datasets based on the Monsanto documents labeled by lawyers involved in the court case. We provide labels following two different labeling approaches, *golden* and *silver*, with the data - in total 8 datasets for the sensitive information detection research community.